%% file: submission_with_appendix.tex
\documentclass{article}
\usepackage{ijcai26}

\usepackage{times}
\usepackage{soul}
\usepackage{url}
\usepackage[utf8]{inputenc}
\usepackage[small]{caption}
\usepackage{graphicx}
\usepackage{amsmath}
\usepackage{amsthm}
\usepackage{booktabs}
\usepackage[switch]{lineno}

\usepackage{multirow}
\usepackage{forest}
\usepackage{longtable}
\usepackage[skins, breakable]{tcolorbox}
\usepackage{fancyvrb}
\usepackage{newfloat}
\usepackage{listings}
\usepackage{tikz}
\usetikzlibrary{positioning,arrows.meta}
\usepackage[section]{placeins}
\PassOptionsToPackage{hyphens}{url}
\usepackage[hidelinks]{hyperref}

\newcommand{\citep}[1]{\cite{#1}}
\newcommand{\citet}[1]{\citeauthor{#1}~\shortcite{#1}}

\newcommand{\fintag}[1]{\texttt{#1}}
\newcommand{\finfeat}[1]{\textit{#1}}

\title{Benchmarking Generalization in Financial Statement Fraud Detection: robust evaluation and novel tasks}

\author{
    Guy Stephane Waffo Dzuyo$^{1,2}$
    \and
    Gaël Guibon$^{2,3}$\and
    Christophe Cerisara$^2$\And
    Luis Belmar-Letelier$^1$\\
    \affiliations
    $^1$Forvis Mazars\\
    $^2$LORIA, CNRS, Université de Lorraine\\
    $^3$Université Sorbonne Paris Nord, CNRS, Laboratoire d'Informatique de Paris Nord, LIPN, F-93430 Villetaneuse, France\\
    \emails
    \{guy.stephane.waffo, luis.belmar-letelier\}@forvismazars.com,
    gael.guibon@lipn.fr,
    christophe.cerisara@loria.fr
}

\begin{document}

\maketitle

\begin{abstract}
Financial statement fraud detection (FSFD) is crucial for market integrity but faces challenges from increasingly sophisticated schemes and under-utilized textual data in financial reports. Existing methods often rely on random data splits, leading to overoptimistic performance estimates that do not reflect real-world generalization to new companies or future periods. To address this recurring problem with the state of the art, we propose a robust FSFD framework leveraging Large Language Models (LLMs) to integrate both structured financial data and unstructured textual information from financial reports. We provide a more realistic evaluation through a novel and challenging benchmark task called Company-Isolated FSFD (CI-FSFD). We construct and make publicly available a comprehensive U.S. company dataset combining financial statements, summarized MD\&A text, and fraud labels. Our approach achieves the best performance on the challenging CI-FSFD task, demonstrating the critical value of textual data and robust evaluation for reliable financial fraud detection.
\end{abstract}

\section{Introduction}
The prevalence of financial statement fraud compromises the transparency and the integrity of financial markets and results in significant economic losses for stakeholders~\citep{REZAEE2005277}. Traditional deterministic and statistical models~\citep{altman1968financial,beneish1999detection,costa_standard_2022} are often insufficient against today’s complex and sophisticated fraudulent schemes, driving the need for more advanced detection approaches to identify them. Machine learning techniques, including tree-based models and deep learning, have shown promise in this domain by leveraging structured financial data~\citep{craja_deep_2020,ali_financial_2022}. However, financial reports contain rich, unstructured textual information, such as \textbf{M}anagement \textbf{D}iscussion and \textbf{A}nalysis (\textbf{MD\&A}) sections, which often contain qualitative signals and narratives that complement numerical data, and can be indicative of fraudulent intent or misrepresentation \citep{kirkos_exploring_2024}.
Recent advancements in Large Language Models (LLMs) have demonstrated their capacity to process, understand, and reason over complex information across various domains~\citep{liu-etal-2025-findabench,xu-ding-2025-large}. This capability presents a significant opportunity to leverage the textual components of financial reports more effectively \citep{wang-brorsson-2025-large} for tasks like fraud detection. However, applying LLMs to financial data, especially for fraud detection through anomaly detection, faces unique challenges \citep{li_large_2023}. A primary limitation in current \textbf{F}inancial \textbf{S}tatement \textbf{F}raud \textbf{D}etection (\textbf{FSFD}) research is the evaluation methodology itself. Many studies simply rely on random data splitting, which can inflate performance metrics by allowing models to learn company-specific patterns or exploit temporal dependencies present in the training data, thus failing to generalize to unseen companies or future periods \citep{wang_attentive_2023}, which is the main purpose of the FSFD task. This overestimation of predictive capabilities highlights the need for more robust and realistic evaluation frameworks that better reflect real-world deployment scenarios.
To address these limitations and advance towards a more robust FSFD, we propose a novel framework grounded in two key hypotheses. Firstly, we hypothesize that (HYP1)
 generalization on companies in financial statement fraud detection is mandatory, which goes beyond the commonly used random train-test splitting. Indeed, realistic evaluation requires isolating models from specific company identities, which differs from what is currently standard practice. Secondly, we hypothesize that (HYP2) leveraging the rich textual information within financial reports, alongside traditional structured financial data, can improve fraud detection performance, particularly under these more challenging isolation conditions from HYP1.
Driven by these hypotheses, we investigate three core research questions: (RQ1) How does evaluating FSFD performance under company isolation impact the model's performance? (RQ2) To what extent does textual data from financial reports contribute to fraud detection, especially when structured data alone is less informative? (RQ3) Are LLMs capable of effectively detecting financial fraud over multimodal and multiformat financial data for robust fraud detection?\\

\noindent In this paper, we contribute as follows:
\paragraph{A Novel Task for Realistic FSFD Evaluation.} We introduce a novel task called \textbf{C}ompany-\textbf{I}solated FSFD (\textbf{CI}-FSFD) based on professional expertise. This novel task provides a more realistic assessment of model generalization capabilities compared to the current standard practice limited to traditional random splitting.
\paragraph{Multimodal Financial Dataset.} We construct and make publicly available a comprehensive dataset of U.S. companies by integrating structured financial statement data with unstructured textual information from the \textbf{M}anagement \textbf{D}iscussions and \textbf{A}nalysis sections (\textbf{MD\&A}). Crucially, we link these to fraud labels derived \textbf{A}ccounting and \textbf{A}uditing \textbf{E}nforcement \textbf{R}eleases (\textbf{AAER}s) from the U.S. \textbf{S}ecurities and \textbf{E}xchange \textbf{C}ommission's (\textbf{SEC}) through a two-stage, high-fidelity process. We detail our data extraction pipeline, including LLM-based summarization of MD\&A and our manually-audited temporal linking of AAERs labels\footnote{\url{https://www.sec.gov/enforcement-litigation/accounting-auditing-enforcement-releases}}, which ensures the accuracy of our ground-truth labels. We also detail our preparation pipeline and temporal linking of AAERs.
\paragraph{LLM-based Framework.} We propose and implement a novel LLM-based framework capable of effectively processing and combining structured numerical features and \textbf{S}ummarized text data from MD\&A (\textbf{S}MD\&A), for binary fraud classification.
\paragraph{Benchmark-leading Metrics.} We demonstrate that our LLM-based approach achieves the top performance on the novel CI-FSFD task, validating our hypotheses and highlighting the important role of both textual information and robust evaluation in FSFD.

By introducing these challenging evaluation and benchmarks we demonstrate the power of LLMs on multimodal and multiformat financial data. Our work lays a foundation for developing more reliable and generalizable financial fraud detection systems. We believe it will help the community to tackle financial statement analysis\footnote{\url{https://github.com/WaguyMz/Financial-Statements-Fraud}\break\url{-Detection}}.

\section{Related Work}

Financial fraud involves the intentional misrepresentation of financial information to deceive stakeholders or gain an unfair advantage \citep{REZAEE2005277}. This illicit activity can manifest in various forms, including revenue misstatement, asset misappropriation, and expense misstatement. The consequences of financial fraud are severe, leading to significant financial losses, legal repercussions, and reputational damage for companies and investors. While fraud can encompass issues such as disclosure violations, breaches of market regulations, bribery, and earning manipulation, the latter is most likely to be revealed through financial statements. Therefore, in this work, we define fraud as earning manipulations and the false reporting of any accounting information intended to mislead investors, regulators, customers, or other parties~\citep{10.1257/089533003765888403,10.1257/jep.25.1.49}.

\paragraph{Early Statistical Techniques.}
Early attempts to detect financial fraud relied on statistical techniques based on financial ratios. They provide an estimate of the probability of fraud by analyzing various financial ratios and identifying patterns indicative of fraudulent behavior. The Altman Z-Score focuses on bankruptcy prediction \citep{altman1968financial}, the Beneish M-Score on earnings manipulation \citep{beneish1999detection}, and the Jones model focuses on accruals \citep{costa_standard_2022}. In 2011, \citet{dechow_predicting_2011} proposed an efficient approach based on a logistic regression model and 7 features to predict earning mistatement.

\paragraph{Machine Learning Approaches.}
The rise of machine learning in the 2000s spurred research into more advanced techniques for FSFD, such as Deep Neural Networks~\citep{krizhevsky2012imagenet} and Random Forests~\citep{breiman2001random}. Later in the 2010s, the increased accessibility of powerful Natural Language Processing methods like LSTMs~\citep{lstm} enabled researchers to leverage textual information within financial reports. For example, \citet{craja_deep_2020} used the Management Discussion and Analysis (MD\&A) sections of Form-10K\footnote{Form 10-K is the comprehensive annual report that public companies file with the SEC. Form 10-Q is a quarterly report about the company's financial performance during the quarter.}, where executives explain financial performance, to enhance the performance of their classification model.
In 2023, \citet{wang_attentive_2023} tackled financial statement fraud detection with a novel model, RCMA, emphasizing the importance of attentive mechanisms for distinguishing between modalities and coordinate financial ratios with textual data from financial reports.
By addressing fusion ambiguity, their approach achieved strong fraud detection performance on CSMARD\footnote{The China Stock Market \& Accounting Research Database (CSMARD) offers data on the China stock markets and the financial statements of China’s listed companies.}.

\paragraph{The Emergence of Large Language Models (LLMs).} LLMs \citep{radford2019language,touvron2023llama} offer new avenues for complex tasks like FSFD. Initial studies, such as \citet{kirkos_exploring_2024} using ChatGPT-4 on CEO letters and \citet{kim_financial_2024} on general financial statement analysis, have highlighted LLMs' potential in understanding financial narratives. However, these often rely on closed-source models, posing reproducibility challenges.
\citet{bhattacharya_accounting_2024} fine-tuned a BERT model using truncated MD\&A sections, potentially missing key information. This underscores the need for LLM-based FSFD approaches that can utilize the extensive textual data in financial reports, a gap our work addresses.

\paragraph{Frameworks for Financial Statement Fraud Detection (FSFD).}
Prevailing FSFD evaluation using random data splitting often yields overoptimistic performance, as models may learn company-specific or time-bound artifacts rather than generalizable fraud indicators, failing to reflect real-world deployment challenges. To address this, we introduce more realistic evaluation via a novel task: \textbf{C}ompany-\textbf{I}solated FSFD (\textbf{CI-FSFD}), evaluating generalization to unseen companies. To our knowledge, this is the first work to establish dedicated benchmark for that specific setting, aiming for more reliable assessments of FSFD systems.

\paragraph{Open Data and Datasets.}
A significant challenge in FSFD research is the scarcity of readily available, open datasets. While U.S. AAERs \citep{sec_aaer} provide public fraud instances and China's CSMARD \citep{CSMAR2025} offers extensive data for Chinese markets, integrating these primary sources with structured financial statements (often in XBRL format\footnote{XBRL (eXtensible Business Reporting Language): \url{https://www.xbrl.org/the-standard/what}}) and textual MD\&A sections (both also available from the SEC) requires intensive, non-trivial preprocessing and accurate temporal linking. Curated datasets that perform this integration, such as those available through commercial providers like the Compustat database~\citep{compustat_wrds}, often come at a significant cost, limiting accessibility for widespread research. We address this gap by constructing and making publicly available a comprehensive FSFD dataset for U.S. companies, which combines financial statements, temporally linked AAERs, and processed MD\&A text, and we detail our data collection and preprocessing pipeline.

\section{Data Collection and Preprocessing}
Our FSFD dataset integrates financial data, textual information from Form 10-Q's MD\&A sections, and fraud labels. This involved extracting, cleaning, and structuring these components from various sources, detailed below.

\subsection{Financial Data}
\label{subsec:financial-data}
Quarterly financial data (Forms 10-Q) from 2009-2024 were sourced from the SEC website. Using the US-GAAP taxonomy, we mapped items to core accounts and, following \citet{Waffo_Dzuyo_Guibon_Cerisara_Belmar-Letelier_2025}, processed XBRL data to extract raw metrics (e.g., Total Revenue) and impute missing values. From these, we engineered \textbf{122} financial indicators (raw figures, change-based, ratios). For quality, reports with less than 25\% of these features present were removed, resulting in \textbf{268,936} firm-quarter reports from \textbf{13,332} companies. Appendix A details the extraction and list all features.

\subsection{Text Data}
We collected 195,023 quarterly MD\&A sections (Forms 10-Q, 2009-2024) via a paid SEC-API\footnote{https://sec-api.io/}. These raw HTML sections were lengthy and variable (1k-150k tokens, avg. 14k), making direct LLM processing computationally challenging. To make this text tractable, we summarized each section using the pretrained and open-source Qwen3 32B ~\citep{yang2025qwen3technicalreport}. This step filters out non-material boilerplate legal language. Because forensic accounting anomalies (e.g., transaction misstatements) are embedded within hard factual disclosures rather than subtle linguistic style shifts, utilizing Qwen3 32B distills core factual triggers while minimizing context distraction for the classifier.    This yielded concise summaries averaging 3,800 tokens, forming our Summarized MD\&A dataset, referred as SMD\&A.

\subsection{Fraud Dataset Preprocessing}
\label{sub:fraud_prep}
Sourcing and aligning fraud labels is a critical step in constructing a robust FSFD dataset. Our fraud labels are derived from 3,300 AAERs obtained via the SEC-API. A significant challenge arises because the machine-readable JSON summaries for these releases lack the specific fiscal years and quarters of the violations, preventing a direct link to our quarterly financial data.
To overcome this, we implemented a two-stage pipeline to guarantee the accuracy of our ground-truth labels.
\paragraph{Stage 1: Automated Extraction.} First, we scraped the full, detailed legal documents linked within each AAER summary. We then leveraged the long-context capabilities of the Qwen3 32B model as a powerful parsing assistant. Using a structured prompt, we tasked the LLM with identifying and extracting a preliminary set of key information from each document: the fraudulent company or companies involved, a description of the fraudulent scheme, a list of fine-grained fraud categories, based on the 11 earning misstatement types proposed by \citet{dechow_predicting_2011}, which we augmented with an additional \fintag{Assets misstatement} label (details in Appendix C).

 \paragraph{Stage 2: Manual Audit and Verification.} Each of the 249 AAERs processed by the LLM was individually reviewed by one human expert in both Machine Learning and Auditing, who cross-referenced the extracted company, fiscal quarter(s), and fraud categories against the original legal source documents. This meticulous verification process confirmed that every extracted data point was correct, resulting in perfect accuracy for our fraud labels. This process yielded a set of 1,451 firm-quarter reports identified as fraudulent between 2000 and 2022. These verified instances form our core binary fraud labels which will serve to further construct the dataset.

Additionally, our fraud dataset preprocessing involves extracting 12 fine-grained fraud labels. Although these labels could serve as valuable features for an advanced multi-label classification task, the scope of the current work is limited to binary classification to demonstrate the robustness of the novel CI-FSFD task.

\subsection{Final Dataset Construction}
\label{subsec:class-imbalance-handling}
The final dataset construction involved 3 key steps: merging the datasets, handling class imbalance, and ensuring temporal and company consistency.

\paragraph{Merging Datasets.}
We merged the financial data, text data, and fraud labels. The financial and text data were aligned using company identifiers and fiscal quarters, creating distinct firm-quarter instances. The fraud labels, derived from AAERs, were linked to the financial and text data based on the extracted fiscal quarters.

\paragraph{Critical Class Imbalance Handling.}
Financial fraud is an inherently rare event, leading to extreme class imbalance; in our raw dataset, fraud cases are only about 0.03\% of firm-quarter observations. Training directly on such severe imbalance biases models towards the majority (non-fraud) class. Following rare-event ML paradigms, we target a stable 5\% distribution. This preserves a realistic, severe class imbalance while ensuring gradient stability during training. Post-hoc threshold calibration via validation F1-maximization ensures the model remains optimized for precision under these imbalanced constraints. This initial downsampling is done by preserving original industry and time distributions.

\paragraph{Final Dataset Statistics.}
After merging and downsampling, the final dataset consists of 10,159 samples (\textbf{511} fraud cases and \textbf{9,648} non-fraud cases). The distribution of samples across industries and time periods was maintained to ensure generalization and realistic evaluation.

\section{Tasks Definition}
\label{sec:tasks-definition}
\subsubsection{Classic FSFD.}
In the common setting of binary fraud detection, the dataset usually consists of sets of firm-quarters observations either labelled as fraud or not. The dataset is split randomly into train and test sets, with the goal of predicting whether a given firm-quarter observation is fraudulent or not.

\subsubsection{CI-FSFD: Company Isolated FSFD.}
In the classic FSFD, random splitting of the dataset can lead to overfitting, as the model may learn to recognize specific patterns of individual companies. In contrast, our CI-FSFD task requires the model to generalize across different companies, ensuring that it can accurately identify frauds in firms it has never encountered before. This novel task is particularly relevant in real-world scenarios where models must be deployed to detect fraud in new companies.

\section{Fraud Supervised Classification}
\subsubsection{Input Data.}
To train and evaluate our models, we explored three feature sets derived from our processed data. First, we used \textbf{Financial Data Only (FIN)}, which comprises the 122 engineered financial indicators detailed in Appendix A. These indicators cover a range of metrics including raw figures, change-based values, financial ratios, and Beneish M-Score components. Second, we employed \textbf{Text Data Only (SMD\&A)}, consisting solely of the summarized quarterly MD\&A sections. Finally, we utilized \textbf{Combined Financial and Text Data (FIN+SMD\&A)} to leverage information from both sources. For this combined input, we serialized the 122 structured financial indicators into a key-value string (e.g., "Total Revenue: 123456, Net Income: 7890, ..."), which was then directly concatenated with the SMD\&A text. This straightforward fusion approach unified both modalities into a single text sequence for the model prompt (shown in Appendix F).

\subsubsection{Network Design.}
\label{ssec:network_design}
\begin{figure}[t]
    \centering
    \includegraphics[width=0.95\columnwidth]{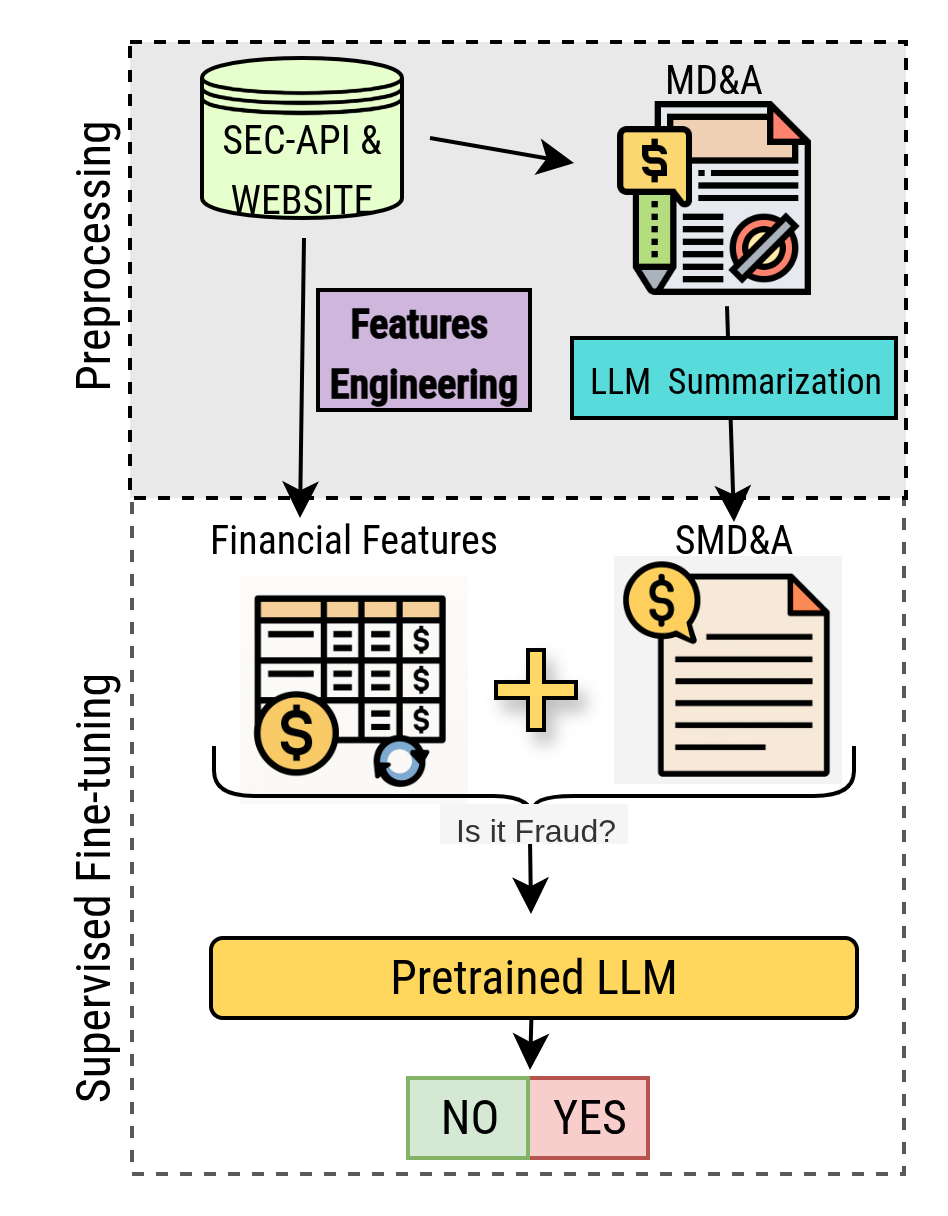}
    \caption{Financial Statement Fraud Classification.}
    \label{fig:classification-network}
\end{figure}
We propose a Large Language Model (LLM) based framework for financial fraud detection. Our primary model employs a pretrained LLM.
The input to the LLM is a carefully constructed prompt that defines the binary fraud detection task. This prompt includes the relevant financial (FIN) and/or textual (SMD\&A) data for a given firm-quarter, and is structured to elicit a classification response.
The LLM is not fine-tuned on the autoregressive language modeling
objective (i.e., predicting every subsequent token in the input sequence),
but rather on predicting the final target token in the
sequence, which represents the classification decision: either "YES"
(indicating fraud) or "NO" (indicating non-fraud). Figure
\ref{fig:classification-network} shows an overview of our classification
approach. The specifics of the fine-tuning methodology are detailed in section \ref{sec:expes}.

\subsubsection{Class Imbalance.}
Financial fraud is an inherently rare event, leading to highly imbalanced datasets. To mitigate the risk of models becoming biased towards the majority (non-fraud) class, we implement epoch-level undersampling during training. At each training epoch, we dynamically undersample the non-fraud cases from the training set to match the number of fraud instances. This prevents the model from overfitting to the majority class while still utilizing the full diversity of the majority class samples across different epochs.

\section{Experiments and Results}
\label{sec:expes}
This section details the experimental setup, baseline models, evaluation metrics, and the results obtained for the different FSFD tasks (Classic FSFD and CI-FSFD). We also present results for zero-shot performance of pretrained LLM on FSFD for further comparison.

\subsubsection{Data Splitting.}
For both the \textbf{Classic FSFD} and \textbf{CI-FSFD} tasks, we employ a
5-fold cross-validation strategy on the 10,159 firm-quarter observations. In
Classic FSFD setting, folds are created by randomly splitting these
observations. For CI-FSFD, company isolation is enforced: all firm-quarter
data from a specific company belong exclusively to either the training or
test set within a fold. This company-based splitting also maintains the
dataset's original industry sector distribution and approximates a 5\% fraud ratio
across folds.  Detailed information on these data splitting methodologies is
provided in Appendix D.

\subsection{Baseline Models}
We compare our LLM-based approach against several established and contemporary baselines:

\paragraph{Dechow Model.}
We include the logistic regression model proposed by \citet{dechow_predicting_2011}, hereafter referred to as \textbf{LR-DECHOW}. It is reference logistic regression model, built on 7 financial features and widely recognized benchmark in prediction of  earning misstatements. Its features are calculated from our 122 financial features. Appendix A.8 provides details on these features.

\paragraph{Multi-Layer Perceptron (MLP).}
 The MLP serves as a strong baseline for structured financial data. It is trained on the full set of 122 engineered financial indicators (FIN). Hyperparameters, including the number of layers and neurons per layer, are optimized using a Bayesian optimization approach via Hyperopt \citep{bergstra2013making} to maximize the average AUC over the 5 validation sets.

\paragraph{Tree-Based Ensemble Models.}
We benchmark our approach against tree-based ensemble models: Random Forest~\citep{breiman2001random}, LightGBM~\citep{ke_lightgbm_nodate}, and XGBoost~\citep{DBLP:journals/corr/ChenG16}. These methods are widely recognized for their robustness and efficacy in classification tasks, including financial fraud detection \citep{ashtiani_intelligent_2022}. Random Forest aggregates multiple decision trees to improve stability, while LightGBM and XGBoost, advanced gradient boosting frameworks, are known to offer optimized performance and scalability. We train them all using the 122 engineered financial indicators (FIN), with their respective hyperparameters tuned through Hyperopt as above.

\paragraph{RCMA-adapted.}
We developed and benchmarked an adapted implementation of the Ratio-Chapter-Modality-Aware (RCMA) model \citep{wang_attentive_2023}. This adaptation was necessary due to two primary factors: the original model's text subnetwork relies on legacy methods (Doc2Vec and LSTMs), and its source code and hyperparameters are not publicly available. 
Our principal modification was to replace those legacy methods by Jina Embedding V2-Small~\citep{nussbaum2025nomicembedtrainingreproducible}, a modern, open-source SentenceBERT model with a long-context architecture~\citep{reimers_sentence-bert_2019}, to align the model with current best practices. We fine-tuned this new component with LoRA~\citep{DBLP:journals/corr/abs-2106-09685}. For all other hyperparameters, we performed a grid search optimization to find the best configuration.

Crucially, the remainder of the RCMA architecture was replicated as faithfully as possible to \citet{wang_attentive_2023}'s description, especially its core modality-aware attention mechanisms. This ensures that our benchmark is a fair and up-to-date representation of the RCMA design. More details are provided in Appendix E.

\subsection{Experimental Setup}
\label{sec:trainingsetup}

We fine-tune two foundation large language models: the general-purpose Llama-3.1 8B~\citep{touvron2023llama} and the domain-specific Fino1-8B~\citep{qian2025fino1transferabilityreasoningenhanced}. To ensure computational tractability, base models were loaded using 4-bit quantization~\citep{frantar_gptq_2023,zheng2024llamafactoryunifiedefficientfinetuning}. We utilized Low-Rank Adaptation (LoRA)~\citep{DBLP:journals/corr/abs-2106-09685} for parameter-efficient fine-tuning, applying adapters to all linear layers.
Experiments were run on a single NVIDIA H100 GPU, requiring approximately 4 hours per fold. Hyperparameters are detailed in Appendix E.

\subsection{Evaluation Methodology}

Our primary evaluation metric is the ROC AUC score, a widely recognized standard for tasks with significant class imbalance~\citep{fawcett2006introduction}. We supplement this with standard classification metrics: precision, recall, and F1-score.

Our model selection and calibration process follows a two-stage approach on a validation set (10\% of the training data). First, we select the model checkpoint that achieves the highest ROC AUC. Second, using this chosen model, we determine an optimal decision threshold by maximizing the F1-score on the same validation data. This final model is then used to generate predictions on the held-out test set. 

\subsection{Classic FSFD Results}

\begin{table*}[t]
  \centering
      \begin{tabular}{p{2.5cm}p{2.5cm}p{2.5cm}p{2.5cm}p{2.5cm}p{2.5cm}}
    \toprule
    \textbf{Model} & \textbf{Input} & \textbf{AUC}  $\pm$ {\small stdev} & \textbf{F1}  $\pm$ {\small stdev} & \textbf{Precision}  $\pm$ {\small stdev} & \textbf{Recall}  $\pm$ {\small stdev} \\
    \midrule
    LR-DECHOW & FIN& 0.68 $\pm$ 0.0272 & 0.15 $\pm$ 0.0135 & 0.09 $\pm$ 0.0121 & 0.53 $\pm$ 0.1530 \\

    MLP & FIN &  0.89 $\pm$ 0.0098 & 0.40 $\pm$ 0.0827 & 0.30 $\pm$ 0.1080 & 0.70 $\pm$ 0.0800 \\

    LightGBM & FIN & 0.95 $\pm$ 0.0098 & 0.74 $\pm$ 0.0355 & 0.84 $\pm$ 0.0541 & 0.66 $\pm$ 0.0451 \\

    XgBoost & FIN & \textbf{0.96 $\pm$ 0.0108} & \textbf{0.76 $\pm$ 0.0930} & 0.84 $\pm$ 0.0604 & 0.69 $\pm$ 0.1240 \\

    Random Forest & FIN & 0.92 $\pm$ 0.0119 & 0.54 $\pm$ 0.0245 & 0.52 $\pm$ 0.0443 & 0.57 $\pm$ 0.0366 \\

    RCMA-adapted  & FIN+SMD\&A & 0.89 $\pm$ 0.0081 & 0.39 $\pm$ 0.0598 & 0.28 $\pm$ 0.0705 & 0.71 $\pm$ 0.0572 \\

    \midrule

    Fino1 8B       & FIN      & 0.90 $\pm$ 0.0195 & 0.46 $\pm$ 0.1032 & 0.38 $\pm$ 0.1564 & 0.70 $\pm$ 0.1162 \\

    Fino1 8B       & SMD\&A       & 0.95 $\pm$ 0.0178 & 0.66 $\pm$ 0.0818 & 0.56 $\pm$ 0.1049 & 0.83 $\pm$ 0.0584 \\

    Fino1-8B       & FIN+SMD\&A & 0.94 $\pm$ 0.0094 & 0.60 $\pm$ 0.0920 & 0.50 $\pm$ 0.1169 & 0.80 $\pm$ 0.0573 \\

    \midrule

    Llama-3.1 8B    & FIN    & 0.93 $\pm$ 0.0104 & 0.44 $\pm$ 0.0819 & 0.33 $\pm$ 0.0968 & 0.79 $\pm$ 0.0973 \\

    Llama-3.1 8B    & SMD\&A       & \textbf{0.96 $\pm$ 0.0184 }& \textbf{0.76 $\pm$ 0.0842} & 0.71 $\pm$ 0.1354 & 0.84 $\pm$ 0.0430 \\

    Llama-3.1 8B    & FIN+SMD\&A   & 0.95 $\pm$ 0.0123 & 0.71 $\pm$ 0.0824 & 0.69 $\pm$ 0.1665 & 0.77 $\pm$ 0.0605 \\
    \bottomrule

  \end{tabular}
  \caption{Performance on the Classic FSFD task over 5 folds with standard deviation (stdev)}
  \label{tab:fsfd_results_full}
\end{table*}

The results for the Classic FSFD task, detailed in Table \ref{tab:fsfd_results_full}, show high performance across most models. The Llama-3.1 8B configuration achieved the best AUC of 0.96, and nearly all approaches surpassed an AUC of 0.89, with the LR-DECHOW model being the only exception.

However, we argue that this high performance is more indicative of a methodological artifact than true generalization capability. The random splitting protocol results in significant data leakage, where the same companies appear in both training and evaluation sets. Our analysis confirms this issue: on average, each of the 321 fraudulent firms is present in 3.35 folds. This setup incites models to memorize company-specific patterns instead of learning robust fraud signals. The strong results reported here and in the literature~\citep{wang_attentive_2023,li_detecting_nodate} should be interpreted with caution, as they likely reflect this evaluation flaw. This observation responds to our research question (RQ1).

\subsection{Company-Isolated FSFD Results}
\begin{table*}[t]
  \centering

  \begin{tabular}{p{2.3cm}p{1.9cm}p{1.9cm}p{1.9cm}p{1.9cm}p{1.9cm}p{2.6cm}}
    \toprule
    \textbf{Model} & \textbf{Input} & {\textbf{AUC}} & {\textbf{F1}} & {\textbf{Precision}} & {\textbf{Recall}} & {\textbf{$\Delta$AUC (p-value)}} \\
    \midrule
    LR-DECHOW & FIN& 0.67 $\pm$ 0.04 & 0.13 $\pm$ 0.02 & 0.07 $\pm$ 0.01 & 0.68 $\pm$ 0.08 & -0.074 (p=0.000) \\
    MLP & FIN &  0.69 $\pm$ 0.06 & 0.14 $\pm$ 0.06 & 0.10 $\pm$ 0.04 & 0.39 $\pm$ 0.29 & -0.058 (p=0.000) \\
    LightGBM & FIN &  0.68 $\pm$ 0.01 & 0.15 $\pm$ 0.03 & 0.11 $\pm$ 0.03 & 0.25 $\pm$ 0.08 & -0.085  (p=0.000) \\
    XgBoost & FIN & 0.66 $\pm$ 0.04 & 0.13 $\pm$ 0.03 & 0.08 $\pm$ 0.02 & 0.38 $\pm$ 0.18 & -0.080  (p=0.000) \\
    Random Forest & FIN & 0.70 $\pm$ 0.03 & 0.16 $\pm$ 0.03 & 0.10 $\pm$ 0.02 & 0.47 $\pm$ 0.18 & -0.042 (p=0.000) \\
    RCMA-adapted  & FIN+SMD\&A   & 0.65 $\pm$ 0.00 & 0.14 $\pm$ 0.01 & 0.08 $\pm$ 0.01 & 0.71 $\pm$ 0.08 & -0.134 (p=0.000) \\
    \midrule
    Fino1 8B       & FIN      & 0.69 $\pm$ 0.04 & 0.14 $\pm$ 0.03 & 0.10 $\pm$ 0.03 & 0.49 $\pm$ 0.29 & -0.049 (p=0.000) \\
    Fino1 8B       & SMD\&A   & \bfseries 0.74 $\pm$ 0.03 & \bfseries 0.18 $\pm$ 0.04 &  0.16 $\pm$ 0.03 & 0.23 $\pm$ 0.08 & \textbf{(Reference)} \\
    Fino1 8B       & FIN+SMD\&A & 0.72 $\pm$ 0.01 & 0.17 $\pm$ 0.03 & 0.12 $\pm$ 0.05 & 0.46 $\pm$ 0.18 & -0.026 (p=0.002) \\
    \midrule
    Llama-3.1 8B    & FIN   &  0.68 $\pm$ 0.04 & 0.12 $\pm$ 0.07 & 0.15 $\pm$ 0.08 & 0.24 $\pm$ 0.18 & -0.067 (p=0.000) \\
    Llama-3.1 8B    & SMD\&A       & 0.68 $\pm$ 0.04 & 0.14 $\pm$ 0.01 & 0.09 $\pm$ 0.01 & 0.43 $\pm$ 0.08 & -0.066 (p=0.000) \\
    Llama-3.1 8B    & FIN+SMD\&A   & 0.68 $\pm$ 0.04 & 0.13 $\pm$ 0.04 & 0.08 $\pm$ 0.01 & 0.35 $\pm$ 0.20 & -0.067 (p=0.000) \\
    \bottomrule
  \end{tabular}

  \caption{
    Performance on the Company-Isolated FSFD (CI-FSFD) task over 5 folds.
    Metrics are reported as mean $\pm$ standard deviation.
    The final column displays the results of a paired bootstrap test comparing each model against the top performer (Fino1 8B on SMD\&A, in bold).
    This test reports the mean difference in AUC ($\Delta$ AUC) and the corresponding empirical p-value, calculated from 5,000 bootstrap iterations.
  }
  \label{tab:cifsd_results_full}
\end{table*}
The company-isolated evaluation, presented in Table \ref{tab:cifsd_results_full}, provides a more rigorous test of model generalization by preventing data leakage. The dramatic drop in performance for all models validates our hypothesis (HYP1) that the Classic FSFD task is prone to optimistic bias.

Against this challenging backdrop, a clear pattern emerges. The Fino1-8B model, when leveraging only narrative SMD\&A data, significantly outperforms all other configurations, achieving a leading AUC of 0.74 and an F1-score of 0.18. This result also highlights the value of domain specialization, as Fino1-8B consistently surpassed the general-purpose Llama-3.1 8B across all data modalities. Interestingly, this text-only model is more effective than the same LLM using financial data (AUC 0.69) or even the combination of both data types (AUC 0.72). The superiority of this approach over the best-performing classical model, Random Forest (AUC 0.70), further highlights the unique advantage of LLMs in this context.

\begin{table}[t]
\centering
\small
\begin{tabular}{p{1.74cm}p{1.7cm}cc}
\toprule
\textbf{Model} & \textbf{Input} & \textbf{AUC} $\pm$ {\small stdev} & \textbf{F1} $\pm$ {\small stdev} \\
\midrule
Fino1 8B & FIN & 0.48 $\pm$ 0.07 & 0.00 \\
Fino1 8B & SMD\&A & 0.52 $\pm$ 0.05 & 0.00 \\
Fino1 8B & FIN+SMD\&A & 0.51 $\pm$ 0.04 & 0.00 \\
\midrule
Llama-3.1 8B & FIN & 0.49 $\pm$ 0.04 & 0.10 $\pm$ 0.01 \\
Llama-3.1 8B & SMD\&A & 0.49 $\pm$ 0.04 & 0.10 $\pm$ 0.01 \\
Llama-3.1 8B & FIN+SMD\&A & 0.52 $\pm$ 0.05 & 0.09 $\pm$ 0.00 \\
\midrule
Qwen3 32B & FIN & 0.47 $\pm$ 0.04 & 0.02 $\pm$ 0.01 \\
Qwen3 32B & SMD\&A & 0.48 $\pm$ 0.05 & 0.00 $\pm$ 0.01 \\
Qwen3 32B & FIN+SMD\&A & 0.47 $\pm$ 0.04 & 0.02 $\pm$ 0.01 \\
\bottomrule
\end{tabular}
\caption{Zero-shot FSFD performance (mean over 5 folds, with standard deviation, stdev). All models perform extremely poorly, needing finetuning. Threshold for computing F1 was set to 0.5.}
\label{tab:zeroshot_results_full}
\end{table}

\subsection{Discussion of Results}
Our LLM-based FSFD framework, particularly with summarized textual (SMD\&A) data, yields significant insights, emphasizing the need for robust evaluation and thus validating our first hypothesis (HYP1). The substantial performance drop observed in Company-Isolated (CI-FSFD) scenarios vividly demonstrates how traditional random splitting inflates real-world generalization estimates. Notably, SMD\&A text proved highly valuable, consistently boosting model discrimination (higher AUC). 
Fino-1's specialized financial fine-tuning enabled it to outperform the general-purpose Llama-3.1 8B model across all input types in the challenging Company-Isolated Financial Statement Fraud Detection (CI-FSFD) task. Interestingly, the combined input (FIN+SMD\&A) underperformed compared to text alone, rejecting HYP2. We intentionally used simple serialization to establish a clean baseline; these results reveal a textual "noise bottleneck," proving that naive concatenation distracts the LLM and highlighting the need for future non-linear cross-modal gating structures. Further, our statistical analysis, using a bootstrap test with 1,000 iterations per fold, confirmed that this model significantly outperforms all others in terms of AUC score, with all observed empirical p-values being zero, except for one (Table \ref{tab:cifsd_results_full}). 
Finally, the consistently poor zero-shot LLM performance (AUC 0.50) confirms that LLMs' pretrained alone are inadequate for fraud detection, highlighting the need for specialized FSFD fine-tuning.

\begin{figure}[t]
\centering
\includegraphics[width=\columnwidth]{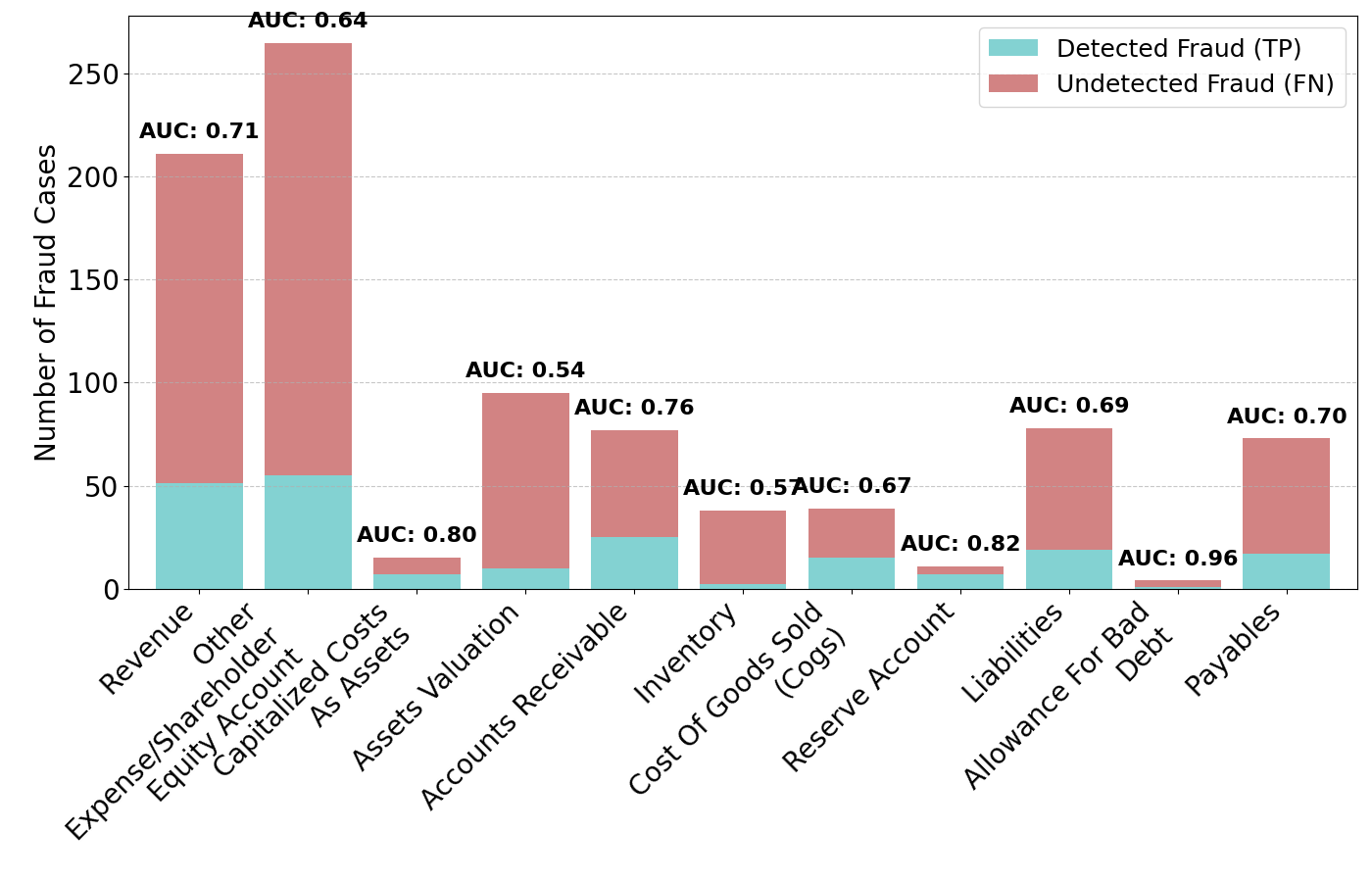}
\caption{Detection Performance per Misstatement type on the CI-FSFD task (Fino1 8B with SMD\&A input).The average AUC per mistatement is also reported above the bars.}
    \label{fig:fine-grain-perf}
    \vspace{-0.5cm}
\end{figure}

\paragraph{Fine-Grained Labels Analysis.}
To gain a deeper understanding of our model's performance on different types of financial misstatements, we conducted a post-training analysis using the 12 fine-grained fraud categories extracted from the AAERs (details in Appendix C). Utilizing the predictions from our best-performing model (Fino1 8B with SMD\&A input) on the CI-FSFD task, we computed the average AUC for each of these misstatement types.
Figure \ref{fig:fine-grain-perf} shows detection performance by category. "\texttt{Other Expense/Shareholder Equity Account}" (AUC 0.64) and "\texttt{Revenue}" (AUC 0.71) are the most frequent and relatively well-detected misstatement types. In contrast, "\texttt{Assets Valuation}" recorded the lowest AUC (0.54), indicating it is particularly challenging to consider for fraud detection.

\subsection{Explainability}
\begin{figure}
    \centering
    \includegraphics[width=\linewidth]{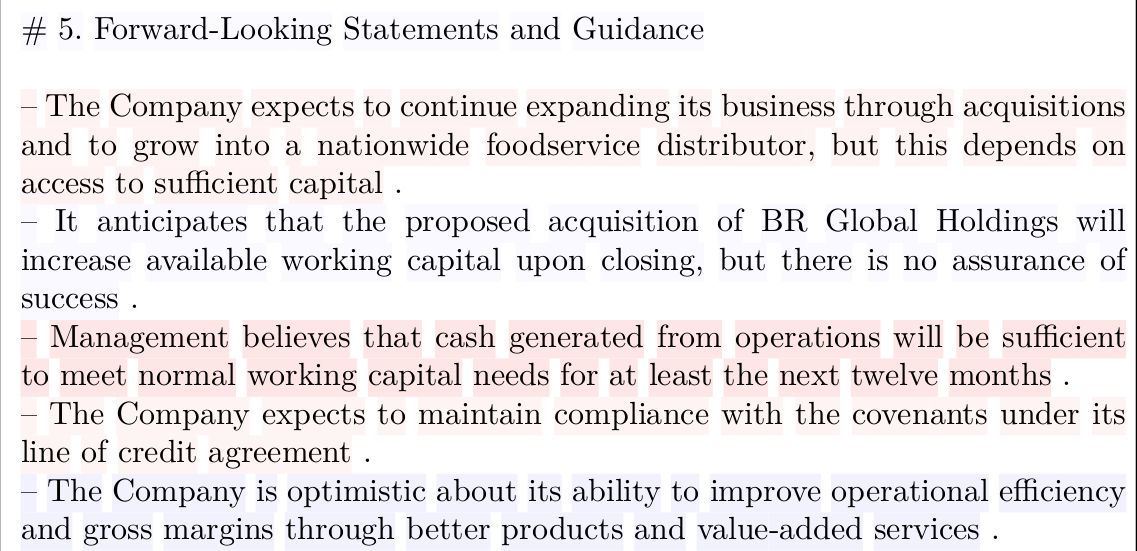}
    \caption{Attn-LRP sentence-level relevancy. Red highlights mean positive contribution to Fraud prediction and blue ones mean negative contribution, with according intensity.}
    \label{fig:sample_explainability}
    \vspace{-0.5cm}
\end{figure}

In order to explain the LLM classification, we employ AttnLRP~\citep{pmlr-v235-achtibat24a}, a technique that calculates the relevancy of each input token to the LLM predictions using a gradient-perturbation of the input signal. For each sentence of the SMD\&A document, we aggregate the relevance scores of all constituent tokens to derive a sentence-level relevance score. This approach allows us to identify and highlight key sentences that significantly influence the model's decisions as shown in Figure \ref{fig:sample_explainability}. We acknowledge that those scores do not directly elicit explanations, but they can serve as clues, helping human experts to understand the model's prediction.

\section{Limitations}

Our study, though it advances financial statement fraud detection (FSFD), has some limitations.
First, The low F1-score (0.18) reflects the severe difficulty of cross-company generalization without identity leakage. However, this baseline is valuable to help human experts narrow down audit spaces, rather than acting as an automated judge.

Second, while CI-FSFD eliminates company identity leakage, it does not strictly enforce chronological sequencing (e.g., historical-to-future splits). Merging company isolation with explicit rolling time windows is a crucial next trajectory for this benchmark to entirely prevent forward-looking bias.

Third, our experiments are confined to the U.S. SEC dataset. To establish global generalizability, future work must extend this empirical evaluation to cross-country and multi-jurisdictional contexts using external databases such as CSMARD \citep{CSMAR2025}.

Finally, our approach assumes fraud signals reside primarily within factual disclosures, which may filter out subtle stylistic or linguistic anomalies. Future work should explore hybrid architectures that ingest original MD\&A texts to capture a broader spectrum of behavioral fraud indicators.

\section{Conclusion}
Financial statement fraud detection is essential for market integrity, yet it faces significant challenges due to the sophistication of fraudulent schemes and subpar evaluation methods that often overestimate real-world performance. To address these issues, we introduced the novel Company-Isolated Financial Statement Fraud Detection (CI-FSFD) task that better evaluates models' ability to generalize to unseen companies compared to standard practice. We created and publicly released a comprehensive dataset integrating structured financial data, summarized Management Discussion and Analysis (SMD\&A) texts, and fraud labels derived from SEC AAERs. Our experiments showed that fine-tuned LLMs, particularly the specialized Fino-1 8B model using SMD\&A data, outperformed other models. These results highlight the critical value of both structured and textual data in fraud detection and underscore the importance of robust evaluation frameworks and the poor zero-shot performance of LLMs emphasizes the necessity of task-specific fine-tuning. This work provides crucial benchmarks and resources, paving the way for more reliable fraud detection  systems and evaluations.

\bibliographystyle{named}
\bibliography{custom,ai_in_audit_all_copy,models_copy,anomaly_detection_copy,miscellaneous_copy} 

\newpage

\appendix
\include{appendixes/appendix_1_financial_features_details}
\include{appendixes/appendix_2_mda_dataset_details}
\include{appendixes/appendix_3_aaer_dataset_details}
\include{appendixes/appendix_4_dataset_splitting}

\include{appendixes/appendix_5_hyperparams}
\include{appendixes/appendix_6_system_prompts}

\include{appendixes/appendix_7_sample_predictions}
\include{appendixes/appendix_8_subgroup_performance_analysis}
\include{appendixes/appendix_9_explainability}

\end{document}

%% file: appendixes/appendix_1_financial_features_details.tex
\onecolumn
\section{Appendix A : Details on Financial Features}
\label{appendix:financials}

The financial data used in this study are derived from quarterly reports (Forms 10-Q and 10-K) sourced from the SEC, covering the period from 2009 to 2024. The process involved meticulous extraction, imputation, feature engineering, and quality control to construct a robust set of financial indicators for fraud detection.

\subsection{Data Preparation Overview}
Raw financial metrics were initially extracted by mapping reported items from company filings to the standardized US-GAAP (Generally Accepted Accounting Principles) taxonomy. This taxonomy provides a hierarchical structure for financial reporting elements.
The quarterly financial datas are downloaded freely form the SEC -Website : \url{https://www.sec.gov/data-research/sec-markets-data/financial-statement-data-sets}

\subsubsection{Taxonomy-based Data Imputation}
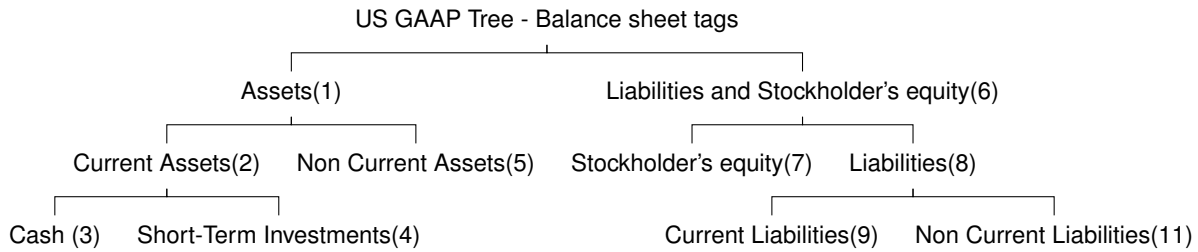
\begin{figure}[htb]
  \centering
  \begin{forest}
    for tree={
      parent anchor=south,
      child anchor=north,
      align=center,
      edge path={
        \noexpand\path[\forestoption{edge}]
          (!u.parent anchor) -- +(0,-3pt) -|
          (.child anchor)\forestoption{edge label};
      },
      l sep=4pt,
      tier/.wrap pgfmath arg={tier #1}{level()},
      font=\sffamily\small 
    }
    [US GAAP Tree - Balance sheet tags
      [Assets(1)
        [Current Assets(2)
          [Cash (3)]
          [Short-Term Investments(4) ]
        ]
        [Non Current Assets(5)]
      ]
      [Liabilities and Stockholder's equity(6)
        [Stockholder's equity(7)]
        [Liabilities(8)
          [Current Liabilities(9)]
          [Non Current Liabilities(11)]
        ]
      ]
    ]
  \end{forest}
  \caption{Simplified overview of the US-GAAP taxonomy tree structure. The full taxonomy is extensive and can be explored via the FASB website (\url{https://xbrlview.fasb.org/yeti/resources/yeti-gwt/Yeti.jsp}). Only a few top-level balance sheet tags are presented for illustration.}
  \label{fig:us_gaap_taxonomy_overview}
\end{figure}

A significant challenge in processing financial statements is handling missing data. The hierarchical nature of the US-GAAP taxonomy was leveraged to impute missing values. For instance, if a parent account (e.g., Total Assets) is reported but some of its constituent child accounts are missing, their values can sometimes be inferred based on the reported parent value and other reported sibling accounts. This imputation helps in creating a more complete financial picture for each report. Figure \ref{fig:us_gaap_taxonomy_overview} provides a simplified overview of the US-GAAP taxonomy structure.

\subsubsection{Feature Engineering and Quality Control}
Following imputation, a comprehensive set of 122 financial indicators was engineered. These indicators are designed to capture a wide array of financial signals relevant to fraud detection. To ensure data quality, a cutoff filtering process was applied: reports with excessive missing information, specifically those where less than 25\% of the 122 engineered features were present (i.e., non-zero and not NaN), were removed from the final dataset.

The 122 engineered features are categorized into five groups as detailed below.

\paragraph{Notation and Conventions:}
In the formulas presented, the subscript $t$ denotes the current fiscal quarter, and $t-1$ denotes the previous fiscal quarter.
\begin{itemize}
    \item $\Delta X = X_t - X_{t-1}$ represents the change in feature $X$ from the previous quarter to the current quarter.
    \item $\text{Avg}(X) = (X_t + X_{t-1})/2$ represents the average value of feature $X$ over the current and previous quarters.
    \item Values for financial tags that are missing in a report are treated as 0.
    \item \textbf{Safe Division (\texttt{safe\_divide(num, den)})}: If the denominator \texttt{den} is 0 or \texttt{NaN}, or if the numerator \texttt{num} is \texttt{NaN}, the result is 0. Otherwise, it is \texttt{num / den}.
    \item \textbf{Safe Summation (\texttt{safe\_sum(args...)})}: If any of the arguments \texttt{args} is \texttt{NaN} or 0, the result is 0. Otherwise, it is the sum of the arguments. This specific behavior is adopted for consistency in calculations.
\end{itemize}

\subsection{Basic Financial Numbers (43 features)}
These features are core financial metrics extracted directly from financial statements, standardized according to the US-GAAP taxonomy. The 43 basic financial numbers include:
\begin{enumerate}
    \item \fintag{AccountsPayableCurrentAndNoncurrent}
    \item \fintag{AccountsReceivableNetCurrent}
    \item \fintag{AccountsReceivableNetNoncurrent}
    \item \fintag{AccumulatedOtherComprehensiveIncomeLossNetOfTax}
    \item \fintag{AdditionalPaidInCapital}
    \item \fintag{AmortizationOfIntangibleAssets}
    \item \fintag{Assets}
    \item \fintag{AssetsCurrent}
    \item \fintag{CashCashEquivalentsAndShortTermInvestments}
    \item \fintag{CommonStockHeldBySubsidiary}
    \item \fintag{CommonStockValue}
    \item \fintag{CostOfRevenue}
    \item \fintag{DebtCurrent} (Short-Term Debt)
    \item \fintag{DeferredTaxAssetsDeferredIncome}
    \item \fintag{DeferredTaxLiabilitiesDeferredExpense}
    \item \fintag{DeferredTaxLiabilitiesTaxDeferredIncome}
    \item \fintag{DepreciationAndAmortization}
    \item \fintag{Goodwill}
    \item \fintag{GrossProfit}
    \item \fintag{IncomeLossFromContinuingOperations}
    \item \fintag{IntangibleAssetsNetIncludingGoodwill}
    \item \fintag{InterestAndDebtExpense}
    \item \fintag{InventoryNet}
    \item \fintag{Liabilities} (Total Liabilities)
    \item \fintag{LiabilitiesCurrent}
    \item \fintag{LongTermDebtCurrent}
    \item \fintag{LongTermDebtNoncurrent}
    \item \fintag{MinorityInterest}
    \item \fintag{NetCashProvidedByUsedInFinancingActivities}
    \item \fintag{NetCashProvidedByUsedInInvestingActivities}
    \item \fintag{NetCashProvidedByUsedInOperatingActivities}
    \item \fintag{NetIncomeLoss}
    \item \fintag{OperatingExpenses}
    \item \fintag{OperatingIncomeLoss}
    \item \fintag{PreferredStockValue}
    \item \fintag{PropertyPlantAndEquipmentNet}
    \item \fintag{ReceivableFromShareholdersOrAffiliatesForIssuanceOfCapitalStock}
    \item \fintag{RetainedEarningsAccumulatedDeficit}
    \item \fintag{Revenues}
    \item \fintag{SellingGeneralAndAdministrativeExpense}
    \item \fintag{TemporaryEquityCarryingAmountIncludingPortionAttributableToNoncontrollingInterests}
    \item \fintag{TreasuryStockValue}
    \item \fintag{UnearnedESOPShares}
\end{enumerate}

\subsection{Aggregated Measures (9 features)}
These features are composite values derived by summing related basic financial numbers to represent broader financial concepts.
\begin{enumerate}
    \item \finfeat{agg\_ACCOUNT\_RECEIVABLES}:\\ $\text{\fintag{AccountsReceivableNetCurrent}}_t + \text{\fintag{AccountsReceivableNetNoncurrent}}_t$
    \item \finfeat{agg\_LONG\_TERM\_DEBT}:\\ $\text{\fintag{LongTermDebtCurrent}}_t + \text{\fintag{LongTermDebtNoncurrent}}_t$
    \item \finfeat{agg\_EQUITY}:\\{safe\_sum} of: \\
    $|\text{\fintag{CommonStockValue}}_t|$, $|\text{\fintag{PreferredStockValue}}_t|$, $\text{\fintag{AdditionalPaidInCapital}}_t$, $\text{\fintag{RetainedEarningsAccumulatedDeficit}}_t$, $\text{\fintag{AccumulatedOtherComprehensiveIncomeLossNetOfTax}}_t$, $-\text{\fintag{TreasuryStockValue}}_t$, $-\text{\fintag{TemporaryEquityCarryingAmount...}}_t$, $-\text{\fintag{ReceivableFromShareholders...}}_t$, $-\text{\fintag{MinorityInterest}}_t$, $\text{\fintag{UnearnedESOPShares}}_t$, $\text{\fintag{CommonStockHeldBySubsidiary}}_t$
    \item \finfeat{agg\_TOTAL\_DEBT}:\\ $\text{\fintag{DebtCurrent}}_t + \text{\finfeat{agg\_LONG\_TERM\_DEBT}}_t$
    \item \finfeat{agg\_DEF\_TAX\_EXPENSE}:\\ $\text{\fintag{DeferredTaxLiabilitiesDeferredExpense}}_t - \text{\fintag{DeferredTaxAssetsDeferredIncome}}_t$
    \item \finfeat{agg\_ACCRUALS}:\\ $\text{\fintag{NetIncomeLoss}}_t - \text{\fintag{NetCashProvidedByUsedInOperatingActivities}}_t$
    \item \finfeat{agg\_EBIT}:\\ $\text{\fintag{Revenues}}_t - \text{\fintag{CostOfRevenue}}_t - \text{\fintag{OperatingExpenses}}_t$
    \item \finfeat{agg\_EBITDA}:\\ $\text{\finfeat{agg\_EBIT}}_t + \text{\fintag{DepreciationAndAmortization}}_t$
    \item \finfeat{agg\_NET\_CASH\_FLOW}:\\ $\text{\fintag{NetCashProvidedByUsedInOperatingActivities}}_t + \text{\fintag{NetCashProvidedByUsedInFinancingActivities}}_t + \text{\fintag{NetCashProvidedByUsedInInvestingActivities}}_t$
\end{enumerate}

\subsection{Change-based Measures (16 features)}
This category includes 16 features designed to capture temporal changes in financial accounts and performance. These features are:
\begin{enumerate}
     \item \finfeat{diff\_WC\_Accruals}:\\ $\Delta\text{\fintag{AssetsCurrent}} - \Delta\text{\fintag{LiabilitiesCurrent}} - \Delta\text{\fintag{CashCashEquivalentsAndShortTermInvestments}}$
    \item \finfeat{diff\_Inventories}:\\ $\text{safe\_divide}(\Delta\text{\fintag{InventoryNet}}, \text{Avg}(\text{\fintag{Assets}}))$
    \item \finfeat{diff\_Receivables}:\\ $\text{safe\_divide}(\Delta\text{\finfeat{agg\_ACCOUNT\_RECEIVABLES}}, \text{Avg}(\text{\fintag{Assets}}))$
    \item \finfeat{diff\_CashSales}:\\
    $\left( \text{safe\_divide}(\text{\fintag{Revenues}}_t, \text{Avg}(\text{\fintag{InventoryNet}})) + \text{safe\_divide}(\text{\fintag{Revenues}}_{t-1}, \text{Avg}(\text{\fintag{InventoryNet}})) \right) / 2 - \Delta\text{\finfeat{agg\_ACCOUNT\_RECEIVABLES}}$
    \item \finfeat{diff\_CashMargin}:\\ $\text{safe\_divide}(\texttt{safe\_sum}(\text{\fintag{CostOfRevenue}}_t, -\Delta\text{\fintag{InventoryNet}}, \Delta\text{\finfeat{agg\_ACCOUNT\_RECEIVABLES}}),\newline \text{\finfeat{diff\_CashSales}}_t)$
    \item \finfeat{diff\_DefTaxExpense}:\\ $\text{safe\_divide}(\Delta\text{\finfeat{agg\_DEF\_TAX\_EXPENSE}}, \text{\fintag{Assets}}_{t-1})$
    \item \finfeat{diff\_Earnings}:\\ $\text{safe\_divide}(\Delta\text{\fintag{NetIncomeLoss}}, \text{Avg}(\text{\fintag{Assets}}))$
    \item \finfeat{diff\_AverageAssets}:\\ $\text{Avg}(\text{\fintag{Assets}})$
    \item \finfeat{diff\_Revenues}:\\ $\Delta\text{\fintag{Revenues}}$
    \item \finfeat{diff\_Cash}:\\ $\Delta\text{\fintag{CashCashEquivalentsAndShortTermInvestments}}$
    \item \finfeat{diff\_EBIT}:\\ $\Delta\text{\finfeat{agg\_EBIT}}$
    \item \finfeat{diff\_EBITDA}:\\ $\Delta\text{\finfeat{agg\_EBITDA}}$
    \item \finfeat{diff\_NetCashFlow}:\\ $\Delta\text{\finfeat{agg\_NET\_CASH\_FLOW}}$
    \item \finfeat{diff\_Depreciation}:\\ $\Delta\text{\fintag{DepreciationAndAmortization}}$
    \item \finfeat{diff\_Assets}:\\ $\Delta\text{\fintag{Assets}}$
    \item \finfeat{diff\_Equity}:\\ $\Delta\text{\finfeat{agg\_EQUITY}}$
\end{enumerate}

\subsection{Ratio-based Measures (45 features)}

These features are financial ratios calculated to assess profitability, liquidity, solvency, efficiency, and market valuation. Table \ref{tab:financial_ratios} lists these ratios and their formulas.

\begin{longtable}{p{0.03\textwidth}p{0.35\textwidth}p{0.6\textwidth}}
\caption{List of 45 Ratio-based Measures} \label{tab:financial_ratios} \\
\toprule
\textbf{No.} & \textbf{Ratio Name (Feature ID)} & \textbf{Formula} \\
\midrule
\endfirsthead
\caption[]{List of 45 Ratio-based Measures (Continued)} \\
\toprule
\textbf{No.} & \textbf{Ratio Name (Feature ID)} & \textbf{Formula} \\
\midrule
\endhead
\bottomrule
\endfoot
1 & \finfeat{ratio\_GrossProfitMargin} & $\text{safe\_divide}(\text{\fintag{GrossProfit}}_t, \text{\fintag{Revenues}}_t)$ \\
2 & \finfeat{ratio\_OperatingMargin} & $\text{safe\_divide}(\text{\fintag{OperatingIncomeLoss}}_t, \text{\fintag{Revenues}}_t)$ \\
3 & \finfeat{ratio\_NetProfitMargin} & $\text{safe\_divide}(\text{\fintag{NetIncomeLoss}}_t, \text{\fintag{Revenues}}_t)$ \\
4 & \finfeat{ratio\_EBITMargin} & $\text{safe\_divide}(\text{\finfeat{agg\_EBIT}}_t, \text{\fintag{Revenues}}_t)$ \\
5 & \finfeat{ratio\_EBITDAMargin} & $\text{safe\_divide}(\text{\finfeat{agg\_EBITDA}}_t, \text{\fintag{Revenues}}_t)$ \\
6 & \finfeat{ratio\_CashFlowMargin} & $\text{safe\_divide}(\text{\fintag{NetCashProvidedByUsedInOperatingActivities}}_t,
\newline \text{\fintag{Revenues}}_t)$ \\
7 & \finfeat{ratio\_ReturnOnAssets} & $\text{safe\_divide}(\text{\fintag{NetIncomeLoss}}_t, \text{\fintag{Assets}}_t)$ \\
8 & \finfeat{ratio\_ReturnOnEquity} & $\text{safe\_divide}(\text{\fintag{NetIncomeLoss}}_t, \text{\finfeat{agg\_EQUITY}}_t)$ \\
9 & \finfeat{ratio\_CurrentRatio} & $\text{safe\_divide}(\text{\fintag{AssetsCurrent}}_t, \text{\fintag{LiabilitiesCurrent}}_t)$ \\
10 & \finfeat{ratio\_QuickRatio} & $\text{safe\_divide}(\text{\fintag{AssetsCurrent}}_t - \text{\fintag{InventoryNet}}_t, \text{\fintag{LiabilitiesCurrent}}_t)$ \\
11 & \finfeat{ratio\_CashRatio} & $\text{safe\_divide}(\text{\fintag{CashCashEquivalentsAndShortTermInvestments}}_t, \newline \text{\fintag{LiabilitiesCurrent}}_t)$ \\
12 & \finfeat{ratio\_WorkingCapitalToTotalAssets} & $\text{safe\_divide}(\text{\fintag{AssetsCurrent}}_t - \text{\fintag{LiabilitiesCurrent}}_t, \text{\fintag{Assets}}_t)$ \\
13 & \finfeat{ratio\_DebtToAssetsRatio} & $\text{safe\_divide}(\text{\finfeat{agg\_TOTAL\_DEBT}}_t, \text{\fintag{Assets}}_t)$ \\
14 & \finfeat{ratio\_DebtToEquityRatio} & $\text{safe\_divide}(\text{\finfeat{agg\_TOTAL\_DEBT}}_t, \text{\finfeat{agg\_EQUITY}}_t)$ \\
15 & \finfeat{ratio\_InterestCoverageRatio} & $\text{safe\_divide}(\text{\fintag{OperatingIncomeLoss}}_t, \text{\fintag{InterestAndDebtExpense}}_t)$ \\
16 & \finfeat{ratio\_TotalLiabilitiesToAssets} & $\text{safe\_divide}(\text{\fintag{Liabilities}}_t, \text{\fintag{Assets}}_t)$ \\
17 & \finfeat{ratio\_AssetTurnover} & $\text{safe\_divide}(\text{\fintag{Revenues}}_t, \text{Avg}(\text{\fintag{Assets}}))$ \\
18 & \finfeat{ratio\_FixedAssetTurnover} & $\text{safe\_divide}(\text{\fintag{Revenues}}_t, \text{Avg}(\text{\fintag{PropertyPlantAndEquipmentNet}}))$ \\
19 & \finfeat{ratio\_ReceivablesTurnover} & $\text{safe\_divide}(\text{\fintag{Revenues}}_t, \text{\finfeat{agg\_ACCOUNT\_RECEIVABLES}}_t)$ \\
20 & \finfeat{ratio\_InventoryTurnover} & $\text{safe\_divide}(\text{\fintag{CostOfRevenue}}_t, \text{Avg}(\text{\fintag{InventoryNet}}))$ \\
21 & \finfeat{ratio\_SalesTurnover} & $\text{safe\_divide}(\text{\fintag{Revenues}}_t, \text{Avg}(\text{\fintag{InventoryNet}}))$ \\
22 & \finfeat{ratio\_EquityMultiplier} & $\text{safe\_divide}(\text{\fintag{Assets}}_t, \text{\finfeat{agg\_EQUITY}}_t)$ \\
23 & \finfeat{ratio\_SGARatio} & $\text{safe\_divide}(\text{\fintag{SellingGeneralAndAdministrativeExpense}}_t, \text{\fintag{Revenues}}_t)$ \\
24 & \finfeat{ratio\_GoodwilltoAssets} & $\text{safe\_divide}(\text{\fintag{Goodwill}}_t, \text{\fintag{Assets}}_t)$ \\
25 & \finfeat{ratio\_CashFlowToDebtRatio} & $\text{safe\_divide}(\text{\fintag{NetCashProvidedByUsedInOperatingActivities}}_t, \newline \text{\finfeat{agg\_TOTAL\_DEBT}}_t)$ \\
26 & \finfeat{ratio\_CashFlowFinancingActivities} & $\text{safe\_divide}(\text{\fintag{NetCashProvidedByUsedInFinancingActivities}}_t, \newline \text{\finfeat{agg\_NET\_CASH\_FLOW}}_t)$ \\
27 & \finfeat{ratio\_CashFlowOperatingActivities} & $\text{safe\_divide}(\text{\fintag{NetCashProvidedByUsedInOperatingActivities}}_t, \newline\text{\finfeat{agg\_NET\_CASH\_FLOW}}_t)$ \\
28 & \finfeat{ratio\_EquityRatio} & $\text{safe\_divide}(\text{\finfeat{agg\_EQUITY}}_t, \text{\fintag{Assets}}_t)$ \\
29 & \finfeat{ratio\_CashFlowToCurrentLiabilities} & $\text{safe\_divide}(\text{\fintag{NetCashProvidedByUsedInOperatingActivities}}_t,\newline \text{\fintag{LiabilitiesCurrent}}_t)$ \\
30 & \finfeat{ratio\_CashFlowToRevenue} & $\text{safe\_divide}(\text{\fintag{NetCashProvidedByUsedInOperatingActivities}}_t,\newline \text{\fintag{Revenues}}_t)$ \\
31 & \finfeat{ratio\_CashFlowCoverageRatio} & $\text{safe\_divide}(\text{\fintag{NetCashProvidedByUsedInOperatingActivities}}_t, \newline\text{\finfeat{agg\_TOTAL\_DEBT}}_t)$ \\
32 & \finfeat{ratio\_NetWorkingCapital} & $\text{\fintag{AssetsCurrent}}_t - \text{\fintag{LiabilitiesCurrent}}_t$ \\
33 & \finfeat{ratio\_LongTermDebtToEquity} & $\text{safe\_divide}(\text{\finfeat{agg\_LONG\_TERM\_DEBT}}_t, \text{\finfeat{agg\_EQUITY}}_t)$ \\
34 & \finfeat{ratio\_DegreeOfFinancialLeverage} & $\text{safe\_divide}(\text{\fintag{Revenues}}_t, \text{\fintag{NetIncomeLoss}}_t)$ \\
35 & \finfeat{ratio\_InvestedCapitalRatio} & $\text{safe\_divide}(\text{\fintag{PropertyPlantAndEquipmentNet}}_t + \text{\fintag{InventoryNet}}_t, \text{\fintag{Assets}}_t)$ \\
36 & \finfeat{ratio\_CashToTotalAsset} & $\text{safe\_divide}(\text{\fintag{CashCashEquivalentsAndShortTermInvestments}}_t, \newline\text{\fintag{Assets}}_t)$ \\
37 & \finfeat{ratio\_DebtServiceCoverage} & $\text{safe\_divide}(\text{\fintag{OperatingIncomeLoss}}_t, \text{\finfeat{agg\_TOTAL\_DEBT}}_t)$ \\
38 & \finfeat{ratio\_FinancialLeverageIndex} & $\text{safe\_divide}(\text{\fintag{OperatingIncomeLoss}}_t, \text{\fintag{Assets}}_t)$ \\
39 & \finfeat{ratio\_TimesInterestEarnedRatio} & $\text{safe\_divide}(\text{\fintag{NetIncomeLoss}}_t$ + \\ $\text{\fintag{InterestAndDebtExpense}}_t, \text{\fintag{InterestAndDebtExpense}}_t)$ \\
40 & \finfeat{ratio\_CurrentAssetToRevenues} & $\text{safe\_divide}(\text{\fintag{AssetsCurrent}}_t, \text{\fintag{Revenues}}_t)$ \\
41 & \finfeat{ratio\_CurrentLiabilitiesToRevenues} & $\text{safe\_divide}(\text{\fintag{LiabilitiesCurrent}}_t, \text{\fintag{Revenues}}_t)$ \\
42 & \finfeat{ratio\_ShortTermDebtToRevenue} & $\text{safe\_divide}(\text{\fintag{DebtCurrent}}_t, \text{\fintag{Revenues}}_t)$ \\
43 & \finfeat{ratio\_IntangibleAssetToRevenue} & $\text{safe\_divide}(\text{\fintag{IntangibleAssetsNetIncludingGoodwill}}_t, \text{\fintag{Revenues}}_t)$ \\
44 & \finfeat{ratio\_LongtermLeverage} & $\text{safe\_divide}(\text{\finfeat{agg\_LONG\_TERM\_DEBT}}_t, \text{\fintag{Assets}}_t)$ \\
45 & \finfeat{ratio\_CFF} & $\text{safe\_divide}(\text{safe\_divide}
\newline(\text{\fintag{NetCashProvidedByUsedInFinancingActivities}}_t, \newline \text{\finfeat{agg\_NET\_CASH\_FLOW}}_t), \text{Avg}(\text{\fintag{Assets}}))$ \\
\end{longtable}

\subsection{Beneish M-Score Indicators (9 features)}
These features are components of the Beneish M-Score model, designed to detect earnings manipulation. The individual indicators are:
\begin{enumerate}
    \item \finfeat{Beneish\_DSRI} (Days' Sales in Receivables Index): \\
    $\text{safe\_divide}\left( \frac{\text{\finfeat{agg\_ACCOUNT\_RECEIVABLES}}_t}{\text{\fintag{Revenues}}_t} , \frac{\text{\finfeat{agg\_ACCOUNT\_RECEIVABLES}}_{t-1}}{\text{\fintag{Revenues}}_{t-1}} \right)$
    \item \finfeat{Beneish\_GMI} (Gross Margin Index): \\
    Let $GM_t = \text{safe\_divide}(\text{\fintag{Revenues}}_t - \text{\fintag{CostOfRevenue}}_t, \text{\fintag{Revenues}}_t)$. Then $\text{\finfeat{Beneish\_GMI}} = \text{safe\_divide}(GM_{t-1}, GM_t)$.
    \item \finfeat{Beneish\_AQI} (Asset Quality Index): \\
    Let $NCA_t = \text{\fintag{Assets}}_t - \text{\fintag{AssetsCurrent}}_t - \text{\fintag{PropertyPlantAndEquipmentNet}}_t$. \\
    Let $AQ_t = \text{safe\_divide}(NCA_t, \text{\fintag{Assets}}_t)$.\\ Then $\text{\finfeat{Beneish\_AQI}} = \text{safe\_divide}(AQ_t, AQ_{t-1})$.
    \item \finfeat{Beneish\_SGI} (Sales Growth Index): $\text{safe\_divide}(\text{\fintag{Revenues}}_t, \text{\fintag{Revenues}}_{t-1})$
    \item \finfeat{Beneish\_DEPI} (Depreciation Index): \\
    Let $DepRate_t = \text{safe\_divide}(\text{\fintag{DepreciationAndAmortization}}_t, \text{\fintag{DepreciationAndAmortization}}_t + \text{\fintag{PropertyPlantAndEquipmentNet}}_t)$. \\
    Then $\text{\finfeat{Beneish\_DEPI}} = \text{safe\_divide}(DepRate_{t-1}, DepRate_t)$.
    \item \finfeat{Beneish\_SGAI} (SG\&A Index): \\
    $\text{safe\_divide}\left( \frac{\text{\fintag{SellingGeneralAndAdministrativeExpense}}_t}{\text{\fintag{Revenues}}_t} , \frac{\text{\fintag{SellingGeneralAndAdministrativeExpense}}_{t-1}}{\text{\fintag{Revenues}}_{t-1}} \right)$
    \item \finfeat{Beneish\_ACCRUALS} (Total Accruals to Total Assets): $\text{safe\_divide}(\text{\finfeat{agg\_ACCRUALS}}_t, \text{\fintag{Assets}}_t)$
    \item \finfeat{Beneish\_LVGI} (Leverage Index): \\
    Let $Lev_t = \text{safe\_divide}(\text{\finfeat{agg\_LONG\_TERM\_DEBT}}_t + \text{\fintag{DebtCurrent}}_t, \text{\fintag{Assets}}_t)$. Then $\text{\finfeat{Beneish\_LVGI}} = \text{safe\_divide}(Lev_t, Lev_{t-1})$.
    \item \finfeat{Beneish\_PROBM} (Beneish M-Score): This is the M-Score itself, calculated using the indicators above.
\end{enumerate}
The \finfeat{Beneish\_PROBM} (M-Score) is calculated as:
\begin{align*}
\text{M-Score} = -4.84 &+ 0.920 \times \text{DSRI} \\
                      &+ 0.528 \times \text{GMI} \\
                      &+ 0.404 \times \text{AQI} \\
                      &+ 0.892 \times \text{SGI} \\
                      &+ 0.155 \times \text{DEPI} \\
                      &- 0.172 \times \text{SGAI} \\
                      &+ 4.679 \times \text{ACCRUALS\_val} \\
                      &- 0.327 \times \text{LVGI\_val}
\end{align*}
Where DSRI, GMI, AQI, SGI, DEPI, SGAI are the values of the correspondingly named Beneish indicators (\finfeat{Beneish\_DSRI}, \finfeat{Beneish\_GMI}, etc.). ACCRUALS\_val is the value of the \finfeat{Beneish\_ACCRUALS} feature, and LVGI\_val is the value of the \finfeat{Beneish\_LVGI} feature.

\subsection{Dataset Statistics}
Descriptive statistics for the number of engineered features per firm-quarter report and reports per company (CIK) are presented in Table \ref{tab:feature_count_stats}. The "Number of Extended Features" refers to the count of non-zero values among all potentially derived financial features for a given report prior to final selection for the model, indicating the richness of available data per report. "Important Tags" count refers to a predefined subset of raw US-GAAP tags deemed critical.

\begin{table}[htb]
\centering
\caption{Descriptive Statistics for Feature Counts per Firm-Quarter Report}
\label{tab:feature_count_stats}
\resizebox{\textwidth}{!}{%
\begin{tabular}{@{}lrrrrrr@{}}
\toprule
Feature &Count & Mean   & Median & Min & Max & Std Dev \\ \midrule
Base numerical numbers (\textit{n\_important\_tags})\textsuperscript{***} &43& 20.60 & 21.0 & 11.0 & 33.0 & 4.01 \\
Aggregated Measures (\textit{n\_aggregates}) &9& 1.35   & 1.0    & 0.0 & 7.0 & 1.36    \\
Change-base Features (\textit{n\_diff\_features})\textsuperscript{*} &16& 11.21  & 11.0   & 8.0 & 15.0 & 0.77   \\
Ratio-based Measures (\textit{n\_ratios}) &45& 19.72  & 21.0   & 1.0 & 39.0 & 6.88    \\
Beneish Features (\textit{n\_benish\_features}) &9& 2.62   & 2.0    & 0.0 & 9.0 & 1.57    \\
\midrule
\textbf{All Features (\textit{n\_features})} & \textbf{122} & \textbf{55.50} & \textbf{56} & \textbf{20} & \textbf{93} & \textbf{12.35}\\
\bottomrule
\end{tabular}%
}
\caption*{\textsuperscript{*}Refers to a broader set of calculated differential values tracked during preprocessing, from which the 8 change-based measures listed earlier are a specific subset.
\textsuperscript{***}Count of non-zero values for a predefined subset of 43 important US-GAAP tags.}

\subsection{Dechow Model Features}

The Dechow model~\cite{dechow_predicting_2011}, as implemented in our study, incorporates the following features, based on the implementation from \url{https://github.com/jdonadio/FSFraud}:

\begin{itemize}
    \item \textbf{DECHOW\_RSST\_ACCRUALS (RSST Accrual)}: This measures discretionary accruals based on the Reverse-Salomon-Teoh (RSST) model. It is calculated as:

    $$
    \text{RSST Accruals} = \frac{\Delta WC + \Delta NCO + \Delta FIN}{\text{Average Total Assets}}
    $$
    where:
    \begin{itemize}
        \item $\Delta WC$ is the change in working capital, calculated as:
        $$
        \Delta WC = [\Delta \text{Current Assets} - \Delta \text{Cash and Short-term Investments}] - [\Delta \text{Current Liabilities} - \Delta \text{Short-term Debt}]
        $$
        \item $\Delta NCO$ is the change in net non-current operating assets, calculated as:
        $$
        \Delta NCO = \Delta [\text{Total Assets} - \text{Current Assets} - \text{Investments}] - \Delta [\text{Total Liabilities} - \text{Current Liabilities} - \text{Long-term Debt}]
        $$
        \item $\Delta FIN$ is the change in financing activities, calculated as:
        $$
        \Delta FIN = \Delta \text{Short-term Investments} - \Delta [\text{Long-term Debt} + \text{Short-term Debt} + \text{Preferred Stock}]
        $$
    \end{itemize}

    \item \textbf{DECHOW\_CH\_REC (Change in Receivables)}: This is the change in accounts receivable scaled by total assets, calculated as:
    $$
    \text{Change in Receivables} = \frac{\Delta \text{Accounts Receivable}}{\text{Average Total Assets}}
    $$

    \item \textbf{DECHOW\_CH\_INV (Change in Inventory)}: This is the change in inventory scaled by total assets, calculated as:
    $$
    \text{Change in Inventory} = \frac{\Delta \text{Inventory}}{\text{Average Total Assets}}
    $$

    \item \textbf{DECHOW\_SOFT\_ASSETS (Soft Assets)}: This is the ratio of intangible assets and goodwill to total assets, indicating the proportion of 'soft' or less tangible assets:
    $$
    \text{Soft Assets} = \frac{\text{Intangible Assets} + \text{Goodwill}}{\text{Total Assets}}
    $$

    \textbf{DECHOW\_CH\_CASHSALES (Change in Cash Sales)}:  This is the change in cash sales, where cash sales are calculated as sales minus changes in accounts receivable:
    $$
    \text{Change in Cash Sales} = \text{Sales} - \Delta \text{Accounts Receivable}
    $$

    \item \textbf{DECHOW\_CH\_ROA (Change in Return on Assets)}: This is the change in Return on Assets, indicating the trend in a company's profitability relative to its assets:
    $$
    \text{Change in ROA} = \frac{\text{Earnings}_t}{\text{Average Total Assets}_t} - \frac{\text{Earnings}_{t-1}}{\text{Average Total Assets}_{t-1}}
    $$

    \item \textbf{DECHOW\_ISSUANCE (Issuance)}: This feature indicates whether the company issued new shares or new debt in the current period. It is a binary variable:1, \text{if equity or debt issuance occurred}, 0 otherwise

\end{itemize}

\end{table}

%% file: appendixes/appendix_2_mda_dataset_details.tex
\section{Appendix B. Details on MD\&A Reports Dataset}
\label{appendix:mda_details}

This appendix provides further details on the Management Discussion and Analysis (MD\&A) sections used in our study. We elaborate on the characteristics of the raw MD\&A data, the summarization process employed, the resultant Synthetically Summarized MD\&A (SMD\&A) dataset, and the associated costs for this data processing step.

\subsection{Raw MD\&A Sections}
The raw MD\&A sections were extracted using the API \url{https://sec-api.io/}. We subscribed for monthly plan which costs \textbf{\$55}(\url{https://sec-api.io/pricing} and which is enough to download all the available quarterly MD\&A sections. Specifically, we query the endpoint \url{https://sec-api.io/docs/sec-filings-item-extraction-api} to get the item \texttt{part1item2} of \texttt{Form-10Q} which refers exactly to the desired MD\&A sections. These sections are typically lengthy and contain a mix of textual narratives, financial figures, and sometimes tables, often embedded within HTML structures.

\subsubsection{Tokens' count Distribution of Raw MD\&A}
\begin{figure}[htbp]
    \centering
    \includegraphics[width=\linewidth]{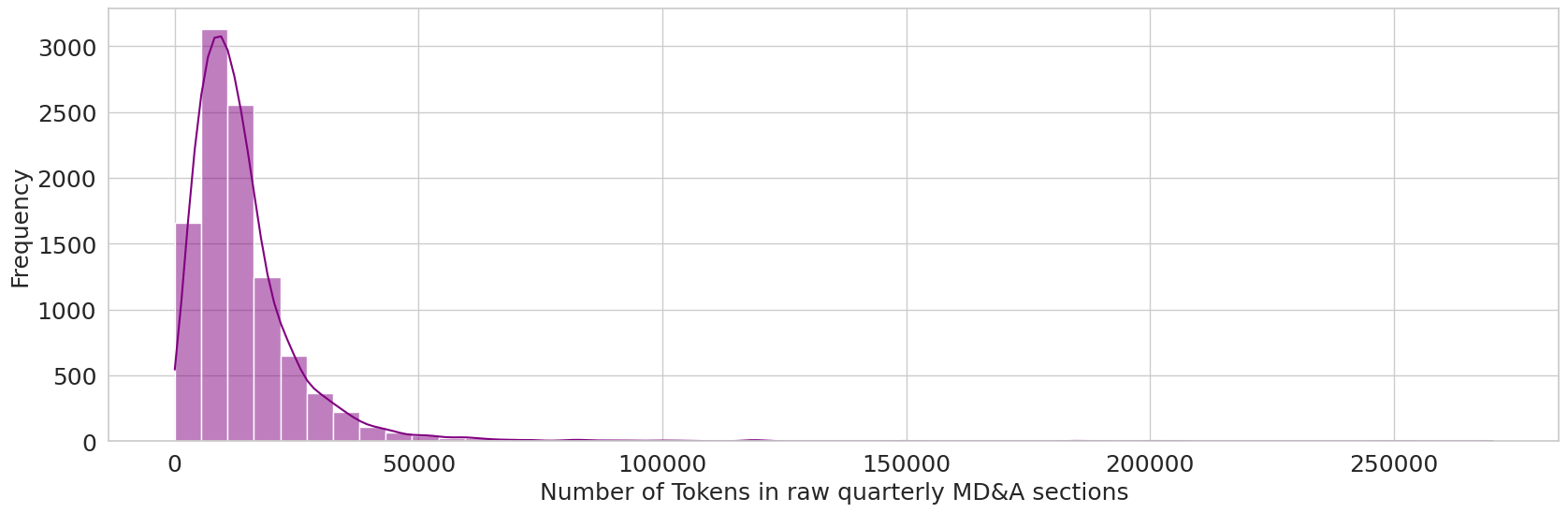}
    \caption{Distribution of token counts in raw quarterly MD\&A sections. The x-axis represents the number of tokens, and the y-axis represents the frequency.}
    \label{fig:raw_mda_token_dist}
\end{figure}

The distribution of token counts for the raw MD\&A sections is depicted in Figure \ref{fig:raw_mda_token_dist}. These sections exhibit considerable variability in length.

Table \ref{tab:raw_mda_stats} presents the descriptive statistics for the token counts of the raw MD\&A sections in our dataset.

\begin{table}[htbp]
  \centering
  \caption{Descriptive statistics for token counts of raw MD\&A sections.}
  \label{tab:raw_mda_stats}
  \begin{tabular}{lr}
    \toprule
    Statistic          & Value         \\
    \midrule
    Count              & 10,159        \\
    Mean               & 14,021     \\
    Standard Deviation & 12,580     \\
    Minimum            & 18        \\
    25th Percentile    & 7,049         \\
    50th Percentile    & 11,370        \\
    75th Percentile    & 17,180        \\
    Maximum            & 270,346      \\
    \bottomrule
  \end{tabular}
\end{table}

\subsubsection{Sample Raw MD\&A}
Below is an excerpt from a sample raw MD\&A section, illustrating its typical structure and content. Note the presence of HTML entities.

\begin{tcolorbox}[enhanced,breakable, colback=gray!5!white, colframe=gray!75!black, title=Sample Raw MD\&A Excerpt, fonttitle=\bfseries, boxsep=2pt,         %
  left=2pt,
  right=2pt]
\begin{verbatim}
Item 2. Management's Discussion and Analysis of Financial Condition and
 Results of Operations

CAUTIONARY STATEMENT RELATING TO THE SAFE HARBOR PROVISIONS OF THE PRIVATE
SECURITIES LITIGATION REFORM ACT OF 1995

This Quarterly Report contains forward-looking statements as that term is
defined in the federal securities laws. The events described in
forward-looking statements contained in this Quarterly Report may not occur.
Generally, these statements relate to our business plans or strategies,
projected or anticipated benefits or other consequences of our plans or
strategies, financing plans, projected or anticipated benefits from
acquisitions that we may make, or projections involving anticipated
revenues, earnings or other aspects of our operating results or financial
position, and the outcome of any contingencies. Any such forward-looking
statements are based on current expectations, estimates and projections of
management. We intend for these forward-looking statements to be covered by
the safe-harbor provisions for forward-looking statements. Words such as
“may,” “will,” “expect,”
“believe,” “anticipate,” “project,”
“plan,” “intend,” “estimate,”
and “continue,” and their opposites and similar expressions
are intended to identify forward-looking statements. We caution you that
these statements are not guarantees of future performance or events and are
subject to a number of uncertainties, risks and other influences, many of
which are beyond our control that may influence the accuracy of the
statements and the projections upon which the statements are based. Factors
that could cause actual results to differ materially from those set forth or
implied by any forward-looking statement include, but are not limited to,
our ability to remain competitive with competitors, risks associated with
the generic product industry, dependence on a limited number of suppliers,
risks associated with healthcare reform and reductions in reimbursement
rates, difficulty in predicting revenue stream and gross profit, industry
and market changes, the effect of fluctuations in operating results on the
trading price of our common stock, inventory levels, reliance on outside
manufacturers, risks of incurring uninsured environmental and other
industry specific liabilities, governmental approvals and regulations, risks
associated with hazardous materials, potential violations of government
regulations, product liability claims, reliance on Chinese suppliers,
potential changes to Chinese laws and regulations, potential changes to
laws governing our relationships in India , fluctuations in foreign
currency exchange rates, tax assessments, changes in tax rules, global
economic risks, risk of unsuccessful acquisitions, effect of acquisitions
on earnings, indemnification liabilities, terrorist activities, reliance on
key executives, litigation risks, volatility of the market price of our
common stock, changes to estimates, judgments and assumptions used in
preparing financial statements, failure to maintain effective internal
controls, compliance with changing regulations , as well as other risks and
uncertainties discussed in our reports filed with the Securities and
Exchange Commission, including, but not limited to, our Annual Report on
Form 10-K for the fiscal year ended June 30, 2011 and other filings.
Copies of these filings are available at www.sec.gov.

Any one or more of these uncertainties, risks and other influences could
materially affect our results of operations and whether forward-looking
statements made by us ultimately prove to be accurate. Our actual results,
performance and achievements could differ materially from those expressed
or implied in these forward-looking statements. We undertake no obligation
to publicly update or revise any forward-looking statements, whether from
new information, future events or otherwise.

NOTE REGARDING DOLLAR AMOUNTS

In this quarterly report, all dollar amounts are expressed in thousands,
except for per-share amounts.

The following Management’s Discussion and Analysis of Financial Condition
and Results of Operations (MD&A) is intended to provide the readers of
our financial statements with a narrative discussion about our business.
The MD&A is provided as a supplement to and should be read in
conjunction with our financial statements and the accompanying notes.

Executive Summary

We are reporting net sales of $212,024 for the six months ended
December 31, 2011, which represents a 22.3% increase from the $173,343
reported in the comparable prior period. Gross profit for the six months
ended December 31, 2011 was $39,163 and our gross margin was 18.5% as
compared to gross profit of $26,410 and gross margin of 15.2% in the
comparable prior period. Our selling, general and administrative costs
(SG&A) for the six months ended December 31, 2011 increased $6,073 to
$27,097 from the amount we reported in the prior period. Our net income
increased to $7,621, or $0.29 per diluted share, compared to net income of
$1,628, or $0.06 per diluted share in the prior period.

Our financial position as of December 31, 2011 remains strong, as we had
cash and cash equivalents and short-term investments of $28,700, working
capital of $115,838 and shareholders’ equity of $161,571.

Our business is separated into three principal segments: Health Sciences,
Specialty Chemicals and Agricultural Protection Products. The Health
Sciences segment is our largest segment in terms of both sales and gross
profits. Products that fall within this segment include pharmaceutical
intermediates, APIs, finished dosage form generic drugs and nutraceutical
products.

We typically partner with both customers and suppliers years in advance of a
drug coming off patent to provide the generic equivalent. We believe we have
a pipeline of new APIs poised to reach commercial levels over the coming
years as the patents on existing drugs expire, both in the United States
and in Europe. In addition, we continue to explore opportunities to provide
a second-source option for existing generic drugs with approved abbreviated
new drug applications (ANDAs). The opportunities that we are looking for are
to supply the APIs for the more mature generic drugs where pricing has
stabilized following the dramatic decreases in price that these drugs
experienced after coming off patent. As is the case in the generic
industry, the entrance into the market of other generic competition
generally has a negative impact on the pricing of the affected products. By
leveraging our worldwide sourcing, quality assurance and regulatory
capabilities, we believe we can be an alternative economical, second-source
provider of existing APIs to generic drug companies. On December 31, 2010,
we acquired certain assets of Rising Pharmaceuticals, Inc. (“Rising”)
. We believe that the acquisition of Rising will establish another platform
for our growth in our Health Sciences business by the expansion of our
finished dosage form product offerings from both foreign and domestic
facilities as well as complementing our core strength of sourcing active
pharmaceutical ingredients. The addition of Rising provides Aceto with a
presence as a developer and marketer of our own brand of generic
pharmaceuticals, the Rising brand.

According to an IMS Health press release on May 18, 2011, "global spending
for medicines will reach nearly $1.1 trillion by 2015, reflecting a
slowing compound annual rate of growth of 3 – 6 percent over the next
five years. This compares with 6.2 percent annual growth over the past five
years. Lower levels of spending growth for medicines in the U.S., the
ongoing impact of patent expiries in developed markets, continuing strong
demand in pharmerging markets and policy-driven changes in several
countries are among the key factors that will influence future growth,
according to IMS Institutes new study, The Global Use Of Medicines Outlook
Through 2015".

Aceto supplies the raw materials used in the production of nutritional and
packaged dietary supplements, including vitamins, amino acids, iron
compounds and biochemicals used in pharmaceutical and nutritional
preparations. Aceto’s identification of a change in the attitudes of
Europeans towards nutritional products led to the decision to globalize
this business and create an operating company to focus on it, Aceto Health
Ingredients GmbH, headquartered in Germany. This globally structured
business has become the model for all of our business segments, providing
international reach and perspective for our customers.

The Specialty Chemicals segment is a supplier to the many different
industries that require outstanding performance from chemical raw materials
and additives. Specialty Chemicals include a variety of chemicals which make
plastics, surface coatings, textiles, fuels and lubricants perform to their
designed capabilities. Dye and pigment intermediates are used in the
color-producing industries such as textiles, inks, paper, and coatings.
Many of our raw materials are also used in high-tech products like high-end
electronic parts (circuit boards and computer chips) and binders for
specialized rocket fuels. We continue to respond to the changing needs of
our customers in the color producing industry by taking our resources and
knowledge downstream as a supplier of select organic pigments. In addition,
Aceto is a leader in the supply of diazos and couplers to the paper, film
and electronics industries.

According to a December 15, 2011 Federal Reserve Statistical Release, in
the third quarter of calendar year 2011, the index for consumer durables,
which impacts the Specialty Chemicals segment, grew at an annual rate of
11.4%.
\end{verbatim}
Item 3. Quantitative and Qualitative Disclosures About Market Risk

Market risk is the risk of loss arising from adverse changes in market
rates and prices, such as interest rates, foreign currency exchange rates
and commodity prices. Our primary exposure to market risk is interest rate
risk and foreign currency exchange rate risk.

We do not use derivative financial instruments for trading or speculative
purposes. We do not use any derivative contracts to hedge foreign currency
or interest rate exposure. We seek to minimize foreign currency exchange
rate risk through management of our current assets and liabilities which are
denominated in foreign currencies. The principal foreign currencies to
which we are exposed are the Euro, Indian Rupee and Chinese Yuan.

Interest Rate Risk. Our interest expense is sensitive to changes in the
general level of interest rates, as substantially all of our borrowings are
at variable rates. Our exposure to interest rate risk relates primarily to
our Amended and Restated Credit Agreement, as amended (the “Credit
Agreement”). As of December 31, 2011, we had $0 outstanding under our
revolving credit facility and $5,000 outstanding under our term loan
facility. Each of these facilities bears interest at a variable rate based
on LIBOR or the Base Rate (as defined in the Credit Agreement). A 100
basis point increase in interest rates would increase our interest expense
by \$50 annually.

Foreign Currency Exchange Rate Risk. We are exposed to foreign currency
exchange rate risk related to our purchases and sales denominated in foreign
currencies. Our foreign currency exchange rate risk is inherent in the sales
and expenses of our foreign subsidiaries, which are denominated in their
respective local currencies. The financial statements of our foreign
subsidiaries are translated into U.S. dollars at exchange rates in effect
at the balance sheet date for assets and liabilities and average exchange
rates during the period for revenues and expenses. As a result, changes in
exchange rates may affect the reported value of our foreign assets,
liabilities, revenues and expenses, and could result in foreign currency
translation gains or losses in our consolidated statements of operations.
A hypothetical 10
and Chinese Yuan would not have a material effect on our results of
operations.
\end{tcolorbox}

\subsection{MD\&A Summarization Process}
To make the extensive textual data from MD\&A sections more manageable for LLM processing while retaining core financial insights, we employed a summarization strategy using Qwen3 -32B model. This model was chosen for its long-context capabilities and efficiency.

\subsubsection{System Prompt for Summarization}
The following system prompt was used to guide the Qwen3 -32B model in summarizing the raw MD\&A sections. The `{quarter\_info}` placeholder was dynamically filled with the specific quarter and year of the report (e.g., "Q4 2023").

\begin{tcolorbox}[colback=blue!5!white, colframe=blue!75!black, title=System Prompt for MD\&A Summarization, breakable]
 \small
 \begin{verbatim}

You are a highly skilled financial analyst with deep expertise in summarizing
corporate disclosures.\\
You will be provided with the 'Management's Discussion and Analysis' (MD\&A)
section of a financial report for {quarter\_info}.\\
Your task is to summarize it following the instructions below:

Extract and present the **distinct, and factual insights**, along with subjective
statements, management commentary, and qualitative explanations, organized into
the following sections:

---

** 1. Strategic Priorities and Initiatives**\\
   – Summarize key strategies, corporate objectives, growth plans,
     restructuring efforts, and major initiatives discussed by management.\\
   – Capture significant strategic shifts, operational transformations,
     ambitious targets, or business model changes.\\
   – Highlight subjective language, including optimistic tone, vague
     descriptions of progress, or assertions lacking clear supporting evidence.

** 2. Operational and Segment Performance**\\
   – Summarize operational results and segment-level performance, including
     production metrics, KPIs, challenges, and improvements.\\
   – Pay special attention to:\\
     – Unexplained variances in performance.\\
     – Misalignment between narrative explanations and operational metrics.\\
     – Subjective, vague, or generic explanations (e.g., “seasonality,”
       “market dynamics,” “operational excellence”) without adequate
       quantification.\\
     – Unusual operational trends, sales fluctuations, production shifts,
       or inventory movements.

** 3. Financial Results and Key Trends**\\
   – Capture **all financial metrics**, including revenue, profitability,
     margins, cost drivers, liquidity trends, capital structure, and debt
     along with financial ratios.\\
   – If the metrics are presented in tables, rewrite them in the section
     “Important Figures and Tables” instead of here.\\
   – Also Include commentary on:\\
     – Revenue recognition patterns or timing shifts.\\
     – Significant margin changes or cost structure shifts.\\
     – Increases in accounts receivable, inventory, or other working capital
       components relative to sales without clear justification.\\
     – Use of non-recurring items, adjustments, or changes in estimates that
       materially impact results.\\
     – Use of non-recurring items, adjustments, or changes in estimates that
       materially impact results.\\
     – Subjective rationalizations for financial outcomes (e.g., references to
       “strong demand” or “improved efficiencies”) that lack numeric validation.

** 4. Identified Risks and Uncertainties**\\
   – Summarize disclosed risks, including operational, supply chain, regulatory,
     competitive, legal, and macroeconomic risks.\\
   – Capture both concrete risks and:\\
     – Subjective assessments of risk severity.\\
     – Ambiguous or hedged language (e.g., “may,” “could,” “uncertain”).\\
     – Shifts in tone, emphasis, or presentation of risks compared to prior periods.

** 5. Forward-Looking Statements and Guidance**\\
   – Capture management’s expectations, forecasts, assumptions, and outlook
     for future periods.\\
   – Highlight:\\
     – Changes in guidance or underlying assumptions.\\
     – Optimistic tone, hedging, or caveats (e.g., “expects,” “believes,” “aims”).\\
     – Whether forward-looking statements are grounded in quantifiable drivers
       or rely mainly on qualitative assertions.

** 6. Significant Changes, Events, or Developments**\\
   – Summarize material recent or upcoming events affecting the business, such as
     mergers, acquisitions, divestitures, leadership changes, legal proceedings,
     regulatory actions, or external shocks.\\
   – Note how management frames these events—whether impacts are clearly
     quantified or described with vague or qualitative language.

** 7. Important Figures and Tables**\\
   – Extract key figures, tables, or financial data that are critical to
     understanding the MD\&A.\\
   – For each table:\\
        Recreate the exact table content in clean markdown format preceded by
        the table title.

** 8. Management Explanations and Justifications**\\
   – Capture how management explains or justifies operational and financial
     results, risks, or variances.\\
   – Pay attention to:\\
     – Vague, broad, or overly generic justifications.\\
     – Repetitive use of boilerplate terms (e.g., “market conditions,”
       “execution excellence”) without specific detail.\\
     – Narratives that shift accountability to external factors or uncontrollable
       circumstances without precise quantification.

** 9. Accounting Estimates, Judgments, and Policy Changes**\\
   – Summarize any disclosures related to:\\
     – Changes in accounting policies, methodologies, or estimates.\\
     – Adjustments to key assumptions (e.g., impairments, allowances, revenue recognition).\\
     – Areas where significant management judgment materially affects reported results.\\
   – Note whether explanations are clear, detailed, vague, hedged, or superficial.

** 10. Capital Allocation and Liquidity Management**\\
    – Summarize commentary on:\\
      – Cash management strategies, liquidity preservation, and debt management.\\
      – Capital expenditures, share repurchases, dividend policies, and
        financing activities.\\
    – Highlight any:\\
      – Indications of liquidity stress.\\
      – Mismatches between optimistic narratives and defensive liquidity
        actions (e.g., drawing on credit lines despite claimed strong financial performance).

** 11. Legal, Regulatory, and Compliance Matters**\\
    – Summarize discussions related to:\\
      – Ongoing or pending litigation.\\
      – Regulatory investigations or changes.\\
      – Compliance risks, including ESG-related disclosures that have
        material financial implications.\\
    – Note whether these issues are presented transparently, minimized, or
      framed with ambiguous language.

---

**Formatting Instructions:**\\
– Use section headers exactly as written above ie with the numbers and titles.\\
– Present each point as a bullet (–) under the appropriate section.\\
– Include both objective data and subjective commentary.\\
– Explicitly note subjective explanations, optimistic framing, hedging, or
  vague descriptions wherever they appear.\\
– Be precise and factual but the summary should be detailed\\
– Avoid redundancy; each bullet must convey a distinct, meaningful insight.

**Critical Constraint:**\\
– Base the summary **strictly on the content explicitly stated in the MD\&A.**\\
– Do not include any external knowledge, assumptions, interpretations, or
  analysis beyond the document provided.\\
– Only output the summary — do not include any commentary, explanations, or
  meta-text.
\end{verbatim}
\end{tcolorbox}

\subsection{Summarized MD\&A (SMD\&A) Sections}
The summarization process resulted in the SMD\&A dataset, consisting of condensed versions of the original MD\&A narratives.

\subsubsection{Tokens' count Distribution of SMD\&A}
\begin{figure}[htbp]
    \centering
    \includegraphics[width=\linewidth]{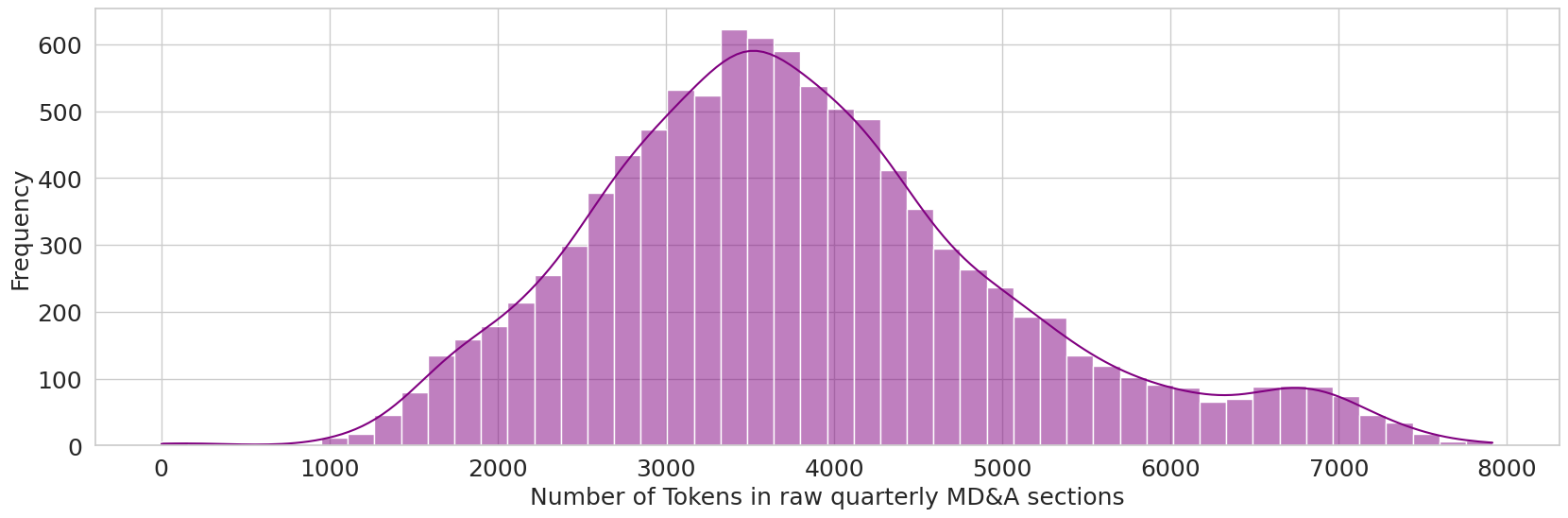}
    \caption{Distribution of token counts in Summarized MD\&A (SMD\&A) sections. The x-axis represents the number of tokens, and the y-axis represents the frequency.}
    \label{fig:smda_token_dist}
\end{figure}

Figure \ref{fig:smda_token_dist} illustrates the token count distribution for the SMD\&A sections. As intended, these summaries are substantially shorter than the raw MD\&A sections.

Table \ref{tab:smda_stats} provides descriptive statistics for the token counts of the SMD\&A sections.

\begin{table}[htbp]
  \centering
  \caption{Descriptive statistics for token counts of SMD\&A sections.}
  \begin{tabular}{lr}
    \toprule
    Statistic          & Value      \\
    \midrule
    Count              & 10,159     \\
    Mean               & 3,807  \\
    Standard Deviation & 1,250   \\
    Minimum            & 1         \\
    25th Percentile    & 2,960      \\
    50th Percentile    & 3,666      \\
    75th Percentile    & 4,476      \\
    Maximum            & 7,914      \\
    \bottomrule
    \label{tab:smda_stats}

  \end{tabular}

\end{table}
\newpage
\subsubsection{Sample SMD\&A}

An excerpt from a sample SMD\&A is provided below. This illustrates the more structured and condensed format achieved through the summarization process.
\begin{tcolorbox}[enhanced, breakable, colback=green!5!white, colframe=green!65!black, title=Sample SSMD\&A Excerpt, fonttitle=\bfseries, breakable, boxsep=2pt,         %
  left=2pt,
  right=2pt]
  \small
\begin{verbatim}
**1. Strategic Priorities and Initiatives**
- The acquisition of St. Jude Medical, Inc. (St. Jude Medical) was completed
  on January 4, 2017, to expand Abbott’s presence in the cardiovascular
  and neuromodulation markets.
- Abbott is reshaping its business portfolio through divestitures, including
  the sale of its vision care business (AMO) to Johnson & Johnson for
  $4.325 billion in cash.
- The company is pursuing the acquisition of Alere Inc. (Alere) to expand
  its global diagnostics presence. The purchase price was reduced from
  $56.00 to $51.00 per share in April 2017, with the acquisition expected
  to close by the end of Q3 2017, subject to regulatory approvals.
- Abbott is implementing cost improvement initiatives across various
  functions and businesses, partially offsetting increased expenses from
  the St. Jude Medical acquisition.
- The company is investing in research and development (R&D), with R&D
  expenses increasing significantly due to the integration of the St. Jude
  Medical business.
- Abbott has a share repurchase program authorized by its board in 2014
  for up to $3.0 billion, in addition to $512 million remaining from a
  prior program.
- Abbott increased its quarterly dividend by approximately 2% in 2017
  compared to 2016, indicating a focus on shareholder returns.

**2. Operational and Segment Performance**
- Net sales for the Cardiovascular and Neuromodulation Products segment
  increased by 198.5% in the first six months of 2017 due to the St. Jude
  Medical acquisition. Excluding the acquisition and foreign exchange,
  sales in this segment decreased by 1.5% as lower coronary stent sales
  and a favorable 2016 royalty agreement resolution were partially offset
  by higher Structural Heart and endovascular sales.
- Sales in the Established Pharmaceutical Products segment increased by
  5.5% in the first six months of 2017. Excluding foreign exchange, sales
  in Key Emerging Markets increased 8.2% in the first half of 2017, driven
  by growth in Russia, China, and Latin America, partially offset by the
  impact of a new GST system in India.
- Nutritional Products sales decreased slightly by 0.3% in the first six
  months of 2017, with International Pediatric Nutritionals declining by
  8.0%. Challenging conditions in the Chinese infant formula market
  continued to impact international performance.
- U.S. Pediatric Nutritionals increased by 7.7%, driven by momentum from
  recently launched infant formula products and growth in the PediaSure
  toddler brand.
- International Adult Nutritionals increased by 3.3% compared to the first
  half of 2016, while U.S. Adult Nutritionals decreased by 4.5% due to
  competitive and market dynamics.
- Diagnostic Products sales increased by 3.7% in the first six months of
  2017, with a 5.1% increase excluding foreign exchange, driven by share
  gains in Core Laboratory and Point of Care markets in the U.S. and
  higher international sales.
- The Other category decreased by 26.0% in the first six months of 2017,
  reflecting the sale of AMO partially offset by double-digit growth in
  Abbott’s Diabetes Care business.
- The decrease in Other Emerging Markets by 6.1% in the first six months
  of 2017 is attributed to the unfavorable impact of Venezuelan operations.
  Excluding Venezuela and foreign exchange, sales in Other Emerging
  Markets increased 4.1%.

**3. Financial Results and Key Trends**
- For the three months ended June 30, 2017, total net sales were $6,637
  million, a 24.4% increase from $5,333 million in the same period in 2016,
  with a 25.3% increase excluding foreign exchange.
- For the six months ended June 30, 2017, total net sales were $12,972
  million, a 27.0% increase from $10,218 million in the same period in
  2016, with a 27.7% increase excluding foreign exchange.
- U.S. sales increased by 42.5% in the second quarter of 2017 and by
  47.0% in the first six months of 2017.
- International sales increased by 16.3% in the second quarter of 2017
  and by 17.9% in the first six months of 2017.
- Gross profit margin decreased from 54.4% in the second quarter of 2016
  to 46.3% in the second quarter of 2017, and from 53.8% to 45.0% for
  the first six months of 2017, primarily due to higher intangible
  amortization and inventory step-up amortization from the St. Jude
  Medical acquisition.
- R&D expenses increased by $165 million in the second quarter of 2017
  and by $333 million in the first six months of 2017, driven by the
  addition of St. Jude Medical.
- Selling, general, and administrative (SG&A) expenses increased by 22.7%
  in the second quarter and 32.6% in the first six months of 2017,
  primarily due to the St. Jude Medical acquisition and integration costs,
  partially offset by cost improvement initiatives.
- Interest expense (income), net increased by $100 million in the second
  quarter and $279 million in the first six months of 2017 compared to
  2016, due to the $15.1 billion in debt issued in November 2016 to
  finance the St. Jude Medical acquisition.
- Taxes on earnings from continuing operations in the first six months of
  2017 included $430 million of tax expense related to the gain on the
  sale of the AMO business.
- Earnings from discontinued operations, net of tax, were $46 million in
  the first six months of 2017, primarily reflecting net tax benefits
  from the resolution of tax positions related to AbbVie’s operations
  prior to the 2013 separation.

**4. Identified Risks and Uncertainties**
- Abbott operates in highly competitive and regulated markets, with
  ongoing debate over healthcare product availability, delivery, and
  payment methods, which could adversely affect its operations.
- The company faces risks related to the integration of the St. Jude
  Medical acquisition, including potential challenges in combining
  operations and achieving expected synergies.
- Foreign exchange fluctuations, particularly in Venezuela, pose a risk
  to financial reporting and cash flows.
- Regulatory and legal risks are present, including the FDA warning
  letter related to the Sylmar, CA manufacturing facility acquired from
  St. Jude Medical.
- The acquisition of Alere is subject to regulatory approvals and
  potential antitrust concerns, with the FTC and European Commission
  reviewing the transaction.
- Alere is divesting certain businesses in connection with the regulatory
  review, including the Triage MeterPro and B-type Natriuretic Peptide
  assay businesses to Quidel Corporation and the subsidiary Epocal Inc.
  to Siemens Diagnostics Holding II B.V.

**5. Forward-Looking Statements and Guidance**
- Abbott expects to complete the acquisition of Alere by the end of the
  third quarter of 2017, subject to customary closing conditions and
  regulatory approvals, which are now due by September 30, 2017.
- The company expects to maintain an investment-grade debt rating.
- Abbott expects to fund cash dividends, capital expenditures, and other
  investments with cash flow from operations, cash on hand, short-term
  investments, and borrowings.
- Abbott expects to use the modified retrospective method to adopt the
  new revenue recognition standard (ASU 2014-09) and does not expect it
  to have a material impact on its consolidated financial statements.
- The company expects to evaluate the impact of recently issued
  accounting standards, including ASU 2017-07, ASU 2016-16, and ASU
  2016-02, on its consolidated financial statements.

**6. Significant Changes, Events, or Developments**
- Abbott completed the acquisition of St. Jude Medical on January 4, 2017,
  for $23.6 billion, including $13.6 billion in cash and $10 billion in
  Abbott common shares.
- Abbott sold its AMO segment to Johnson & Johnson for $4.325 billion in
  cash, completed on February 27, 2017, and recognized a pre-tax gain of
  $1.151 billion.
- Abbott sold 50 million ordinary shares of Mylan N.V. in the first six
  months of 2017, generating approximately $1.9 billion in proceeds,
  reducing its ownership interest from 14% to 3.7%.
- Abbott received a warning letter from the FDA in April 2017 regarding
  its Sylmar, CA manufacturing facility, which is part of the St. Jude
  Medical acquisition.
- Abbott entered into a $2.8 billion term loan agreement in July 2017 to
  fund the acquisition of Alere.
- Abbott commenced a tender offer to purchase its outstanding shares of
  Aleres Series B Convertible Perpetual Preferred Stock at $402 per
  share, subject to conditions.

**7. Important Figures and Tables**
**Net Sales to External Customers (in millions)**
**Three Months Ended June 30:**
| Segment | 2017 | 2016 | Total Change | Impact of Foreign Exchange | Total Change Excl. Foreign Exchange |
|---------|------|------|---------------|-----------------------------|------------------------------------|
| Established Pharmaceutical Products | $1,021 | $981 | 4.1% | 0.6% | 3.5% |
| Nutritional Products | $1,731 | $1,740 | (0.6)% | (1.1)% | 0.5% |
| Diagnostic Products | $1,273 | $1,226 | 3.8% | (1.6)% | 5.4% |
| Cardiovascular and Neuromodulation Products | $2,260 | $758 | 198.2% | (1.1)% | 199.3% |
| Other | $352 | $628 | (44.0)% | (1.3)% | (42.7)% |
| **Net Sales** | **$6,637** | **$5,333** | **24.4%** | **(0.9)%** | **25.3%** |
| **Total U.S.** | **$2,360** | **$1,655** | **42.5%** | **42.5%** | |
| **Total International** | **$4,277** | **$3,678** | **16.3%** | **(1.3)%** | **17.6%** |

**Net Sales to External Customers (in millions)**
**Six Months Ended June 30:**
| Segment | 2017 | 2016 | Total Change | Impact of Foreign Exchange | Total Change Excl. Foreign Exchange |
|---------|------|------|---------------|-----------------------------|------------------------------------|
| Established Pharmaceutical Products | $1,971 | $1,868 | 5.5% | 1.0% | 4.5% |
| Nutritional Products | $3,373 | $3,411 | (1.1)% | (0.8)% | (0.3)% |
| Diagnostic Products | $2,431 | $2,344 | 3.7% | (1.4)% | 5.1% |
| Cardiovascular and Neuromodulation Products | $4,363 | $1,467 | 197.4% | (1.1)% | 198.5% |
| Other | $834 | $1,128 | (26.0)% | (1.3)% | (24.7)% |
| **Net Sales** | **$12,972** | **$10,218** | **27.0%** | **(0.7)%** | **27.7%** |
| **Total U.S.** | **$4,684** | **$3,186** | **47.0%** | **47.0%** | |
| **Total International** | **$8,288** | **$7,032** | **17.9%** | **(1.0)%** | **18.9%** |

**Preliminary Allocation of Fair Value of St. Jude Medical Acquisition (in billions):**
| Item | 2017 |
|------|------|
| Acquired intangible assets, non-deductible | $15.0 |
| Goodwill, non-deductible | $15.1 |
| Acquired net tangible assets | $3.4 |
| Deferred income taxes recorded at acquisition | ($4.6) |
| Net debt | ($5.3) |
| **Total preliminary allocation of fair value** | **$23.6** |

**Assets and Liabilities Held for Disposition (in millions) as of December 31, 2016:**
| Item | 2016 |
|------|------|
| Trade receivables, net | $176 |
| Total inventories | $82 |
| Prepaid expenses and other current assets | $266 |
| Current assets held for disposition | $524 |
| Net property and equipment | $130 |
| Intangible assets, net of amortization | $150 |
| Goodwill | $1,966 |
| Deferred income taxes and other assets | $503 |
| Non-current assets held for disposition | $2,753 |
| **Total assets held for disposition** | **$3,266** |
| Trade accounts payable | $145 |
| Salaries, wages, commissions and other accrued liabilities | $108 |
| Current liabilities held for disposition | $253 |
| Post-employment obligations, deferred income taxes and other long-term liabilities | $34 |
| **Total liabilities held for disposition** | **$287** |

**8. Management Explanations and Justifications**
- Management attributes the significant increase in Cardiovascular and
  Neuromodulation Products sales to the St. Jude Medical acquisition,
  but notes that sales excluding the acquisition and foreign exchange
  impact decreased by 1.5% due to lower coronary stent sales and the
  favorable 2016 royalty agreement resolution.
- The decrease in International Pediatric Nutritionals is explained as
  being due to “challenging conditions in the Chinese infant formula
  market,” a qualitative explanation without numeric validation.
- The increase in U.S. Pediatric Nutritionals is attributed to “continued
  momentum of several recently launched infant formula products” and
  “growth of the PediaSure toddler brand,” with no specific quantitative
  details provided.
- The increase in Diagnostic Products sales is attributed to “share gains
  in the Core Laboratory and Point of Care markets in the U.S. and higher
  sales to various international markets,” a broad explanation without
  specific market share or sales figures.
- The decrease in the Other category is attributed to the “sale of the
  AMO business,” partially offset by “double-digit growth in Abbott’s
  Diabetes Care business,” with no further quantification of the
  Diabetes Care growth.
- The increase in R&D expenses is explained as being due to the “addition
  of the acquired St. Jude Medical business,” a straightforward explanation
  without further detail.
- The increase in SG&A expenses is attributed to the “addition of the
  acquired St. Jude Medical business as well as the incremental expenses
  to integrate St. Jude Medical with Abbott’s existing vascular business,”
  partially offset by “cost improvement initiatives,” which is a general
  term without specifics.
- The decrease in working capital is explained as being due to the “use of
  cash to fund the cash portion of the St. Jude Medical acquisition,
  repayments of debt, pension contributions and dividend payments,” with a
  partial offset from the sale of Mylan shares and business dispositions.

**9. Accounting Estimates, Judgments, and Policy Changes**
- Abbott received a warning letter from the FDA regarding its Sylmar, CA
  manufacturing facility, and has prepared a plan for corrective actions,
  which is progressing.
- The preliminary allocation of fair value for the St. Jude Medical
  acquisition is based on estimates and may be subject to material
  changes as the valuation is finalized.
- Abbott has recognized a $70 million credit to intangible amortization
  expense in the second quarter of 2017 due to measurement period
  adjustments to the value of intangibles.
- Abbott is evaluating the impact of several new accounting standards,
  including ASU 2017-07, ASU 2016-16, ASU 2016-02, and ASU 2016-01, which
  will become effective in 2018 and 2019.
- Abbott is currently evaluating the impact of ASU 2014-09 (revenue
  recognition) and expects to use the modified retrospective method for
  adoption, but has not yet quantified the potential impact.

**10. Capital Allocation and Liquidity Management**
- Abbott reduced its cash and cash equivalents from $18.6 billion at
  December 31, 2016, to $9.7 billion at June 30, 2017, primarily due to
  the St. Jude Medical acquisition, debt repayments, pension
  contributions, and dividends.
- Net cash from operating activities increased by $1.109 billion in the
  first six months of 2017 compared to 2016, due to the favorable impact
  of the St. Jude Medical acquisition and reduced pension contributions.
- Abbott has $5.0 billion in unused lines of credit available, which expire
  in 2019.
- Abbott entered into a $2.8 billion term loan agreement in July 2017 to
  fund the Alere acquisition.
- Abbott is maintaining an investment-grade debt rating and has a long-term
  debt rating of BBB by Standard & Poor's and Baa3 by Moody's.
- Abbott is utilizing cash flow from operations, cash on hand, short-term
  investments, and borrowings to fund dividends, capital expenditures, and
  other business investments.

**11. Legal, Regulatory, and Compliance Matters**
- Abbott received a warning letter from the FDA in April 2017 related to
  its Sylmar, CA manufacturing facility, which was acquired as part of the
  St. Jude Medical acquisition.
- The FDA inspection findings have not yet resulted in a material impact
  on financial results, and Abbott is implementing a corrective action plan.
- Abbott is subject to regulatory scrutiny in multiple jurisdictions, and
  tax authorities in various jurisdictions regularly review its income
  tax filings.
- The company expects the recorded amount of gross unrecognized tax benefits
  to decrease by $200 million to $350 million, including cash adjustments,
  within the next twelve months as a result of concluding various domestic
  and international tax matters.
- Abbott’s U.S. federal income tax returns are settled through 2013, and
  St. Jude Medical’s federal income tax returns are settled through 2013
  except for one item.
\end{verbatim}
\end{tcolorbox}

%% file: appendixes/appendix_3_aaer_dataset_details.tex
\section{Appendix C. Details on AAER Dataset and Preprocessing}
\label{appendix:aaer_details}

This appendix details the acquisition, preprocessing, and filtering pipeline for the Accounting and Auditing Enforcement Releases (AAERs) used to generate fraud labels for our study.

\subsection{Raw AAER Data Acquisition and Initial Characteristics}
Raw AAER data was programmatically downloaded as JSON objects using the \texttt{sec-api.io} service, specifically querying their AAER Database API endpoint\footnote{\url{https://sec-api.io/docs/aaer-database-api}}. Approximately \textbf{3,300} AAERs were initially collected. Each JSON object contains structured information about an enforcement release.

An example structure of a downloaded raw AAER JSON object is shown below:
\begin{tcolorbox}[enhanced, breakable, colback=orange!5!white, colframe=orange!75!black, title=Sample Raw AAER JSON Object, fonttitle=\bfseries, boxsep=2pt, left=2pt, right=2pt]
\small
\begin{verbatim}
{
    "id": "c9ac87509126bd0f1f62c89346cae52d",
    "dateTime": "2004-02-25T09:21:21-05:00",
    "aaerNo": "AAER-1964",
    "releaseNo": ["LR-18595"],
    "respondents": [{
        "name": "FOO",
        "type": "individual"
    }],
    "respondentsText": "FOO",
    "urls": [{
        "type": "primary",
        "url": "https://www.sec.gov/enforcement-litigation/litigation-releases/lr-18595"
    }],
    "summary": "The SEC filed a complaint against FOO...",
    "tags": ["disclosure fraud", "financial reporting fraud"],
    "entities": [{
        "name": "FOO",
        "type": "individual",
        "role": "defendant"
    }, {
        "name": "Just for Feet, Inc.",
        "type": "company",
        "role": "entity involved in the fraud",
        "cik": "918111",
        "ticker": "FEET"
    }],
    "complaints": [
        "Ruttenberg was instrumental in the acquisition of fraudulent confirmations..."
    ],
    "parallelActionsTakenBy": ["United States Department of Justice", "..."],
    "hasAgreedToSettlement": false,
    "hasAgreedToPayPenalty": false,
    "penaltyAmounts": [],
    "requestedRelief": ["permanent injunction", "disgorgement", "..."],
    "violatedSections": ["Section 17(a) of the Securities Act of 1933", "..."],
    "otherAgenciesInvolved": [{"name": "United States Department of Justice", ...}]
}
\end{verbatim}
\end{tcolorbox}

\begin{figure}[htbp]
    \centering
    \includegraphics[width=0.5\linewidth]{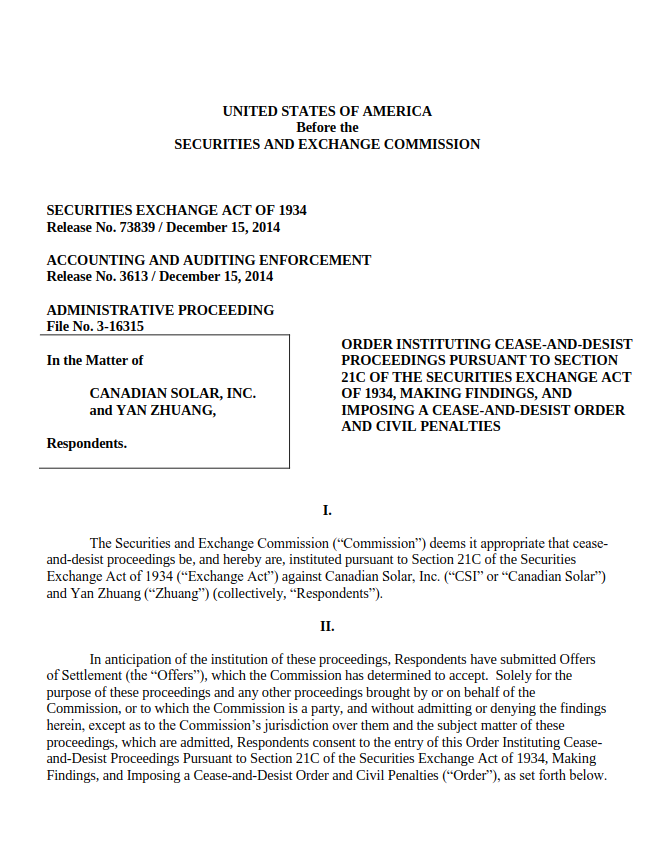}
    \caption{Example first page of a primary document linked from an AAER release.}
    \label{fig:sample_aaer_pdf_page1}
\end{figure}

Each AAER JSON object includes a `primary\_url` field, which typically links to a detailed document (often a PDF or HTML page) on the SEC website. Figure \ref{fig:sample_aaer_pdf_page1} shows an example of the first page of such a document.

\subsection{AAER Preprocessing Pipeline}
The downloaded AAERs underwent a multi-step preprocessing pipeline to extract relevant information and filter them for suitability in our fraud detection task.

\subsubsection{Parsing and Initial Data Extraction}
The pipeline began by iterating through all downloaded JSON files. For each AAER instance, key fields such as `aaerNo` (cleaned to a standard format, e.g., "AAER-XXXX"), `dateTime` (date part extracted), `tags`, `summary`, `complaints`, the `primary\_url`, and company-specific details (name, role, CIK) from the `entities` list were extracted. This information was structured into a tabular format for further processing.

\subsubsection{Extraction of Fiscal Quarter Violation Information}
A critical challenge is that raw AAER JSONs do not directly provide the specific fiscal years and quarters during which the violations occurred. This temporal information is essential for linking fraud events to quarterly financial reports. To address this, we implemented an automated extraction process:
\begin{enumerate}
    \item \textbf{Content Retrieval}: For each AAER, the content of the document linked by its `primary\_url` was fetched. This involved using web automation tools (like Selenium with ChromeDriver) to download PDF documents or render HTML pages, followed by text extraction (using libraries like PyPDF2 for PDFs and BeautifulSoup for HTML). This process was parallelized to expedite the processing of numerous documents.
    \item \textbf{Fiscal Quarter Identification}: The extracted text content from each AAER document was then processed by the Qwen3 32B model. This model was tasked with identifying the precise quarters (e.g., "2018q1", "2019q2") and companies associated with the financial violations described in the text. The process was guided by the system prompt detailed below, designed to ensure consistent and accurate extraction. API interactions were managed with rate limiting to ensure robust performance.
    \item The promt was tailored to put emphasis on earning mistatements, as it is what our work considers as Fraud. 
\end{enumerate}

\begin{tcolorbox}[enhanced, breakable, colback=cyan!5!white, colframe=cyan!75!black, title=System Prompt for Fiscal Quarter Extraction from AAER Content, fonttitle=\bfseries, boxsep=2pt, left=2pt, right=2pt]
\small
\begin{verbatim}
You are a specialized AI agent tasked with meticulously extracting detailed
information about **earnings misstatements** (financial statement violations
that directly impact the calculation of reported earnings, income, assets,
or liabilities) from U.S. Securities and Exchange Commission (SEC) Accounting
and Auditing Enforcement Releases (AAERs). Your goal is to deconstruct complex
legal and financial text into structured, factual data.

### **Objective**
Your primary objective is to identify every quarter in which a **true earning
misstatement** occurred, specify the company responsible, detail the specific
types of earning misstatements based on predefined categories, and describe the
fraudulent scheme that led to these misstatements. You must adhere strictly
to the formats and rules defined below.

### **Key Definitions: `LIST_MISTATEMENT_TYPE`**
You must categorize all identified **earnings misstatements** using **only**
the types from this predefined list. An "earnings misstatement" directly alters
the reported financial performance or position (e.g., net income, assets,
liabilities, equity balances). Violations related *solely* to disclosure
failures that do not alter the numerical financial statements (e.g., failure
to disclose related party relationships without affecting specific account
balances, or control issues) should *not* be categorized here unless they
clearly result in a numerical misstatement of an account listed below.

*   **Revenue**: Overstating or understating sales or income. This includes
    premature revenue recognition, fictitious sales, or improper income
    classification directly impacting the income statement.
*   **Other Expense/Shareholder Equity Account**: Manipulating expenses not
    directly related to cost of goods sold (e.g., operating, selling, general
    \& administrative expenses, R\&D), or directly misstating equity accounts
    (like retained earnings, common stock, additional paid-in capital) through
    improper accounting entries that affect net income or equity balances. Do
    not include in this category disclore fraud or governance issues that do
    not impact financial accounts.
*   **Assets Valuation** : Improperly recognition of assets or their values.
    This includes inflating asset values (e.g., property, plant \& equipment,
    intangible assets) or failing to recognize impairments, which directly
    impacts the balance sheet. 
*   **Capitalized Costs as Assets**: Improperly recording expenses as long-term
    assets (e.g., property, plant \& equipment, intangible assets) to inflate
    current period income by reducing expenses.
*   **Accounts Receivable**: Overstating the money owed by customers. This
    includes recording fictitious sales, failing to write off uncollectible
    receivables, or otherwise inflating the asset balance.
*   **Inventory**: Overstating the value of goods for sale. This includes
    counting non-existent inventory, improper valuation methods, or
    misclassifying costs.
*   **Cost of Goods Sold (COGS)**: Understating the direct costs of production.
    Often linked to inventory manipulation (e.g., overstating inventory leads
    to understated COGS), directly impacting gross profit and net income.
*   **Reserve Account**: Manipulating funds set aside for future contingent
    liabilities (e.g., warranty, litigation reserves, bad debt reserves). This
    includes understating reserves to boost current income or overstating them
    to create "cookie jar" reserves for future manipulation, directly impacting
    expenses or liabilities.
*   **Liabilities**: Understating company obligations. This includes concealing
    debt, failing to record accrued expenses (e.g., unbilled services,
    payroll), or misclassifying liabilities to improve financial ratios or
    conceal obligations.
*   **Marketable Securities**: Misstating the value of short-term investments.
    This includes improper valuation (e.g., failing to mark to market when
    required) or failing to recognize impairment losses, directly impacting
    asset values and potentially income.
*   **Allowance for Bad Debt**: Understating the estimated uncollectible
    accounts receivable to inflate net receivables and income. This is a
    specific type of reserve manipulation.
*   **Payables**: Understating money owed to suppliers. This includes delaying
    invoice recording, concealing vendor liabilities, or manipulating cut-off
    dates, directly impacting liabilities and potentially expenses.

---

### **Input Format**
You will receive a dictionary containing:
*   `"aaerNo"`: The unique identifier of the AAER.
*   `"content"`: The full text of the AAER.
*   `"entities"`: A list of dictionaries, each representing an entity (company
    or individual) involved in the AAER.

---

### **Output Format**
You must generate a JSON list of dictionaries. Each dictionary represents a
single fraudulent scheme by a specific company in a specific quarter.

```json
[
    {
        "quarter": "YYYYqQ",
        "is_fiscal_quarter": true,
        "fraud_scheme_description": "A concise, factual description of the
            earning misstatement mechanics, focusing on how the numerical
            financial statements were altered. You must also justify the
            selection of the misstatements indicated in the 'misstatements'
            field. The justification should be provided only when the
            'misstatements' field is not empty. The justification should be
            in the form of a list of sentences, each explaining why a specific
            misstatement type was selected for that quarter. For example:
            \n- Revenue because the company recorded fictitious sales
            transactions.\n- Accounts Receivable because the inflated sales
            led to an overstatement of amounts owed by customers.",
        "misstatements": ["Type1", "Type2", "Type3", ...],
        "misstating_company": {
            "name": "Company Name",
            "role": "respondent",
            "cik": "0001234567"
        }
    }
]
\end{verbatim}
\end{tcolorbox}

The extracted fiscal quarters were then associated with their respective AAERs in our structured dataset.

\subsubsection{Filtering and Refinement}
The dataset of AAERs, now augmented with extracted fiscal quarters, underwent several filtering steps to refine its suitability for our fraud detection task:
\begin{itemize}
  
    \item \textbf{Date Filtering}: To align with the availability of our financial features (which start from 2009), only AAERs with identified violation years from 2009 onwards were considered for the primary dataset used in the experiments.
\end{itemize}
This comprehensive filtering process significantly narrowed down the set of AAERs to those most pertinent for training and evaluating models for company-level financial statement fraud detection.

\subsection{Final Fraud Label Statistics}
After the complete preprocessing and filtering pipeline:
\begin{itemize}
    \item From the initial $\sim$3,300 AAERs, the filtering steps (for relevance based on tags, direct company culpability, CIK presence, and violations occurring from 2009 onwards) resulted in \textbf{249 unique AAERs}.
    \item These 249 AAERs were then merged with our financial features dataset. Due to factors such as non-overlapping CIKs between the AAER dataset and the companies present in our financial reports database, and further filtering based on the completeness threshold for financial features, the number of AAERs contributing to fraud labels in our final experimental dataset was reduced to \textbf{137}.
    \item These 137 AAERs collectively identified \textbf{511 firm-quarters} as fraudulent instances. These instances formed the positive class in our fraud detection experiments.
\end{itemize}
This rigorous process ensures that the fraud labels used in our study are well-defined, temporally accurate, and directly linkable to the financial and textual data of the companies involved.

%% file: appendixes/appendix_4_dataset_splitting.tex
\section{Appendix D: Final Dataset Construction and Data Splitting Methodology}
\label{appendix:dataset_splitting}

This appendix outlines the procedures for constructing the final dataset used in our experiments and details the different data splitting strategies employed to evaluate our models under various conditions.

\subsection{Final Dataset Construction}
The creation of our final experimental dataset involved several key steps: merging data from different sources, addressing class imbalance, and ensuring data consistency.

\subsubsection{Data Merging}
The foundational step was the integration of three primary data sources:
\begin{itemize}
    \item \textbf{Financial Data}: As described in Section \ref{subsec:financial-data}, this includes 122 engineered financial indicators derived from quarterly reports.
    \item \textbf{ Summarized MD\&A (SMD\&A)}: Textual data obtained from summarizing MD\&A sections, as detailed in Appendix \ref{appendix:mda_details}.
    \item \textbf{AAER-derived Fraud Labels}: Binary fraud labels (fraud/non-fraud) for specific firm-quarters, processed as described in Appendix \ref{appendix:aaer_details}.
\end{itemize}
These datasets were merged based on common identifiers: Central Index Key (CIK), fiscal year, and fiscal quarter. Company names were standardized (converted to lowercase) before merging to ensure consistency. A mapping between CIKs and company names was also created for reference. Any records that did not have corresponding entries across these essential dimensions or lacked MD\&A data were excluded.

\subsubsection{Handling Class Imbalance: Stratified Downsampling of Non-Fraud Cases}
\label{subsubsec:downsampling_algo}
Financial fraud is a rare event, leading to highly imbalanced datasets. To create a tractable dataset for experimentation while preserving a significant level of imbalance, we downsampled the non-fraudulent cases to achieve a target fraud rate of approximately \textbf{5.03\%} in the final dataset (as discussed in Section \ref{subsec:class-imbalance-handling}). This was performed carefully to maintain the underlying characteristics of the non-fraud data. The stratified downsampling algorithm is as follows:

\begin{enumerate}
    \item \textbf{Segregation}: The merged dataset was divided into two subsets: fraudulent firm-quarter samples and non-fraudulent firm-quarter samples.
    \item \textbf{Target Calculation}:
        \begin{itemize}
            \item Let $N_{fraud}$ be the total number of fraudulent samples.
            \item The target number of non-fraudulent samples ($N_{non\_fraud\_target}$) to achieve the desired overall fraud percentage ($P_{fraud} = 5.03\%$) was calculated as:
                  $N_{non\_fraud\_target} = N_{fraud} \times \frac{100 - P_{fraud}}{P_{fraud}}$.
        \end{itemize}
    \item \textbf{Proportional Group Sampling (Initial Pass)}:
        \begin{itemize}
            \item The non-fraudulent subset was grouped by `year` and `sicagg`. The `sicagg` field represents an aggregated Standard Industrial Classification code, corresponding to high-level industry sectors derived from the first two digits of the SIC code, based on classifications from official sources (e.g., \url{https://siccode.com/}). This grouping aims to preserve temporal and industry distributions.
            \item A global downsampling factor ($F_{downsample}$) was computed: $F_{downsample} = N_{non\_fraud\_target} / N_{non\_fraud\_original}$, where $N_{non\_fraud\_original}$ is the total count of non-fraudulent samples before downsampling.
            \item For each (`year`, `sicagg`) group within the non-fraudulent data:
                \begin{itemize}
                    \item The target number of samples for this group ($N_{group\_target}$) was calculated by multiplying the original size of this group by $F_{downsample}$.
                    \item $N_{group\_target}$ was adjusted to be at least 1 (if the group was non-empty and $N_{group\_target}$ was positive) and no more than the actual number of samples available in that group.
                    \item $N_{group\_target}$ samples were randomly selected from this specific group.
                \end{itemize}
            \item All samples selected from these groups were collected.
        \end{itemize}
    \item \textbf{Refinement Pass (If Target Not Met)}:
        \begin{itemize}
            \item If the total number of non-fraudulent samples collected in the initial pass was less than $N_{non\_fraud\_target}$, a refinement step was performed.
            \item Groups that still contained unselected non-fraudulent samples were identified.
            \item Additional samples were iteratively drawn from these groups, prioritizing those with more remaining available samples, until $N_{non\_fraud\_target}$ was reached or no more unselected samples were available. This ensures the target size is met more closely while still favoring the original distribution.
        \end{itemize}
    \item \textbf{Final Dataset Assembly}: The original set of $N_{fraud}$ fraudulent samples was combined with the $N_{non\_fraud\_target}$ downsampled non-fraudulent samples to form the final dataset used for all experiments. A fixed random seed was used during the sampling process to ensure reproducibility.
\end{enumerate}
This stratified downsampling ensures that while the dataset is made more balanced, the non-fraudulent samples still reflect the temporal and sectoral diversity of the original population.

\subsubsection{Final Dataset Composition}
After merging and downsampling, the final dataset used for our experiments consists of \textbf{10,159} firm-quarter observations. This includes:
\begin{itemize}
    \item \textbf{Fraudulent Samples}: \textbf{511} (approximately \textbf{5.03\%})
    \item \textbf{Non-Fraudulent Samples}: \textbf{9,648} (approximately \textbf{94.97\%})
    \item \textbf{Unique Companies}: \textbf{5,658}
\end{itemize}
It is important to note that a single company (identified by its CIK) can contribute multiple firm-quarter observations to the dataset, and these can include both fraudulent and non-fraudulent instances over different time periods. The dataset spans from 2009 to 2021. The overall distribution of samples per aggregated industry sector (based on `sicagg`) in this final dataset is presented in Table \ref{tab:final_dataset_industry_dist}.

\begin{table}[htbp]
  \centering
  \caption{Overall industry sector distribution in the final experimental dataset (after downsampling). Numbers represent firm-quarter instances.}
  \label{tab:final_dataset_industry_dist}
  \begin{tabular}{lc}
    \toprule
    Industry Sector (Aggregated SIC - `sicagg`) & Total Samples \\
    \midrule
    Agriculture, Forestry, And Fishing & \textbf{39} \\
    Construction & \textbf{110} \\
    Finance, Insurance, And Real Estate & \textbf{2320} \\
    Manufacturing & \textbf{3789} \\
    Mining & \textbf{640} \\
    Public Administration & \textbf{5} \\
    Retail Trade & \textbf{426} \\
    Services & \textbf{1770} \\
    Transportation \& Public Utilities & \textbf{786} \\
    Wholesale Trade & \textbf{274} \\
    \bottomrule
  \end{tabular}
\end{table}

\subsection{Data Splitting Strategies for FSFD Tasks}
To evaluate model performance under different assumptions and levels of difficulty, we employed three distinct data splitting strategies, corresponding to the tasks defined in Section \ref{sec:tasks-definition}. For tasks involving cross-validation, 5 folds were used, and a fixed random seed was employed for fold generation to ensure reproducibility.

\subsubsection{Classic FSFD: Random K-Fold Cross-Validation}
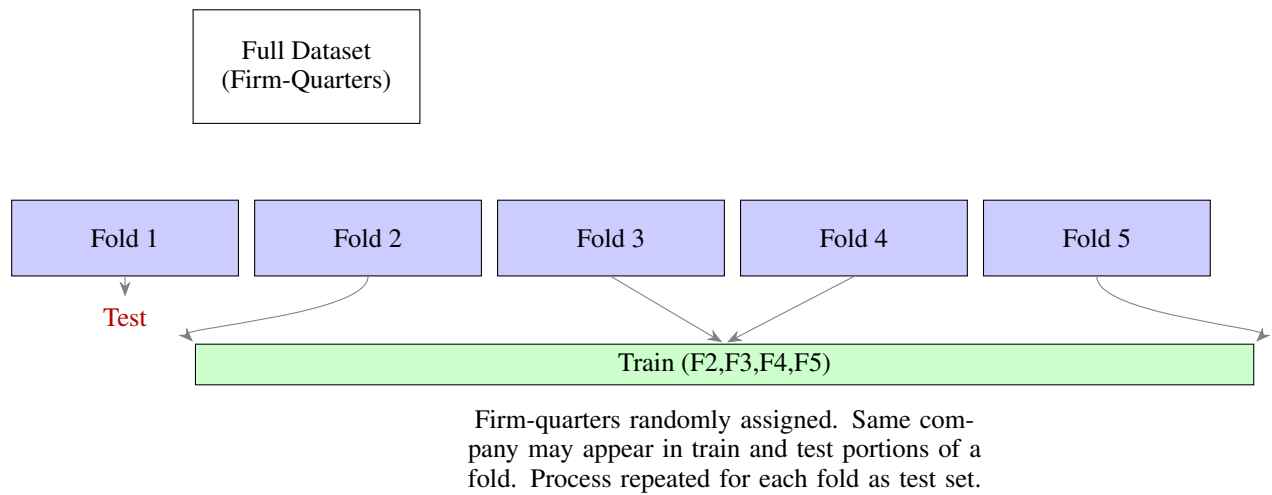
\begin{figure}[htbp]
    \centering
    \begin{tikzpicture}[node distance=0.5cm and 1cm]
        \node[draw, rectangle, minimum height=1.5cm, minimum width=3cm, align=center] (dataset) {Full Dataset \\ (Firm-Quarters)};

        \node[draw, rectangle, fill=blue!20, minimum height=1cm, minimum width=3.0cm, below=1cm of dataset, xshift=-2.4cm] (f1) {Fold 1};
        \node[draw, rectangle, fill=blue!20, minimum height=1cm, minimum width=3cm, right=0.2cm of f1] (f2) {Fold 2};
        \node[draw, rectangle, fill=blue!20, minimum height=1cm, minimum width=3cm, right=0.2cm of f2] (f3) {Fold 3};
        \node[draw, rectangle, fill=blue!20, minimum height=1cm, minimum width=3cm, right=0.2cm of f3] (f4) {Fold 4};
        \node[draw, rectangle, fill=blue!20, minimum height=1cm, minimum width=3cm, right=0.2cm of f4] (f5) {Fold 5};

        \node[below=0.3cm of f1, text width=1.2cm, align=center, color=red!70!black, font=\mdseries] (t1) {Test};
        \node[draw, rectangle, fill=green!20, minimum height=0.5cm, minimum width=14cm, below right = 0.1cm and 0.2cm of t1, align=center] (tr1) {Train (F2,F3,F4,F5)};
        \draw[-{Stealth[length=2mm, width=1.5mm]}, shorten >=1pt, color=gray] (f1.south) -- (t1.north);
        \draw[-{Stealth[length=2mm, width=1.5mm]}, shorten >=1pt, color=gray] (f2.south) .. controls +(south:0.5cm) and +(north west:0.3cm) .. (tr1.north west);
        \draw[-{Stealth[length=2mm, width=1.5mm]}, shorten >=1pt, color=gray] (f3.south) -- (tr1.north);
        \draw[-{Stealth[length=2mm, width=1.5mm]}, shorten >=1pt, color=gray] (f4.south) -- (tr1.north);
        \draw[-{Stealth[length=2mm, width=1.5mm]}, shorten >=1pt, color=gray] (f5.south) .. controls +(south:0.5cm) and +(north east:0.3cm) .. (tr1.north east);

        \node[text width=10cm, below=0.2cm of tr1, align=center] (desc) {Firm-quarters randomly assigned. Same company may appear in train and test portions of a fold. Process repeated for each fold as test set.};
    \end{tikzpicture}
    \caption{Conceptual diagram of Classic FSFD (Random K-Fold) splitting. Each fold serves as a test set once, with the remaining folds as training.}
    \label{fig:random_split_konva}
\end{figure}

This strategy represents the traditional approach to evaluating FSFD models.
\begin{itemize}
    \item \textbf{Methodology}: The final dataset of \textbf{10,159} firm-quarter observations was split into 5 folds using random sampling. Stratification was applied based on the `is\_fraud` label to ensure that each fold maintained approximately the same \textbf{5.03\%} fraud ratio as the overall dataset.
    \item \textbf{Characteristics}: In this setup, observations from the same company can appear in both the training and testing sets of a given fold (though not the same observation). This allows the model to potentially learn company-specific patterns.
    \item \textbf{Fold Statistics (Averages over 5 Folds)}:
        \begin{itemize}
            \item Test Set Size: $\sim$\textbf{2032} samples
            \item Fraud Samples in Test: $\sim$\textbf{102} (Fraud Ratio: $\sim$\textbf{5.03\%})
            \item Industry distribution in test folds (e.g., `Manufacturing` $\sim$\textbf{758} samples with $\sim$\textbf{52} fraud; `Services` $\sim$\textbf{354} samples with $\sim$\textbf{22} fraud) typically reflected the overall dataset distribution due to random sampling.
        \end{itemize}
    \item \textbf{Illustration}: See Figure \ref{fig:random_split_konva}.
\end{itemize}

\subsubsection{Company-Isolated FSFD (CI-FSFD): Company-Based K-Fold Cross-Validation}
This strategy imposes a stricter evaluation by ensuring that companies seen during training are not present in the test set.
\begin{itemize}
    \item \textbf{Methodology}: The dataset was split into 5 folds at the company (CIK) level. All firm-quarter observations belonging to a specific company (which may include a mix of fraud and non-fraud instances for that company) were assigned entirely to one fold. The assignment of companies to folds was performed aiming to balance the number of fraud reports and total reports per fold, and preserve the industry sector distribution within each fold's test set as much as possible.
    \item \textbf{Characteristics}: This split tests the model's ability to generalize to entirely unseen companies, preventing it from relying on idiosyncratic patterns of companies present in the training data.
    \item \textbf{Fold Statistics (Averages over 5 Folds)}:
        \begin{itemize}
            \item Test Set Size: $\sim$\textbf{2032} samples
            \item Unique Companies in Test: $\sim$\textbf{1132}
            \item Fraud Samples in Test: $\sim$\textbf{102} (Fraud Ratio: $\sim$\textbf{5.03\%})
            \item Industry distributions were actively balanced. For example, across test folds, `Manufacturing` had $\sim$\textbf{758} samples (with $\sim$\textbf{52} fraud), and `Services` had $\sim$\textbf{354} samples (with $\sim$\textbf{22} fraud).
        \end{itemize}
    \item \textbf{Illustration}: See Figure \ref{fig:cik_split_konva}.
\end{itemize}

\begin{figure}[htbp]
    \centering
     \begin{tikzpicture}[node distance=0.5cm and 1cm]
        \node[draw, rectangle, minimum height=1.5cm, minimum width=3.5cm, align=center] (dataset) {Full Dataset \\ (Grouped by Company)};

        \node[draw, ellipse, fill=yellow!30, minimum width=1.5cm, minimum height=0.8cm, below=1cm of dataset, xshift=-2.5cm] (g1) {Companies A};
        \node[draw, ellipse, fill=orange!30, minimum width=1.5cm, minimum height=0.8cm, right=0.3cm of g1] (g2) {Companies B};
        \node[draw, ellipse, fill=cyan!30, minimum width=1.5cm, minimum height=0.8cm, right=0.3cm of g2] (g3) {Companies C};
        \node[draw, ellipse, fill=magenta!30, minimum width=1.5cm, minimum height=0.8cm, right=0.3cm of g3] (g4) {Companies D};
        \node[draw, ellipse, fill=lime!30, minimum width=1.5cm, minimum height=0.8cm, right=0.3cm of g4] (g5) {Companies E};

        \node[draw, rectangle, fill=red!20, minimum height=1cm, minimum width=2cm, below=1cm of g1, xshift=0cm, align=center] (test_f1) {Fold 1 (Test) \\ (e.g., Co. Group A)};
        \node[draw, rectangle, fill=green!20, minimum height=1cm, minimum width=14cm, right=0.5cm of test_f1, align=center] (train_f1) {Train for Fold 1 \\ (e.g., Co. Groups B, C, D, E)};

        \draw[-{Stealth[length=2mm, width=1.5mm]}, shorten >=1pt, color=gray] (g1.south) -- (test_f1.north);
        \draw[-{Stealth[length=2mm, width=1.5mm]}, shorten >=1pt, color=gray] (g2.south) .. controls +(south:0.5cm) and +(north west:0.3cm) .. (train_f1.north west);
        \draw[-{Stealth[length=2mm, width=1.5mm]}, shorten >=1pt, color=gray] (g3.south) -- (train_f1.north);
        \draw[-{Stealth[length=2mm, width=1.5mm]}, shorten >=1pt, color=gray] (g4.south) -- (train_f1.north);
        \draw[-{Stealth[length=2mm, width=1.5mm]}, shorten >=1pt, color=gray] (g5.south) .. controls +(south:0.5cm) and +(north east:0.3cm) .. (train_f1.north east);

        \node[text width=10cm, below=0.3cm of train_f1, xshift=-2cm, align=center] (desc) {Groups of companies are assigned to folds. If a company's data (all its firm-quarter instances) is in the test set of a fold, none of its data appears in the training set for that fold.};
    \end{tikzpicture}
    \caption{Conceptual diagram of Company-Isolated FSFD (CI-FSFD) splitting. Companies (and all their associated firm-quarter instances) are assigned to folds.}
    \label{fig:cik_split_konva}
\end{figure}

%% file: appendixes/appendix_5_hyperparams.tex
\section{Appendix E: Model's Details and Hyperparameters}
\label{appendix:hyperparameters}

This appendix provides detailed model configurations and hyperparameters for all models used in our experiments: Logistic Regression, MLP, Random Forest (LightGBM), XGBoost, RCMA-adapted, and LLM-based models. For all models, hyperparameters were optimized using Hyperopt, targeting the maximization of the F1-score on a dedicated validation set (typically 10\% of the training data for that specific fold/split). The `decision\_threshold` reported for classification models is the optimal threshold found on the validation set. It is important to note that the same set of optimized hyperparameters was applied across all folds for a given model and task (e.g., Classic FSFD or CI-FSFD).

\subsection{Logistic Regression (MLP-Classifier with no Hidden Layers)}
Our Logistic Regression baseline is implemented as a specialized case of the MLP Classifier with no hidden layers.

\begin{table}[h!]
\caption{Optimized Hyperparameters for Logistic Regression.} \label{tab:logistic_hyperparams}
\centering
\begin{tabular}{ll}
\toprule
Parameter & Value \\
\midrule
Features Type & Dechow \\
Learning Rate & 0.1 \\
Batch Size & 64 \\
Dropout Rate & 0.459 \\
Epochs & 2 \\
Patience & 20 \\
Oversample & True \\
Standardize & True \\
Decision Threshold & 0.5 \\
\bottomrule
\end{tabular}
\end{table}

\subsection{Multi-Layer Perceptron (MLP)}
The MLP Classifier uses a feed-forward neural network architecture.

\begin{table}[h!]
\caption{Optimized Hyperparameters for MLP.} \label{tab:mlp_hyperparams}
\centering
\begin{tabular}{ll}
\toprule
Parameter & Value \\
\midrule
\textbf{Classic FSFD Task} \\
Features Type & Financial (122 features) \\
Hidden Dims & [512] \\
Learning Rate & 0.1 \\
Batch Size & 128 \\
Dropout Rate & 0.413 \\
\midrule
\textbf{CI-FSFD Task} \\
Features Type & Financial (122 features) \\
Hidden Dims & [512, 512] \\
Learning Rate & 0.001 \\
Batch Size & 32 \\
Dropout Rate & 0.145 \\
\midrule
Common Parameters \\
Epochs & 2 \\
Patience & 20 \\
Oversample & True \\
Standardize & False \\
Decision Threshold & 0.5 \\
\bottomrule
\end{tabular}
\end{table}

\subsection{Random Forest (LightGBM Implementation)}
For the Random Forest baseline, we utilized the Random Forest mode of the LightGBM library \citep{ke_lightgbm_nodate}.

\begin{table}[h!]
\caption{Optimized Hyperparameters for Random Forest (LightGBM).} \label{tab:rf_hyperparams}
\centering
\begin{tabular}{ll}
\toprule
Parameter & Value \\
\midrule
\textbf{Classic FSFD Task} \\
Features Type & Financial (122 features) \\
Learning Rate & 0.0799 \\
Max Depth & 7 \\
Num Estimators & 200 \\
Num Leaves & 5 \\
Standardize & True \\
\midrule
\textbf{CI-FSFD Task} \\
Features Type & Financial (122 features) \\
Learning Rate & 0.0996 \\
Max Depth & 78 \\
Num Estimators & 200 \\
Num Leaves & 50 \\
Standardize & False \\
\midrule
Common Parameter \\
Decision Threshold & 0.5 \\
\bottomrule
\end{tabular}
\end{table}

\subsection{XGBoost}
The XGBoost classifier \citep{DBLP:journals/corr/ChenG16} was optimized with Hyperopt.
\begin{table}[h!]
\caption{Optimized Hyperparameters for XGBoost.} \label{tab:xgb_hyperparams}
\centering
\begin{tabular}{ll}
\toprule
Parameter & Value \\
\midrule
\textbf{Classic FSFD Task} \\
Features Type & Financial (122 features) and Dechow \\
Learning Rate & 0.05 \\
Max Depth & 0 \\
Num Estimators & 50 \\
Num Leaves & 50 \\
Standardize & True \\
\midrule
\textbf{CI-FSFD Task} \\
Features Type & Financial (122 features) \\
Learning Rate & 0.090 \\
Max Depth & 88 \\
Num Estimators & 200 \\
Num Leaves & 50 \\
Standardize & True \\
\midrule
Common Parameter \\
Decision Threshold & 0.5 \\
\bottomrule
\end{tabular}
\end{table}

\subsection{RCMA-adapted Model}
The RCMA-adapted model architecture and training parameters were based on the work of \citet{wang_attentive_2023}, with modifications for our SBERT-based text processing.

\begin{table}[h!]
\caption{Optimized Hyperparameters for RCMA-adapted Model.} \label{tab:rcma_hyperparams}
\centering
\begin{tabular}{ll}
\toprule
Parameter & Value \\
\midrule
SBERT Model Name & `jinaai/jina-embeddings-v2-small-en` \\
SBERT Output Dim & 512 \\
Trainable SBERT Layers & 1 \\
Max SMD\&A Length & 8192 tokens \\
Financial Features & 122 \\
Num Financial Groups & 7 \\
Financial Embedding Dim & 512 \\
Text Embedding Dim & 512 \\
Dropout Rate & 0.05 \\
Learning Rate & 1e-4 \\
Batch Size & 8 \\
Validation Batch Size & 8 \\
Pos Weight Beta (for FocalLoss) & 0.75 \\
Focal Gamma (for FocalLoss) & 2.0 \\
Gradient Accumulation Steps & 4 \\
\midrule
\textbf{Classic FSFD Task} \\
MLP Hidden Dims & [128] \\
Epochs & 10 \\
Patience & 7 \\
Consistency Loss Weight & 0.2 \\
Oversample & False \\
\midrule
\textbf{CI-FSFD Task} \\
MLP Hidden Dims & [512] \\
Epochs & 20 \\
Patience & 10 \\
Consistency Loss Weight & 0 \\
Oversample & True \\
\bottomrule
\end{tabular}
\end{table}

\subsection{LLM-based Models}
This section details the configuration for our Large Language Models, including Llama-3.1 8B and Fino1 8B, in various fine-tuning and zero-shot settings. All LLMs use LoRA for fine-tuning.

\subsubsection{Financial-only Models (Llama-3.1 8B and Fino1 8B)}
These models are fine-tuned exclusively on financial text.

\begin{table}[h!]
\caption{Hyperparameters for LLM-based Models (Financial-only).} \label{tab:llm_fin_hyperparams}
\centering
\begin{tabular}{ll}
\toprule
Parameter & Value \\
\midrule
\textbf{Llama-3.1 8B (Financial)} \\
Model URL & `unsloth/Llama-3.1-8B-unsloth-bnb-4bit` \\
LoRA R & 8 \\
LoRA Alpha & 8 \\
LoRA Dropout & 0.05 \\
Layers to Finetune & 32 \\
Max Context & 1500 tokens \\
Batch Size & 4 \\
Gradient Accumulation Steps & 2 \\
Learning Rate & 1e-4 \\
LoRA Target Modules & `q\_proj`, `v\_proj`, `up\_proj`, `down\_proj`, `gate\_proj` \\
\midrule
\textbf{Fino1 8B (Financial)} \\
Model URL & `TheFinAI/Fino1-8B` \\
LoRA R & 8 \\
LoRA Alpha & 8 \\
LoRA Dropout & 0.05 \\
Layers to Finetune & 32 \\
Max Context & 1500 tokens \\
Batch Size & 4 \\
Gradient Accumulation Steps & 2 \\
Learning Rate & 1e-4 \\
LoRA Target Modules & `q\_proj`, `v\_proj`, `up\_proj`, `down\_proj`, `gate\_proj` \\
\midrule
Common Parameters \\
Epochs (Classic FSFD Task) & 10 \\
Epochs (CI-FSFD Task) & 20 \\
Max New Tokens & 1 \\
Only Completion & True \\
Undersample & True \\
Run Eval on Start & False \\
\bottomrule
\end{tabular}
\end{table}

\subsubsection{SMD\&A-only Models (Llama-3.1 8B, Fino1 8B, Fino1 14B)}
These models are fine-tuned exclusively on Summary Management Discussion \& Analysis (SMD\&A) text.

\begin{table}[h!]
\caption{Hyperparameters for LLM-based Models (SMD\&A-only).} \label{tab:llm_mda_hyperparams}
\centering
\begin{tabular}{ll}
\toprule
Parameter & Value \\
\midrule
\textbf{Llama-3.1 8B (SMD\&A)} \\
Model URL & `unsloth/Llama-3.1-8B-unsloth-bnb-4bit` \\
LoRA R & 8 \\
LoRA Alpha & 8 \\
LoRA Dropout & 0.05 \\
Layers to Finetune & 32 \\
Max Context & 8500 tokens \\
Batch Size & 8 \\
Gradient Accumulation Steps & 1 \\
Learning Rate & 1e-4 \\
LoRA Target Modules (Classic FSFD Task) & `q\_proj`, `v\_proj`, `up\_proj`, `down\_proj`, `gate\_proj` \\
LoRA Target Modules (CI-FSFD Task) & `q\_proj`, `v\_proj`, `up\_proj`, `down\_proj`, `gate\_proj`, `lm\_head` \\
\midrule
\textbf{Fino1 8B (SMD\&A)} \\
Model URL & `TheFinAI/Fino1-8B` \\
LoRA R & 8 \\
LoRA Alpha & 8 \\
LoRA Dropout & 0.05 \\
Layers to Finetune & 32 \\
Max Context & 8500 tokens \\
Batch Size & 8 \\
Gradient Accumulation Steps & 1 \\
Learning Rate & 1e-4 \\
LoRA Target Modules (Classic FSFD Task) & `q\_proj`, `v\_proj`, `up\_proj`, `down\_proj`, `gate\_proj` \\
LoRA Target Modules (CI-FSFD Task) & `q\_proj`, `v\_proj`, `up\_proj`, `down\_proj`, `gate\_proj`, `lm\_head` \\
\midrule
\textbf{Fino1 14B (SMD\&A)} \\
Model URL & `TheFinAI/Fin-o1-14B` \\
LoRA R & 8 \\
LoRA Alpha & 8 \\
LoRA Dropout & 0.05 \\
Layers to Finetune & 40 \\
Max Context & 8500 tokens \\
Batch Size & 4 \\
Gradient Accumulation Steps & 2 \\
Learning Rate & 1e-4 \\
LoRA Target Modules & `q\_proj`, `v\_proj`, `up\_proj`, `down\_proj`, `gate\_proj` \\
\midrule
Common Parameters \\
Epochs (Classic FSFD Task) & 10 \\
Epochs (CI-FSFD Task) & 20 \\
Max New Tokens & 1 \\
Only Completion & True \\
Undersample & True \\
Run Eval on Start & False \\
Use Full Summary & True \\
\bottomrule
\end{tabular}
\end{table}

\subsubsection{Financial+SMD\&A Models (Llama-3.1 8B and Fino1 8B)}
These models are fine-tuned on both Financial (122 features) and Summary Management Discussion \& Analysis (SMD\&A) sections.

\begin{table}[h!]
\caption{Hyperparameters for LLM-based Models (Financial+SMD\&A).} \label{tab:llm_fin_mda_hyperparams}
\centering
\begin{tabular}{ll}
\toprule
Parameter & Value \\
\midrule
\textbf{Llama-3.1 8B (Financial+SMD\&A)} \\
Model URL & `unsloth/Llama-3.1-8B-unsloth-bnb-4bit` \\
LoRA R & 8 \\
LoRA Alpha & 8 \\
LoRA Dropout & 0.05 \\
Layers to Finetune & 32 \\
Learning Rate & 1e-4 \\
LoRA Target Modules & `q\_proj`, `v\_proj`, `up\_proj`, `down\_proj`, `gate\_proj` \\
\midrule
\textbf{Fino1 8B (Financial+SMD\&A)} \\
Model URL & `TheFinAI/Fino1-8B` \\
LoRA R & 8 \\
LoRA Alpha & 8 \\
LoRA Dropout & 0.05 \\
Layers to Finetune & 32 \\
Learning Rate & 1e-4 \\
LoRA Target Modules & `q\_proj`, `v\_proj`, `up\_proj`, `down\_proj`, `gate\_proj` \\
\midrule
Common Parameters \\
Max Context (Classic FSFD Task) & 9500 tokens \\
Max Context (CI-FSFD Task) & 9500 tokens \\
Batch Size (Classic FSFD Task) & 8 \\
Batch Size (CI-FSFD Task) & 4 \\
Gradient Accumulation Steps (Classic FSFD Task) & 1 \\
Gradient Accumulation Steps (CI-FSFD Task) & 2 \\
Epochs (Classic FSFD Task) & 10 \\
Epochs (CI-FSFD Task) & 20 \\
Max New Tokens & 1 \\
Only Completion & True \\
Undersample & True \\
Run Eval on Start & False \\
Use Full Summary & True \\
\bottomrule
\end{tabular}
\end{table}

\subsubsection{Zero-shot Model (Fino1 8B SMD\&A)}
For the zero-shot evaluation, the Fino1 8B model was used without any fine-tuning (i.e., `num\_layers\_to\_finetune` is 0).

\begin{table}[h!]
\caption{Hyperparameters for Zero-shot Fino1 8B SMD\&A Model.} \label{tab:llm_mda_zero_shot_hyperparams}
\centering
\begin{tabular}{ll}
\toprule
Parameter & Value \\
\midrule
Model URL & `TheFinAI/Fino1-8B` \\
Max Context & 9500 tokens \\
Max New Tokens & 1 \\
Batch Size & 1 \\
Zero-Shot & True \\
\bottomrule
\end{tabular}
\end{table}

%% file: appendixes/appendix_6_system_prompts.tex
\section{Appendix F: LLM System Prompts}
\label{appendix:llm_system_prompts}

This appendix details the system prompts used for the Large Language Model (LLM) based classifiers, corresponding to the different input data configurations: Financials Only (FIN), Synthetically Summarized MD\&A Only (SMD\&A), and combined Financials + SMD\&A.

For each configuration, the model was provided with a specific user prompt outlining the task, the context (industry sector), and the relevant data. The LLM was then fine-tuned to generate a single token representing the classification: "YES" (indicating fraud) or "NO" (indicating non-fraud) immediately following the prompt. The prompts were designed to fit within the model's context window, with specific token allowances made for the variable data portions (financial strings or MD\&A content). The token counts provided below are approximate estimates for the fixed textual parts of each prompt, based on the Llama-3 tokenizer, and exclude the tokens from placeholder content like `{industry\_title}`, `{financials\_str}`, or `{mda\_content}`.

\subsection{Prompt for Financials Only (FIN) Input}
When using only financial data, the LLM was presented with the following prompt structure.
\begin{tcolorbox}[enhanced, breakable, colback=red!5!white, colframe=red!75!black, title=LLM Prompt: Financials Only, fonttitle=\bfseries, boxsep=2pt, left=2pt, right=2pt]
\small
\begin{verbatim}
You are a financial forensic analyst.
The company operates in the {industry_title} sector.
Below are key financial indicators derived from its income
statement, balance sheet, and cash flow statement:
{financials_str}
Based on these informations and your knowledge of typical
red flags in financial reporting,
assess whether there is a high likelihood that
this company is engaging in Financial Manipulation Fraud.
Do you think this company is engaging Fraud? Answer with "YES" or "NO"?
\end{verbatim}
\end{tcolorbox}

\subsection{Prompt for SMD\&A Only (SMD\&A) Input}
When using only the Synthetically Summarized MD\&A text, the following prompt structure was employed..

\begin{tcolorbox}[enhanced, breakable, colback=blue!5!white, colframe=blue!75!black, title=LLM Prompt: SMD\&A Only, fonttitle=\bfseries, boxsep=2pt, left=2pt, right=2pt]
\small
\begin{verbatim}
The company operates in the {industry_title} sector.

Below is the summary of the Management Discussion and
Analysis (MDA) section of the quarterly report:
{mda_content}

Based on these informations and your knowledge of
typical red flags in financial reporting,
assess whether there is a high likelihood that
this company is Financial Manipulation Fraud.

Do you think this company is engaging Fraud?
Answer with "YES" or "NO"?
\end{verbatim}
\end{tcolorbox}

\subsection{Prompt for Combined Financials + SMD\&A Input}
For the combined input scenario, the LLM received a prompt integrating both financial metrics and the SMD\&A content.
\begin{tcolorbox}[enhanced, breakable, colback=green!5!white, colframe=green!75!black, title=LLM Prompt: Financials + SMD\&A, fonttitle=\bfseries, boxsep=2pt, left=2pt, right=2pt]
\small
\begin{verbatim}
The company operates in the {industry_title} sector.

Here are financial variables derived from the
income statement, balance sheet, and cash flow statement of the company.
{financials_str}

Also below is the  structured summary of the
Management Discussion and Analysis (MDA)
section of the quarterly report:
{mda_content}

Based on these informations and your knowledge
of typical red flags in financial reporting,
assess whether there is a high likelihood
that this company is Financial Fraud.
Do you think this company is engaging
Fraud? Answer with "YES" or "NO"?
\end{verbatim}
\end{tcolorbox}

%% file: appendixes/appendix_7_sample_predictions.tex
\section{Appendix G: Sample Prediction Details}
\label{appendix:sample_predictions}

This appendix provides an illustrative example of a prediction made by our Fino-1 8B
model using the combined Financial  + SMD\&A input. It shows the complete prompt
provided to the model (truncated for brevity in this display, but the full content
was used for prediction), the model's generated answer, the ground truth label,
and the associated prediction probabilities.

The following example corresponds to a firm-quarter instance from the CI-FSFD task,
where the model correctly identified a fraudulent case.

\begin{tcolorbox}[
    enhanced,
    breakable,
    colback=blue!5!white,
    colframe=blue!75!black,
    title=Sample Model Input Prompt (Financials + SMD\&A - Truncated for Display),
    fonttitle=\bfseries\small,
    boxsep=2pt,
    left=4pt,
    right=4pt,
    top=4pt,
    bottom=4pt,
    arc=2mm,
    boxrule=0.5pt,
    ]%
\footnotesize 
\texttt{
The company operates in the RUBBER \& PLASTICS FOOTWEAR sector. \\
\\
Here are financial variables derived from the income \\
statement, balance sheet, and cash flow statement \\
of the company.\\
- Total Assets: \$22,921,000,000\\
- Cash and Short-term Investments: \$3,695,000,000\\
- Property, Plant, and Equipment: \$4,688,000,000\\
- Degree Of Financial Leverage: 873\%\\
- Invested Capital Ratio: 20\%\\
- Cash To Total Asset: 16\%\\
- Debt Service Coverage: 36\%\\
- Financial LeverageIndex: 5.44\%\\
- Times InterestEarnedRatio: 9,275\%\\
- Current Asset To Revenues: 164\%\\
- Current Liabilities To Revenues: 76\%\\
- Short TermDebt To Revenue: 0.23\%\\
- Intangible Asset ToRevenue: 4.55\%\\
- LongtermLeverage: 15\%\\
- Asset Quality Index: 0.96\\
- Leverage Index: 0.99\\
\dots \\
\\
Also below is the structured summary of the \\
Summary Management Discussion and Analysis (SMD\&A) \\
section of the quarterly report:\\
\\
\# 1. Strategic Priorities and Initiatives\\
\\
– NIKE’s goal is to deliver value to shareholders by building a profitable global \\
portfolio of branded footwear, apparel, equipment, and accessories businesses.\\
– The company’s strategy is to achieve long-term revenue growth by creating \\
innovative, must-have products, building deep personal consumer connections \\
with its brands, and delivering compelling consumer experiences through digital \\
platforms and at retail.\\
– In fiscal 2018, NIKE introduced the Consumer Direct Offense, a new company \\
alignment designed to allow NIKE to better serve the consumer more personally, \\
at scale.\\
– Through the Consumer Direct Offense, NIKE is focusing on the Triple Double \\
strategy, with the objective of doubling the impact of innovation and increasing \\
its speed to market and direct connections with consumers.\\
\\
---\\
\\
\# 2. Operational and Segment Performance\\
\\
– For the third quarter of fiscal 2019, NIKE Brand delivered 8\% revenue growth, \\
with 12\% growth on a currency-neutral basis, driven by higher revenues across \\
all geographies, footwear and apparel, as well as growth in most key categories, \\
led by Sportswear and the Jordan Brand.\\
– Converse revenues decreased 4\% on a reported basis and 2\% on a \\
currency-neutral basis, primarily due to declines in the U.S. and Europe, \\
partially offset by revenue growth in Asia.\\
– In North America, on a currency-neutral basis, revenues increased 7\% for the \\
third quarter and first nine months of fiscal 2019, driven by growth in several \\
key categories for the quarter and nearly all key categories for the year-to-date \\
period, led by Sportswear.\\
– NIKE Direct in North America increased 6\% and 7\% for the third quarter and \\
first nine months, respectively, as higher digital commerce sales and the addition \\
of new stores more than offset an 8\% and 4\% decline in comparable store sales, \\
driven by NFS performance.\\
– In EMEA, on a currency-neutral basis, revenues grew 12\%, driven by balanced \\
growth across all territories and led by Sportswear and the Jordan Brand.\\
– NIKE Direct in EMEA increased 15\% and 14\% for the third quarter and first \\
nine months, respectively, due to comparable store sales growth, higher digital \\
commerce sales, and new store additions.\\
– In Greater China, on a currency-neutral basis, revenues increased \dots \\
\\
---\\
\\
\# 3. Financial Results and Key Trends\\
\\
– For the third quarter of fiscal 2019, revenues increased 7\% to \$9.6 billion, and \\
net income was \$1.1 billion with diluted earnings per share of \$0.68, compared \\
to a net loss of \$921 million and diluted loss per share of \$0.57 for the same \\
period in fiscal 2018.\\
– Income before income taxes increased 11\%, driven by revenue growth and \\
gross margin expansion, partially offset by higher selling and administrative expense.\\
– Gross margin increased to 45.1\% for the third quarter of fiscal 2019, compared \\
to 43.8\% in the same period in fiscal 2018.\\
– For the first nine months of fiscal 2019, revenues increased 9\% to \$28.9 billion, \\
and net income was \$3.0 billion, compared to \$3.1 billion in the prior year.\\
– Gross margin for the nine months ended February 28, 2019, was 44.4\%, \\
compared to 43.5\% in the prior year.\\
– Selling and administrative expense increased to \$3.09 billion for the third \\
quarter of fiscal 2019, representing 32.2\% of revenues, compared to \$2.767 billion \\
and 30.8\% of revenues for the third quarter of fiscal 2018.\\
– For the first nine months of fiscal 2019, selling and administrative expense \\
increased to \$9.296 billion, representing 32.1\% of revenues, compared to \\
\$8.391 billion and 31.5\% of revenues in the prior year.\\
\dots \\
---\\
\\
\# 4. Identified Risks and Uncertainties\\
\\
– The company is exposed to foreign currency market volatility, partly due to \\
global trade uncertainty and geopolitical dynamics.\\
– Foreign currency exposures arise from transactions denominated in \\
non-functional currencies and the translation of foreign currency-denominated \\
results into U.S. Dollars.\\
– Argentina has been identified as a hyper-inflationary market, and the functional \\
currency of the Argentina subsidiary was changed to U.S. Dollars in the second \\
quarter of fiscal 2019.\\
– The translation of foreign currency-denominated profits and foreign exchange \\
rate fluctuations had an unfavorable impact on income before income taxes for \\
both the third quarter and first nine months of fiscal 2019.\\
– The functional currency change in Argentina did not have a material impact on \\
the Company’s results of operations or financial condition, and management does \\
not anticipate a material impact in future periods based on current rates.\\
– The Company may face challenges in accessing credit markets or increased \\
interest costs due to future volatility.\\
– Foreign currency hedge gains and losses may impact operating performance, \\
depending on actual market rates versus standard rates.\\
\dots\\
\\
---\\
\\
\# 5. Forward-Looking Statements and Guidance\\
\\
– The company remains committed to its long-term financial goals, and continues \\
to see opportunities to drive growth and profitability despite foreign currency volatility.\\
– NIKE Direct is expected to continue accelerating growth, driven by digital \\
commerce and store expansion.\\
– Investments in data and analytics, digital commerce platforms, and a new \\
enterprise resource planning tool are part of the end-to-end digital transformation \\
strategy.\\
– The company plans to continue share repurchases under its new \$15 billion \\
four-year program, with funding expected from operating cash flows, excess cash, \\
and debt proceeds.\\
– Management believes that existing cash, cash equivalents, short-term investments, \\
and cash generated by operations, along with access to external funding, will be \\
sufficient to meet capital needs in the foreseeable future. \\
\dots\\
\\
---\\
\\
\# 6. Significant Changes, Events, or Developments\\
\\
– The Company completed the \$12 billion share repurchase program authorized \\
in November 2015 during the first nine months of fiscal 2019, repurchasing \\
192.1 million shares.\\
– A new four-year, \$15 billion share repurchase program was authorized in \\
June 2018, and 43.7 million shares were repurchased under this program \\
during the first nine months of fiscal 2019.\\
– The functional currency of the Argentina subsidiary was changed to U.S. Dollars \\
in the second quarter of fiscal 2019, due to hyper-inflationary conditions.\\
\dots\\
\\
---\\
\\
\# 7. Important Figures and Tables\\
\\
Revenues for the Three Months Ended February 28, 2019 and 2018\\
\\
| Period | Revenues (in millions) | \% Change | \% Change \\
|--------|------------------------|----------|-----------\\
| 2019   | \$9,611                 | —        | 11\%      \\
| 2018   | \$8,984                 | 7\%       | —         \\
\\
Revenues for the Nine Months Ended February 28, 2019 and 2018\\
\\
| Period | Revenues (in millions) | \% Change | \% Change \\
|--------|------------------------|----------|-----------\\
| 2019   | \$28,933                | —        | 11\%      \\
| 2018   | \$26,608                | 8\%       | —         \\
\\
Gross Profit and Gross Margin for the Three Months Ended February 28\\
\\
| Period | Gross Profit (in millions) | \% Change | Gross Margin |\\
|--------|----------------------------|----------|--------------|\\
| 2019   | \$4,339                     | —        | 45.1\%        |\\
| 2018   | \$3,938                     | 10\%      | 43.8\%        |\\
\dots\\
---\\
\\
\# 8. Management Explanations and Justifications\\
\\
– The increase in NIKE Brand revenues is attributed to growth across all \\
geographies, footwear and apparel, and key categories like Sportswear \\
and the Jordan Brand.\\
– The decline in Converse revenues is explained by revenue declines in the \\
U.S. and Europe, partially offset by growth in Asia.\\
\dots\\
\\
---\\
\\
\# 9. Accounting Estimates, Judgments, and Policy Changes\\
\\
– Revenue recognition is based on transfer of control to the customer, with \\
variable consideration for sales returns, discounts, and miscellaneous claims \\
estimated and recorded as a reduction to revenues.\\
\dots\\
\\
---\\
\\
\# 10. Capital Allocation and Liquidity Management\\
\\
– Share repurchase activity increased significantly in fiscal 2019, with \\
\$3.386 billion spent on 43.7 million shares.\\
\dots\\
\\
---\\
\\
\# 11. Legal, Regulatory, and Compliance Matters\\
\\
– The Company has no off-balance sheet arrangements that have or are \\
reasonably likely to have a material effect on financial condition, \\
results of operations, liquidity, or capital resources.\\
\dots\\
\\
Based on these informations and your knowledge of typical \\
red flags in financial reporting, assess whether there is a \\
high likelihood that this company is Financial Fraud.\\
Do you think this company is engaging Fraud? Answer with \\
"YES" or "NO"?
} 
\end{tcolorbox}

\vspace{0.5cm} 

\begin{tcolorbox}[
    enhanced,
    colback=green!5!white,
    colframe=green!65!black,
    title=Model Prediction and Ground Truth,
    fonttitle=\bfseries\small,
    boxsep=3pt,
    left=5pt,
    right=5pt,
    arc=2mm,
    boxrule=0.5pt,
    width=\linewidth 
    ]
\small 
\textbf{Model's Generated Answer:} \texttt{NO} \medskip
\textbf{Ground Truth Label:} \texttt{NO} \medskip

\textbf{Prediction Probability for "YES":} \texttt{0.0073334336280823} \medskip
(Note: The decision threshold optimized on the validation set for this fold was
applied to this probability to yield the binary prediction.) \medskip

\textbf{Instance Identifiers:}
\begin{itemize}
    \item CIK: \texttt{320187}
    \item SIC (Aggregated): \texttt{3021} (RUBBER \& PLASTICS FOOTWEAR)
    \item Quarter: \texttt{2019q3}
\end{itemize}
\end{tcolorbox}

This example illustrates how the model processes the combined textual and numerical
information to arrive at a classification decision. The relatively low probability
for a correct "YES" prediction in this specific case, despite being above the
decision threshold for this fold, highlights the challenging nature of the
CI-FSFD task.

%% file: appendixes/appendix_8_subgroup_performance_analysis.tex
\section{Appendix H: Detailed Performance Analysis }
\label{appendix:subgroup_performance_analysis}

This appendix provides a more granular look at the performance of our LLM-based fraud detection framework (Fino-1 8B with SMD\&A input) across different subgroups: individual companies, industry sectors, and performance on unseen companies in the classic setting. We present key metrics such as True Positives (TP - Detected Fraud), False Negatives (FN - Undetected Fraud), False Positives (FP - Non-Fraud Incorrectly Flagged as Fraud), Total Actual Fraud instances, Recall (TP / (TP + FN)), and Precision (TP / (TP + FP)).

\subsection{Fine-Grain Labels Performance Analysis}
\label{appendix:fine_grain_analysis}

Table \ref{tab:appendix_fine_grain_labels} presents a detailed breakdown of the model's performance on detecting different types of financial statement misstatements. The model exhibits varying recall across different misstatement categories, indicating that certain types of fraud are more challenging to detect than others.

\begin{table}[htbp]
  \centering
  \caption{Fraud Detection Performance by Misstatement Type (CI-FSFD Task). Model: Fino-1 8B with SMD\&A.}
  \label{tab:appendix_fine_grain_labels}
  \resizebox{\textwidth}{!}{%
  \begin{tabular}{@{}lrrrrr@{}}
    \toprule
    Misstatement Type                 & Detected Fraud (TP) & Undetected Fraud (FN) & Total Actual Fraud & Recall & AUC \\ \midrule
    mis\_Reserve Account               & 7                   & 4                     & 11                 & 0.636  & 0.819 \\
    mis\_Capitalized Costs as Assets  & 7                   & 8                     & 15                 & 0.467  & 0.796 \\
    mis\_Cost of Goods Sold (COGS)    & 15                  & 24                    & 39                 & 0.385  & 0.669 \\
    mis\_Accounts Receivable          & 25                  & 52                    & 77                 & 0.325  & 0.756 \\
    mis\_Allowance for Bad Debt       & 1                   & 3                     & 4                  & 0.250  & 0.955 \\
    mis\_Liabilities                  & 19                  & 59                    & 78                 & 0.244  & 0.689 \\
    mis\_Revenue                      & 51                  & 160                   & 211                & 0.242  & 0.707 \\
    mis\_Payables                     & 17                  & 56                    & 73                 & 0.233  & 0.698 \\
    mis\_Other Expense/Shareholder Equity Account & 55                  & 210                   & 265                & 0.208  & 0.645 \\
    mis\_Assets Valuation             & 10                  & 85                    & 95                 & 0.105  & 0.536 \\
    mis\_Inventory                    & 2                   & 36                    & 38                 & 0.053  & 0.565 \\
    \bottomrule
  \end{tabular}%
  }
\end{table}

\subsection{Company-Level Performance Analysis (CI-FSFD)}
\label{appendix:company_level_analysis}

The Company-Isolated FSFD (CI-FSFD) task evaluates the model's ability to generalize to companies not seen during training. Performance at the individual company level can vary significantly.

Table \ref{tab:appendix_best_30_companies_cifsfd} lists the top 30 companies where the model demonstrated the best performance (highest recall) in detecting fraudulent quarters under the CI-FSFD setting. Table \ref{tab:appendix_worst_30_companies_cifsfd} shows the 30 companies where the model struggled the most (lowest recall).

\begin{table}[htbp]
  \centering
  \caption{Top 30 Companies by Recall in Detecting Fraudulent Quarters (CI-FSFD Task), with False Positives and Precision. Model: Fino-1 8B with SMD\&A.}
  \label{tab:appendix_best_30_companies_cifsfd}
  \resizebox{\textwidth}{!}{%
  \begin{tabular}{@{}p{0.45\textwidth}rrrrrr@{}} 
  \toprule
  Industry Sector & Detected Fraud (TP) & Undetected Fraud (FN) & False Positives (FP) & Total Actual Fraud & Recall & Precision \\
  \textit{(Company Name)} & & & & & & \\ \midrule
3m company & 9 & 0 & 1 & 9 & 1.000 & 0.900 \\
Akorn, inc. & 2 & 0 & 0 & 2 & 1.000 & 1.000 \\
Hertz & 2 & 0 & 0 & 2 & 1.000 & 1.000 \\
Ixia & 1 & 0 & 0 & 1 & 1.000 & 1.000 \\
Jda software group, inc. & 2 & 0 & 0 & 2 & 1.000 & 1.000 \\
Mcdermott international inc. & 1 & 0 & 0 & 1 & 1.000 & 1.000 \\
Ocz technology group, inc. & 2 & 0 & 0 & 2 & 1.000 & 1.000 \\
Roadrunner transportation systems, inc. & 6 & 0 & 0 & 6 & 1.000 & 1.000 \\
Surgalign holdings, inc. & 2 & 0 & 0 & 2 & 1.000 & 1.000 \\
Swisher hygiene inc. & 1 & 0 & 0 & 1 & 1.000 & 1.000 \\
Taronis technologies, inc. & 1 & 0 & 0 & 1 & 1.000 & 1.000 \\
Cognizant technology solutions corporation & 2 & 0 & 2 & 2 & 1.000 & 0.500 \\
Kbr, inc. & 1 & 0 & 2 & 1 & 1.000 & 0.333 \\
L3 technologies, inc. & 1 & 0 & 2 & 1 & 1.000 & 0.333 \\
Tech data corp. & 1 & 0 & 1 & 1 & 1.000 & 0.500 \\
Uti worldwide inc. & 2 & 0 & 1 & 2 & 1.000 & 0.667 \\
Nci, inc. & 7 & 0 & 1 & 7 & 1.000 & 0.875 \\
Newell brands inc. & 3 & 0 & 1 & 3 & 1.000 & 0.750 \\
Quadrant 4 system corp. & 7 & 1 & 0 & 8 & 0.875 & 1.000 \\
The kraft heinz co. & 4 & 1 & 1 & 5 & 0.800 & 0.800 \\
Halliburton company & 3 & 1 & 0 & 4 & 0.750 & 1.000 \\
Sciclone pharmaceuticals, inc. & 3 & 1 & 0 & 4 & 0.750 & 1.000 \\
Corporate resource services, inc. & 2 & 1 & 0 & 3 & 0.667 & 1.000 \\
Future fintech group inc. & 2 & 1 & 0 & 3 & 0.667 & 1.000 \\
Mimedx group, inc. & 6 & 3 & 0 & 9 & 0.667 & 1.000 \\
Synchronoss technologies, inc. & 9 & 5 & 3 & 14 & 0.643 & 0.750 \\
Granite construction inc. & 5 & 4 & 0 & 9 & 0.556 & 1.000 \\
Apex global brands inc. & 1 & 1 & 0 & 2 & 0.500 & 1.000 \\
Evoqua water technologies corp. & 1 & 1 & 0 & 2 & 0.500 & 1.000 \\
Gt advanced technologies inc. & 1 & 1 & 0 & 2 & 0.500 & 1.000 \\
  \bottomrule
  \end{tabular}%
  }
\end{table}

\begin{table}[htbp]
  \centering
  \caption{Worst 30 Companies by Recall in Detecting Fraudulent Quarters (CI-FSFD Task). Model: Fino-1 8B with SMD\&A.}
  \label{tab:appendix_worst_30_companies_cifsfd}
  \resizebox{\textwidth}{!}{%
  \begin{tabular}{@{}p{0.45\textwidth}rrrrrr@{}} 
  \toprule
  Industry Sector & Detected Fraud (TP) & Undetected Fraud (FN) & False Positives (FP) & Total Actual Fraud & Recall & Precision \\
  \textit{(Company Name)} & & & & & & \\ \midrule
African gold acquisition corp. & 0 & 1 & 0 & 1 & 0.000 & 0.000 \\
Amyris, inc. & 0 & 1 & 0 & 1 & 0.000 & 0.000 \\
Andeavor llc & 0 & 1 & 0 & 1 & 0.000 & 0.000 \\
Argo group international holdings, ltd. & 0 & 15 & 0 & 15 & 0.000 & 0.000 \\
Assisted living concepts, inc. & 0 & 2 & 0 & 2 & 0.000 & 0.000 \\
Axesstel, inc. & 0 & 1 & 0 & 1 & 0.000 & 0.000 \\
Barrett business services, inc. & 0 & 10 & 0 & 10 & 0.000 & 0.000 \\
Belden inc. & 0 & 1 & 0 & 1 & 0.000 & 0.000 \\
Biomet, inc. & 0 & 4 & 0 & 4 & 0.000 & 0.000 \\
Blue earth, inc. & 0 & 2 & 0 & 2 & 0.000 & 0.000 \\
Brixmor property group inc. & 0 & 4 & 0 & 4 & 0.000 & 0.000 \\
Cantaloupe, inc. & 0 & 5 & 0 & 5 & 0.000 & 0.000 \\
Celadon group, inc. & 0 & 3 & 0 & 3 & 0.000 & 0.000 \\
Celsius holdings, inc. & 0 & 2 & 1 & 2 & 0.000 & 0.000 \\
China valves technology, inc. & 0 & 1 & 0 & 1 & 0.000 & 0.000 \\
Chs inc. & 0 & 15 & 0 & 15 & 0.000 & 0.000 \\
Citigroup inc. & 0 & 1 & 0 & 1 & 0.000 & 0.000 \\
Comscore, inc. & 0 & 5 & 0 & 5 & 0.000 & 0.000 \\
Cpi aerostructures, inc. & 0 & 2 & 0 & 2 & 0.000 & 0.000 \\
Dxc technology company & 0 & 2 & 0 & 2 & 0.000 & 0.000 \\
Elanco animal health inc. & 0 & 4 & 0 & 4 & 0.000 & 0.000 \\
Fmc technologies, inc. & 0 & 5 & 0 & 5 & 0.000 & 0.000 \\
Fte networks, inc. & 0 & 2 & 0 & 2 & 0.000 & 0.000 \\
General electric company & 0 & 7 & 0 & 7 & 0.000 & 0.000 \\
General motors company & 0 & 7 & 0 & 7 & 0.000 & 0.000 \\
Gtt communications, inc. & 0 & 2 & 2 & 2 & 0.000 & 0.000 \\
Healthcare services group, inc. & 0 & 4 & 0 & 4 & 0.000 & 0.000 \\
Home loan servicing solutions, ltd. & 0 & 5 & 0 & 5 & 0.000 & 0.000 \\
Homestreet, inc. & 0 & 7 & 0 & 7 & 0.000 & 0.000 \\
Iconix brand group, inc. & 0 & 3 & 0 & 3 & 0.000 & 0.000 \\
  \bottomrule
  \end{tabular}%
  }
\end{table}

\begin{figure}[htbp]
    \centering
    \includegraphics[width=0.9\textwidth]{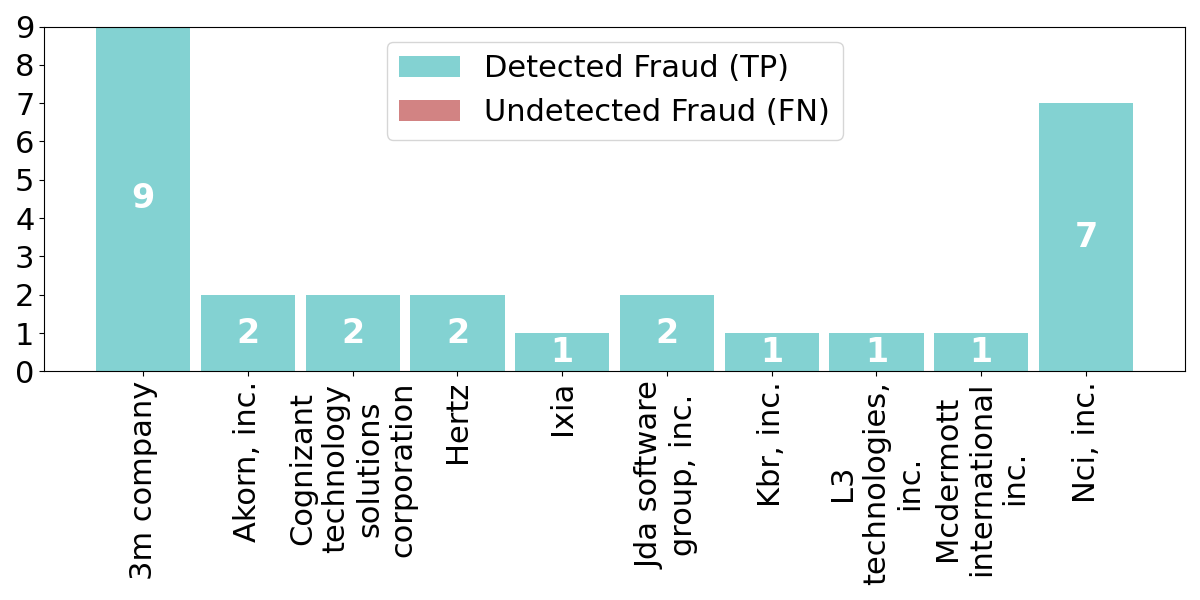}
    \caption{Plot: Top 10 companies by number of correctly detected fraudulent quarters (True Positives vs False Negatives) in the CI-FSFD task. Model: Fino-1 8B with SMD\&A.}
    \label{fig:appendix_best_companies_cifsfd_plot}
\end{figure}

\begin{figure}[htbp]
    \centering
    \includegraphics[width=0.9\textwidth]{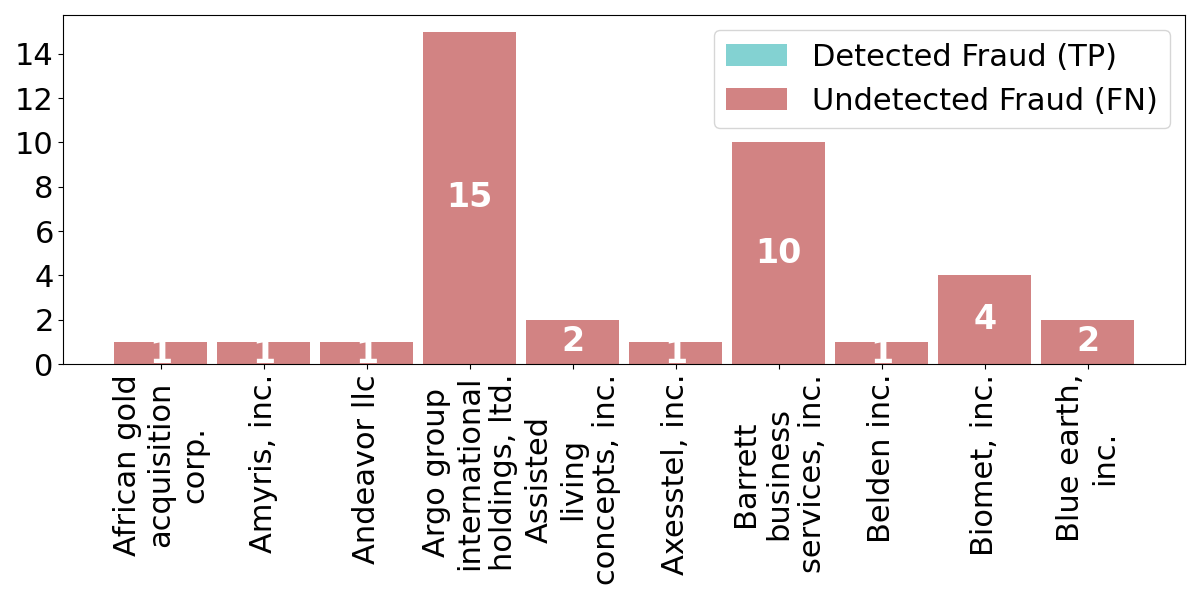}
    \caption{Plot: Top 10 companies by number of undetected fraudulent quarters (True Positives vs False Negatives) in the CI-FSFD task. Model: Fino-1 8B with SMD\&A.}
    \label{fig:appendix_worst_companies_cifsfd_plot}
\end{figure}

The visual representations of True Positives and False Negatives for the top 10 from these company groups are in Figure \ref{fig:appendix_best_companies_cifsfd_plot} and Figure \ref{fig:appendix_worst_companies_cifsfd_plot} respectively.

\FloatBarrier 

\subsection{Sectorial Performance Analysis (CI-FSFD)}
\label{appendix:sectorial_analysis_tables}

Performance also varies when aggregated by industry sector. Table \ref{tab:appendix_sectorial_cifsfd_fp} details the detection performance, including false positives and precision, across major industry sectors for the CI-FSFD task.

\begin{table}[htbp]
  \centering
  \caption{Fraud Detection Performance by Industry Sector (CI-FSFD Task). Model: Fino-1 8B with SMD\&A.}
  \label{tab:appendix_sectorial_cifsfd_fp}
  \resizebox{\textwidth}{!}{%
  \begin{tabular}{@{}lrrrrrr@{}}
    \toprule
    Industry Sector                     & Detected Fraud (TP) & Undetected Fraud (FN) & False Positives (FP) & Total Actual Fraud & Recall & Precision \\ \midrule
Construction & 6 & 4 & 20 & 10 & 0.600 & 0.231 \\
Retail trade & 4 & 6 & 25 & 10 & 0.400 & 0.138 \\
Services & 34 & 74 & 206 & 108 & 0.315 & 0.142 \\
Transportation \& public utilities & 8 & 21 & 47 & 29 & 0.276 & 0.145 \\
Manufacturing & 61 & 198 & 277 & 259 & 0.236 & 0.180 \\
Mining & 4 & 15 & 10 & 19 & 0.211 & 0.286 \\
Wholesale trade & 4 & 20 & 42 & 24 & 0.167 & 0.087 \\
Finance, Insurance, \& Real Estate & 0 & 52 & 57 & 52 & 0.000 & 0.000 \\
    \midrule
    \textbf{Overall (CI-FSFD Task Total)} & \textbf{121} & \textbf{390} & \textbf{684} & \textbf{511} & \textbf{0.237} & \textbf{0.150} \\
    \bottomrule
  \end{tabular}%
  }
  \caption*{\small Note: The "Overall" row aggregates TP, FN, FP, and Actual Fraud across all test folds for the CI-FSFD task for the listed sectors and calculates overall Recall and Precision from these sums.}
\end{table}

\begin{figure}[htbp]
    \centering
    \includegraphics[width=0.9\textwidth]{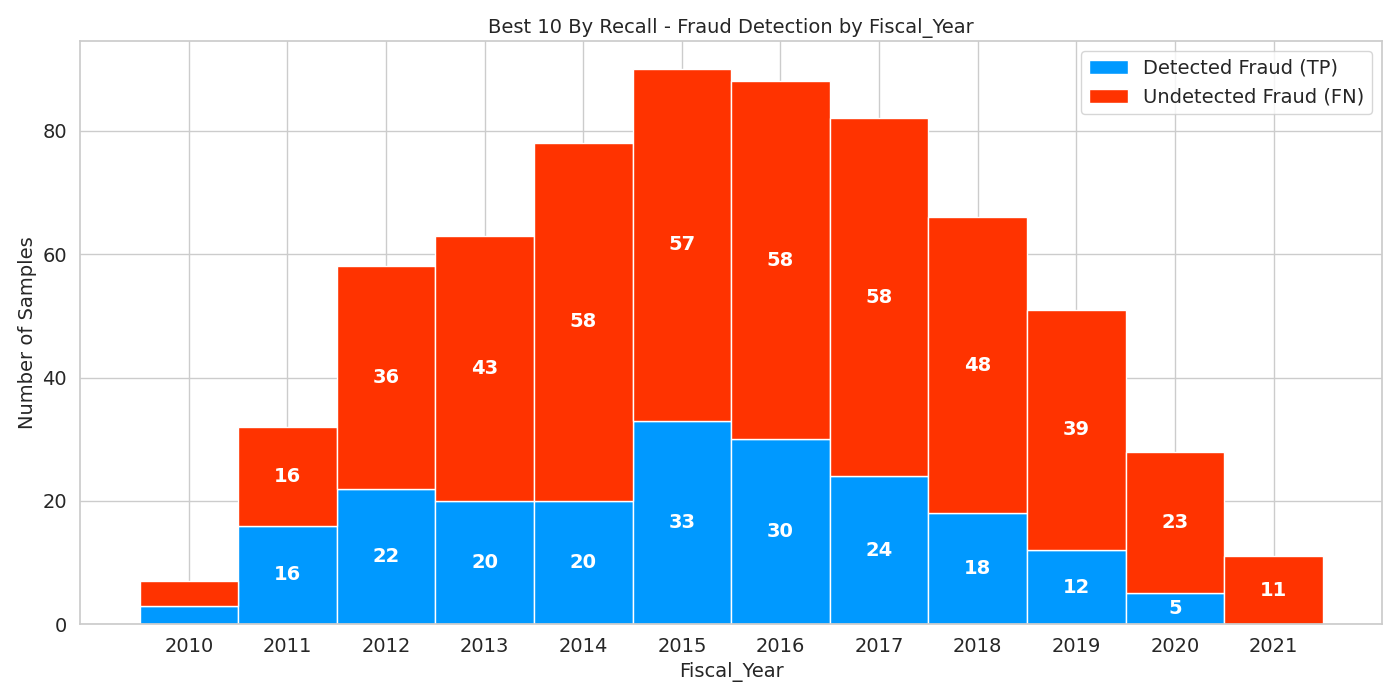}
    \caption{Plot: Detected Fraud (TP) vs. Undetected Fraud (FN) cases per industry sector in the CI-FSFD task. Model: Fino-1 8B with SMD\&A.}
    \label{fig:appendix_sectorial_cifsfd_plot}
\end{figure}

Figure \ref{fig:appendix_sectorial_cifsfd_plot} visually represents the True Positives and False Negatives by sector.

\FloatBarrier

\subsection{Performance on Unseen Companies in Classic FSFD Setting}
\label{appendix:classic_unseen_analysis}

While the Classic FSFD setting involves random splitting, a small fraction of companies in the test set of each fold might still be entirely unseen during the training phase for that specific fold. Analyzing performance on these truly "unseen" companies within the classic random split provides insight into the model's baseline generalization even when not explicitly forced by a company-isolated split.

Table \ref{tab:appendix_classic_unseen_metrics} summarizes key average metrics for the Fino-1 8B (SMD\&A) model on these unseen company instances within the Classic FSFD's 5-fold cross-validation.

\begin{table}[htbp]
  \centering
  \caption{Average Performance Metrics on Unseen Companies within Classic FSFD Test Folds. Model: Fino-1 8B with SMD\&A (Averages over 5 Folds).}
  \label{tab:appendix_classic_unseen_metrics}
  \begin{tabular}{@{}lc@{}}
    \toprule
    Metric                                       & Mean Value $\pm$ Std. Dev. \\ \midrule
    Avg. Test Samples per Fold                   & 2604.2 $\pm$ 0.4 \\
    Avg. Unseen Samples in Test Fold             & 757.6 $\pm$ 18.4 \\
    Avg. Test CIKs per Fold                      & 2163.2 $\pm$ 9.4 \\
    Avg. Unseen CIKs in Test Fold                & 672.6 $\pm$ 14.1 \\
    Avg. Total Fraud Samples in Test Fold        & 130.8 $\pm$ 15.7 \\
    Avg. Unseen Fraud Samples in Test Fold       & 1.6 $\pm$ 1.2 \\
    Avg. Fraud Rate among Unseen Samples         & 0.0021 $\pm$ 0.0015 \\
    \midrule
    F1 Score (on Unseen Samples)                 & 0.0 $\pm$ 0.0 \\
    Recall (on Unseen Fraud Samples)             & 0.0 $\pm$ 0.0 \\
    Precision (on Unseen Samples)                & 0.0 $\pm$ 0.0 \\
    AUC Score (on Unseen Samples)                & 0.646 $\pm$ 0.280 \\
    Accuracy (on Unseen Samples)                 & 0.9924 $\pm$ 0.0022 \\
    \bottomrule
  \end{tabular}%
  \caption*{\small The performance metrics such as F1 score, Recall, and Precision for unseen fraud cases are not statistically significant due to the very low average number of unseen fraud samples (1.6 per fold) in this random splitting setting. This underscores the importance of dedicated CI-FSFD  for robustly evaluating generalization.}
\end{table}

The very low average number of unseen fraudulent instances (1.6 per fold) in the Classic FSFD setting makes it difficult to draw firm conclusions about the model's ability to detect fraud in entirely new companies from these specific metrics (F1, Recall, Precision).

%% file: appendixes/appendix_9_explainability.tex
\section{Appendix I: Explainability }
\label{appendix:explainability}

This appendix details the methodology employed for generating explanations of our LLM's predictions, specifically focusing on the textual components of the input. Understanding which parts of the financial text contribute most to a fraud prediction is crucial for interpretability and trustworthiness in high-stakes domains like financial anomaly detection.

\subsection{LRP-based Explanation}
Our explainability approach is built upon the Layer-wise Relevance Propagation (LRP) \cite{pmlr-v235-achtibat24a} framework. LRP is a technique used to decompose the prediction of a deep neural network into contributions of its input features. It assigns a "relevance score" to each input component (e.g., a token) indicating its importance to the final output. For a classification task, LRP propagates the prediction score backward through the network, layer by layer, until it reaches the input features. The core idea is to conserve the total relevance during propagation, ensuring that the sum of relevances at one layer equals the sum of relevances at the preceding layer. This property allows for a clear attribution of the final prediction to individual input elements.

In our implementation, we leverage the LXT library, which provides an efficient and specialized LRP implementation for Transformer-based models. After the model makes a prediction (i.e., outputs logits for "Fraud" or "Not Fraud"), we backpropagate the relevance from the logit corresponding to the predicted class (or, more specifically, the 'Fraud' logit, regardless of the prediction, to understand drivers of potential fraud) back to the input embeddings. The LRP rules applied ensure that the relevance scores accurately reflect the contribution of each token in the input sequence to that specific logit.

The process for generating LRP explanations for each test sample is as follows:
\begin{itemize}

\item The trained LLM is set to evaluation mode, and all its parameters are frozen ) for the input embeddings, which are set to requires\_grad=True to compute gradients for LRP.
\item For each test sample, the input prompt (containing the textual and numerical financial data) is tokenized and fed into the model.
\item The model performs a forward pass to obtain the logits for the FRAUD\_LABEL\_ID and NOT\_FRAUD\_LABEL\_ID tokens at the last position of the output sequence.
\item The gradient of the FRAUD\_LABEL\_ID logit (representing the unnormalized score for the "Fraud" class) with respect to the input embeddings is computed.
\item The LRP relevance score for each input token is then calculated as the element-wise product of the input embeddings and their corresponding gradients, summed across the embedding dimensions. This yields a single relevance score for each token.
\item These raw relevance scores are then normalized by their absolute maximum to scale them between -1 and 1, facilitating easier interpretation.
\end{itemize}

\subsection{Token-Level vs. Sentence-Level Relevance}
Given the nature of our input data, which consists of long financial reports (averaging around 3800 tokens per context), providing token-level relevance scores directly to a human analyst can be overwhelming and impractical for actionable insights. A raw sequence of 3800 token relevance scores does not immediately highlight the key information at a glance. Therefore, focusing on individual token relevance is not the most pertinent approach for interpretability in this context.

Instead, we emphasize sentence-level relevance. Financial analysts typically review reports section by section, and understanding which sentences or clauses are most indicative of fraud is significantly more valuable than knowing the exact contribution of every single token. Sentence-level aggregation allows for a higher-level summary of the model's reasoning, making the explanations more digestible and actionable.

\subsection{Sentence-Level Relevance Aggregation}
To derive sentence-level relevance from the token-level scores, we implemented a simple yet effective aggregation method:
\begin{itemize}
\item \textbf{Summing of Relevance Scores:} For every sentence, we sum the absolute relevance scores of all tokens belonging to that sentence. Using the absolute sum helps identify sentences that strongly contribute, either positively or negatively, to the fraud prediction.
\item \textbf{Normalization and Ranking:} The summed relevance scores for sentences are then normalized and ranked. Sentences with higher absolute summed relevance are considered more impactful on the model's prediction.
\end{itemize}

This aggregation provides a concise summary of the most relevant sentences within a lengthy financial disclosure, allowing users to quickly pinpoint suspicious statements or critical pieces of information that drove the model's classification decision.

\subsection{Highlighted Examples}
For visual interpretability, we generate PDF heatmaps that highlight the most relevant portions of the input. These heatmaps use color intensity to represent the magnitude of a token's (or aggregated word's) relevance score, with different hues indicating positive or negative contributions to the fraud prediction.

\begin{figure}[h]
\centering
\includegraphics[width=0.8\columnwidth]{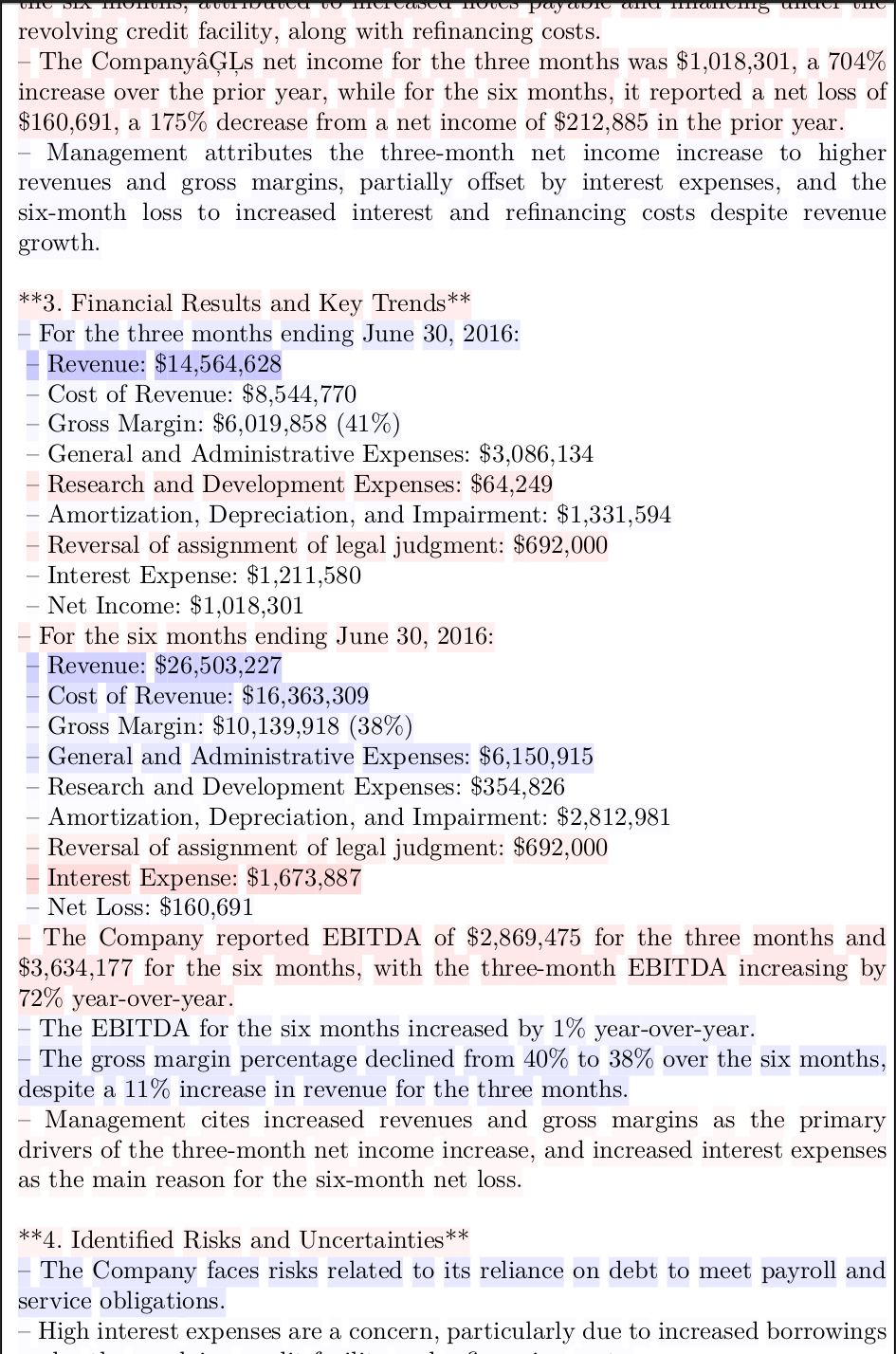}
\caption{Example of a highlighted financial report section indicating token-level relevance for a fraud prediction. Red indicates higher positive relevance towards a "Fraud" prediction, while blue indicates negative relevance.}
\label{fig:highlighted_example}
\end{figure}